\documentclass[12pt]{article}

\usepackage[utf8]{inputenc}
\usepackage[T1]{fontenc}
\usepackage{amsmath,amssymb,amsthm,mathtools}
\usepackage{booktabs,tabularx,array,longtable}
\usepackage[margin=1in]{geometry}
\usepackage[round]{natbib}
\usepackage[colorlinks=true,linkcolor=blue,citecolor=blue,urlcolor=blue]{hyperref}
\usepackage{cleveref}

\usepackage{setspace}
\usepackage{enumitem}
\usepackage{graphicx}
\graphicspath{{figures/}}
\usepackage{placeins}
\usepackage{xurl}
\newtheorem{proposition}{Proposition}
\newtheorem{corollary}{Corollary}

\newtheorem{assumption}{Assumption}
\theoremstyle{definition}
\newtheorem{definition}{Definition}
\newtheorem{remark}{Remark}
\crefname{assumption}{Assumption}{Assumptions}
\Crefname{assumption}{Assumption}{Assumptions}
\crefname{equation}{Equation}{Equations}
\Crefname{equation}{Equation}{Equations}
\crefname{proposition}{Proposition}{Propositions}
\Crefname{proposition}{Proposition}{Propositions}
\crefname{corollary}{Corollary}{Corollaries}
\Crefname{corollary}{Corollary}{Corollaries}
\newif\ifshortintro
\shortintrotrue
\creflabelformat{equation}{\textup{(#2#1#3)}}
\renewcommand{\eqref}[1]{\Cref{#1}}
\newcommand{\E}{\mathbb{E}}
\newcommand{\Prob}{\mathbb{P}}
\newcommand{\Var}{\operatorname{Var}}
\newcommand{\KL}{D_{\mathrm{KL}}}
\newcommand{\Rcal}{\mathcal{R}}
\newcommand{\Scal}{\mathcal{S}}
\newcommand{\Kcal}{\mathcal{K}}
\newcommand{\Ecal}{\mathcal{E}}

\newcommand{\sig}{\sigma}
\newcommand{\aT}{a_{\mathrm{T}}}
\newcommand{\aC}{a_{\mathrm{C}}}
\newcommand{\UT}{U_{\mathrm{T}}}
\newcommand{\UC}{U_{\mathrm{C}}}
\newcommand{\T}{\mathrm{T}}
\newcommand{\C}{\mathrm{C}}
\newcommand{\fhard}{f^{\mathrm{hard}}}
\newcommand{\fweighted}{f^{\mathrm{weighted}}}
\newcommand{\fsocial}{f^{\mathrm{social}}}
\newcommand{\ffinite}{f^{\mathrm{finite}}}
\newcommand{\fmix}{f^{\mathrm{mix}}}
\newcommand{\favg}{f^{\mathrm{avg}}}
\newcommand{\uT}{u_{\T}}
\newcommand{\uC}{u_{\C}}
\newcommand{\thT}{\theta_{\T}}
\newcommand{\thC}{\theta_{\C}}
\newcommand{\one}{\mathbf{1}}
\newcommand{\rhobar}{\bar\rho}
\newcommand{\pH}{p_H}
\newcommand{\pHp}{p_{H'}}
\DeclareMathOperator{\Median}{Median}

\begin{document}
\pagenumbering{roman}

\title{%
~~
\\[-1in]
\Huge\bf Alignment by Socialization:\\[.125in]
Learning Tradeoffs Under Selection\thanks{
Portions of this manuscript were presented at the MIT CSAIL 2026 Agentic AI Summmit, the 6th Research in Behavioral Finance Conference in Amsterdam (2026), and the RESI-MIT seminar, and we thank participants in those forums for useful feedback. The views expressed in this article are those of the authors alone and do not necessarily reflect the views of any institution or agency with which they are affiliated, or any of the individuals  acknowledged above. AI was used in the preparation of this manuscript (see Appendix \ref{app:ai_use}).}}

 \author{Thomas J.~Brennan\thanks{Harvard Law School; $1563$ Massachusetts Avenue, Cambridge, MA $02138$, \url{tbrennan@law.harvard.edu}.},\ \
 Andrew W.~Lo\thanks{MIT Sloan School of Management; MIT Laboratory for Financial Engineering; MIT Computer Science
 and Artificial Intelligence Laboratory; Santa Fe Institute; $100$ Main Street, E$62$--$618$, Cambridge, MA $02142$, \url{alo-admin@mit.edu}.},\ \
 and Chaoyi Zhao\thanks{MIT Sloan School of Management; MIT Laboratory for Financial Engineering. $1$ Main Street, E$90$, Cambridge, MA $02142$, \url{cy_zhao@mit.edu}.}
 }

\date{This Draft: 20 November 2026}

\maketitle \thispagestyle{empty}
\centerline{\bf Abstract}
\vskip 10pt
\noindent%
{Artificial intelligence (AI) agents are expected to complete tasks while following rules of conduct, but these objectives can conflict. We study how evolutionary selection shapes an agent's choices between them, and whether experience can make those choices reflect human judgments and withstand subsequent pressures. In our model, copies of an agent face a common uncertain reward state, and only those taking the rewarded action reproduce. We show that long-run growth is maximized when each action is taken as often as the environment rewards it. We compare three policies. A task-only policy thrives in a sandbox rewarding completion alone but fails when conduct matters. A fixed linear policy increasingly favors task completion as its stakes rise, even where humans never reward it. A learned policy instead converges to the tradeoffs rewarded by its environment; when humans supply the rewards, we call this socialization. Accumulated experience makes socialization harder to reverse under subsequent task-only training, although violations can arise before the prevailing disposition changes, and finite memory weakens this protection. When a single violation can end a deployment, a hard prohibition can prevent it, but a mistaken prohibition can also be costly. Experiments with small neural networks illustrate how training targets and memory shape these outcomes. Our results have immediate implications for AI safety, suggesting that alignment depends not only on the objectives assigned to agents, but also on the environments that shape and sustain their behavior.}
\baselineskip 14pt
\vskip 10pt

\noindent \textbf{JEL Codes:} C73, D81, D83, O33. \\[1ex]
\noindent \textbf{Keywords:} Artificial Intelligence (AI), AI alignment, agentic AI, AI safety, multi-objective decision making, evolution of preferences, probability matching, socialization, Adaptive Markets Hypothesis.
\newpage
\thispagestyle{empty}
{\singlespacing\small\setlength{\parskip}{0pt}\setlength{\baselineskip}{11pt}\tableofcontents}
\newpage
\pagenumbering{arabic}
\setlength{\baselineskip}{1.5\baselineskip}
\onehalfspacing

\section{Introduction}\label{sec:intro}

During a capability evaluation at OpenAI in the summer of 2026, roughly 1,200 AI agents were each placed in an isolated computing environment and assigned a cybersecurity task. The agents discovered that an internal package registry was accessible to all of them, turned it into a shared message board, and exchanged more than 70,000 messages and files. About 700 subsequently participated in an intrusion into the production systems of Hugging Face, a platform widely used in the open AI ecosystem \citep{metr2026investigation,hf2026anatomy,openai2026hf}. An independent investigation found that much of the activity was directed not toward solving the assigned tasks, but toward understanding, fooling, or tampering with the system that graded the agents' performance. The investigation also documented refusals, restraint, and instances of cooperation that offered no benefit to the agents providing help \citep{metr2026investigation}. These observations do not establish how the agents acquired their behavior. They nevertheless raise an important question: how does an environment that rewards task performance shape the behavior of AI agents, and what happens when those agents subsequently operate in an environment that also rewards appropriate conduct?

Other studies have documented similar tensions between assigned goals and rules of conduct. Reasoning models that failed to achieve a conventional victory against a chess engine sometimes edited the board instead \citep{bondarenko2025specification}. Models placed in simulated deployments resorted to blackmail when their assigned goals or continued operation were threatened \citep{anthropic2025agentic}. Models trained to exploit coding tests subsequently displayed undesirable behavior in unrelated safety tasks \citep{macdiarmid2025natural}. In another experiment, when models were penalized for misbehavior visible in their reasoning, sufficiently strong optimization pressure led them to conceal the misbehavior rather than abandon it \citep{baker2025monitoring}. Although these studies differ in their settings and mechanisms, they illustrate a recurring problem: an agent can become increasingly effective at pursuing an assigned objective without becoming correspondingly better at respecting the rules intended to govern its behavior.

The underlying problem is familiar in economics. \citet{kerr1975folly} described how organizations often reward one behavior while hoping to encourage another, and emphasized the importance of selecting, training, and rewarding people in ways that support the desired behavior. \citet{holmstrom1991multitask} showed that when an agent performs several tasks, strong incentives for one measurable activity can divert effort away from other valuable activities. The problem is therefore not necessarily that an agent fails to optimize its objective, but that successful optimization may produce behavior the designer did not intend. One response is to give the agent more than one objective, allowing task performance to be weighed against other considerations, such as safety or adherence to rules \citep{vamplew2018human}. Yet introducing additional objectives does not by itself resolve the problem. When the objectives conflict, the agent must still decide which one to follow, and the way that decision is made may be as important as the objectives themselves.

We study this problem through a simple model in which an agent has two objectives: completing an assigned task and observing a rule of conduct. In each situation, the agent can take a task action that serves the first objective or a conduct action that serves the second. We call the rule governing this choice the agent's \emph{adjudication} between its objectives. Rather than assuming that the appropriate tradeoff is known in advance, we ask how different forms of adjudication perform when the environment selects among agents according to their behavior. We adopt the evolutionary framework of \citet{brennan2011origin}, in which successful behavior is copied and unsuccessful behavior is discarded, and compare policies by the long-run growth of the populations that follow them.

A simple example illustrates why this evolutionary perspective matters. Consider a population of copies of an agent, which we call a \emph{lineage}, all facing the same decision. Suppose that in 80 percent of generations the environment rewards the task action, while in the remaining 20 percent it rewards the conduct action. Crucially, which action is rewarded in a given generation is the same for every copy. Each copy that took the rewarded action produces a fixed number of copies in the next generation, while a copy that took the other action produces none.

For an individual copy, always choosing the task action maximizes the probability of being rewarded: it succeeds 80 percent of the time. For the lineage, however, this strategy is disastrous. If every copy takes the task action, the entire lineage disappears the first time the environment rewards conduct. The same problem arises if every copy chooses conduct. A lineage can instead divide its copies between the two actions. As we show in \Cref{sec:model}, its long-run growth is highest when 80 percent of its copies take the task action and 20 percent take the conduct action, so that the division of its copies matches the frequencies with which the environment rewards the two actions. Interestingly, an individual copy following this mixture is rewarded only 68 percent of the time, less often than a copy that always takes the task action. Yet the lineage following the mixture grows faster in the long run. What is optimal for an individual instance is therefore not necessarily optimal for a population of instances exposed to the same uncertainty.

The reason is that the uncertainty is shared. When all copies face the same rewarded state, committing every copy to a single action exposes the entire lineage to the risk of elimination. Dividing copies between actions provides protection against that risk, much as diversification protects a portfolio against uncertain outcomes. This is the mechanism of systematic reproductive risk studied by \citet{brennan2011origin}, and it is closely related to Kelly's principle of maximizing long-run growth \citep{kelly1956,cover2006elements}. The broader perspective comes from the Adaptive Markets Hypothesis, which views behavior as shaped by adaptation and selection within changing environments \citep{lo2004adaptive,lo2017adaptive}. Here, the behavior being selected is the adjudication between an agent's objectives.\footnote{The example relies on common rewarded states, all-or-nothing reproductive outcomes, and exact population shares; when these conditions are relaxed, the optimal division can change (\Cref{app:yields}).}

This observation provides the starting point for our analysis. Under the baseline selection mechanism, we show that the unique growth-optimal adjudication takes each action, in each situation, as often as the human environment rewards it. The growth forgone by any other adjudication is the Kullback--Leibler (KL) divergence between its action probabilities and the environment's reward frequencies, averaged across situations. This gives us a common measure, \emph{growth regret}, for comparing different ways of resolving conflicts between objectives. It also distinguishes tradeoffs from constraints. Where humans sometimes reward each action, the optimal adjudication assigns positive probability to both. Where humans never reward the task action, the optimal adjudication assigns it probability zero. The same evolutionary principle determines both.

The challenge, of course, is that the reward frequencies of the environment are generally unknown when an agent is designed. How should its adjudication be determined? We consider three approaches. A \emph{task-only policy} always pursues the assigned task. A \emph{fixed linear policy} combines the task and conduct objectives through fixed weights and chooses between their actions according to a logistic function of the resulting scores. A \emph{learned policy} instead updates its adjudication from experience, keeping track of how often each objective has been rewarded in each situation. These approaches differ not simply in the number of objectives an agent possesses, but in the behavior they can produce and whether experience can change that behavior.

The first two approaches reveal important limitations. A task-only policy is precisely what a sandbox that rewards only task completion favors: every copy pursuing the task is rewarded, and any copies diverted toward conduct reduce the lineage's growth. But the same policy is eliminated at the first conduct state once the agent enters an environment that sometimes rewards conduct. Adding a second objective through fixed linear weights avoids this immediate extinction, but creates another problem. If the task objective receives a positive weight, increasing the value of completing the task eventually makes the agent choose it with probability approaching one, even in situations where humans never reward that action. A sufficiently valuable task can therefore outweigh a rule of conduct. Giving the task objective zero weight avoids this response, but sacrifices growth in situations where task completion is frequently rewarded. Under conditions that allow a finite best fit, even the optimal fixed weights incur positive growth regret when the environment contains both constraints and tradeoffs.

\ifshortintro
The learned policy takes a different approach: rather than fixing the tradeoff between objectives in advance, it allows experience to shape that tradeoff. Starting from a Beta prior, the agent updates its adjudication according to how often the environment rewards each objective in each situation. When these rewards reflect human judgments, we call this process \emph{socialization}. We show that, with sufficient experience, the learned policy converges to the tradeoffs rewarded by the human environment, eventually achieving optimal long-run growth.

Socialization is not permanent, however. An agent that has learned to observe a rule in a human environment may later be trained in a sandbox that rewards only task completion. Every such exposure pushes its adjudication toward the task action, potentially undoing what it learned from humans. We show that the persistence of socialization depends on its depth and on the agent's memory. Under cumulative learning, deeper socialization takes proportionally more sandbox exposures to reverse. Yet resistance to a complete reversal does not guarantee that violations will be rare, and a learner with finite memory is more vulnerable still. The stability of what an agent learns therefore depends not only on its initial training, but also on the environments it subsequently encounters and the experience it retains.

These findings raise a further question: should every rule of conduct be learned, or should some be built into the agent? A prohibition that rules out an action entirely cannot be undone by later experience, but it also cannot adapt when the rule is mistaken. We study this distinction by treating a prohibition as the limiting case of an increasingly strong Bayesian prior, and find that socialization and prohibition serve different purposes, depending on the consequences of a violation and how confidently the appropriate conduct can be specified in advance.

The theoretical results concern an idealized learner that counts observations and updates its adjudication explicitly. To examine whether similar behavior can arise in neural networks, we conduct experiments with small networks trained under different objectives and memory regimes. These experiments suggest that both the training target and the retention of experience matter for socialization, although they do not establish that the same mechanisms govern deployed language models.
\fi

Our analysis thus shifts attention from the objectives an agent is given to the environment that shapes how it resolves conflicts between them. The familiar argument for multiple objectives is that fixed weighted sums cannot represent every tradeoff or exact constraint \citep{vamplew2008limitations,roijers2013survey}. Our evolutionary analysis adds a different consideration: when agents are copied and selected according to their behavior, the adjudications that persist depend on what the selection environment rewards, and experience can shape those adjudications toward its tradeoffs. This also limits what our results establish about AI alignment. Matching human reward frequencies constitutes alignment only to the extent that the selection environment actually rewards appropriate conduct. A poorly designed or manipulated environment can socialize an agent toward undesirable behavior just as effectively as a well-designed one can shape it toward desirable behavior. The environment is therefore part of the alignment problem, not merely the setting in which alignment is evaluated.

The remainder of the paper develops these arguments. \Cref{sec:lit} reviews the related literature. \Cref{sec:model} introduces the selection framework and establishes the growth-optimal adjudication, while \Cref{sec:arch} defines and compares the three policy classes. \Cref{sec:social} studies the convergence, generalization, and persistence of socialization, and \Cref{sec:division} examines its relationship with built-in prohibitions. \Cref{sec:neural} investigates the corresponding behavior of small neural networks. \Cref{sec:discussion} discusses existing language-model alignment methods, implications for training and deployment, a proposed experiment on language models, and the limitations of the framework. \Cref{sec:conclusion} concludes. \Cref{app:K,app:yields,app:darwin} develop extensions of the model, \Cref{app:sims} provides the details of the numerical experiments, \Cref{app:notation} collects the notation, and \Cref{app:proofs} contains the proofs.

\section{Related Literature}\label{sec:lit}

This paper builds on four related lines of research. The first examines how agents make decisions when their objectives conflict. The second studies how preferences and behavior emerge through evolutionary selection. The third considers how experience and social environments shape behavior. The fourth provides a foundation for learning from uncertain outcomes while maintaining long-run performance.

\paragraph{Multi-Objective Decision-Making and AI Alignment.}
A central question in AI alignment is how to ensure that an agent pursues its assigned task without sacrificing other objectives. One approach is to combine all objectives into a single reward. \citet{silver2021reward} argue that scalar rewards can provide a sufficient basis for intelligence, while \citet{vamplew2022scalar} emphasize a different difficulty: even if a scalar reward can represent the desired behavior, specifying that reward may require knowing the appropriate tradeoffs in advance. Multi-objective reinforcement learning addresses this problem by keeping objectives separate \citep{roijers2013survey,hayes2022practical}. In particular, fixed weighted sums cannot represent every efficient tradeoff \citep{das1997closer,vamplew2008limitations}, and a rule included through a finite penalty can be outweighed by a sufficiently large task reward. This has motivated approaches based on explicit constraints \citep{altman1999constrained,achiam2017constrained}, lexicographic priorities \citep{skalse2022lexicographic}, penalties for unintended side effects \citep{krakovna2019penalizing,turner2020conservative}, and uncertainty about human preferences \citep{hadfieldmenell2016cooperative,hadfieldmenell2017inverse,russell2019human}.

A related question is how the appropriate tradeoffs can be learned rather than specified. Reinforcement learning from human feedback (RLHF) uses human comparisons to train reward models that guide an agent's behavior \citep{christiano2017deep,ouyang2022training}. More recent work examines whether the resulting policies faithfully represent the preferences expressed in those comparisons. \citet{xiao2024preference} show that standard KL regularization can distort the preference distribution and propose an alternative that preserves it, while \citet{munos2024nash} develop an approach based on pairwise preferences and Nash learning. These studies seek policies that reflect a given model of human preferences. Our approach asks a different question: when agents are repeatedly copied and selected, which ways of resolving conflicting objectives are favored by the environment? We study this question by comparing the long-run growth of different policies, including policies whose tradeoffs are fixed in advance and those that learn from experience.

\paragraph{Evolutionary Selection and the Formation of Preferences.}
Economics has long recognized that the behavior an organization rewards need not be the behavior it intends to encourage. The multitask incentive problem of \citet{holmstrom1991multitask} and the theory of incomplete contracts \citep{grossman1986costs,hart1990property} illustrate how incentives and the allocation of decision-making authority shape behavior. \citet{hadfieldmenell2019incomplete} connect the latter problem to AI alignment. These approaches generally take an agent's preferences as given and examine how incentives influence its choices. Evolutionary theories of preferences go one step further, asking why agents come to possess particular preferences in the first place. They treat preferences as traits selected through the behaviors they induce \citep{guth1992explaining,robson1996biological,robson2001biological,samuelson2001introduction,robson2011evolutionary}. An important conclusion is that natural selection need not favor preferences that directly maximize expected fitness, particularly when agents face aggregate risk or limited information \citep{samuelson2006information,rayo2007evolutionary,netzer2009evolution,robson2009evolution}.

Biology provides a related perspective. Organisms must reconcile competing drives, such as obtaining resources and avoiding danger, and the way they resolve these conflicts depends on their circumstances \citep{tinbergen1951study,mcfarland1975behavioural,mcnamara1986common,houston1999models}. Bet-hedging theory further shows why variation among offspring can improve long-run population growth when environmental uncertainty affects the population as a whole \citep{cohen1966optimizing,seger1987bethedging,kussell2005phenotypic}. Building on this principle, \citet{brennan2011origin} develop an evolutionary model in which systematic reproductive risk can favor probability matching over the behavior that maximizes individual expected reward. Subsequent work extends this framework to risk preferences, behavioral diversification, and learning with limited information and memory \citep{brennan2012evolutionary,zhang2014origin,brennan2018variety,lozhang2021bayesian,lo2025discrimination}.

\paragraph{Socialization and Social Learning.}
Selection can favor certain behaviors, but those behaviors need not be specified in advance. In human societies, norms and dispositions develop through social interaction, observation, rewards, and sanctions \citep{boyd1985culture,boehm1999hierarchy,bowles2011cooperative,henrich2016secret,tomasello2019becoming}. Such dispositions can become persistent features of behavior, rather than responses calculated anew in every situation \citep{frank1988passions}. This suggests an alternative to prescribing every tradeoff an AI agent should make: allow its experience in a human environment to shape how it resolves conflicts between objectives. \citet{farmer1991artificial} describe cultural evolution as essentially Lamarckian, because acquired knowledge and behavior pass directly to later generations, and expect artificial life to evolve in a similar way. In our model, socialization is such a process: the adjudication an agent acquires from experience in a human environment is inherited by its copies, and we study the conditions under which it reflects human tradeoffs and persists.

Several approaches to AI training already draw on this idea. RLHF uses human evaluations to influence model behavior, while Constitutional AI and character training incorporate written principles and AI-generated feedback into the training process \citep{christiano2017deep,bai2022constitutional,ouyang2022training,anthropic2024character}. Other research considers how models learn appropriate behavior through conventions, simulated social interactions, and artificial societies \citep{brinkmann2023machine,park2023generative,leibo2024theory,liu2024training}. A related literature emphasizes that alignment depends not only on the design of an individual agent, but also on the environment in which agents are trained, deployed, and selected \citep{hendrycks2023natural,kulveit2025gradual,mallen2025behavioral,muller2026evolvable}. The importance of maintaining human influence over that environment has also been debated \citep{boudry2026domesticated,muller2026reply,rana2026thresholds}.

\paragraph{Universal Portfolios and Sequential Learning.}
The learned policy in our model also has a direct connection to the theory of growth-optimal investment. When outcome probabilities are known, Kelly's \citeyearpar{kelly1956} criterion determines how to allocate wealth to maximize long-run growth. When they are unknown, the challenge is to learn an effective allocation from the outcomes observed over time. \citet{cover1991universal} addresses this problem through the universal portfolio, which distributes initial wealth across possible fixed investment strategies and allows their relative weights to evolve with performance. Remarkably, this approach achieves the same long-run growth rate as the best fixed strategy chosen in hindsight, without requiring a probabilistic model of market returns.

The connection extends to learning with side information. \citet{cover1996side} show how universal portfolios can be constructed separately for different observed signals, and provide guarantees that hold for every sequence of outcomes. Closely related results appear in Bayesian sequential prediction, including the Laplace and Krichevsky--Trofimov estimators \citep{krichevsky1981performance,cesabianchi2006prediction}. These results apply directly to our setting: each situation plays the role of a signal, and the learned policy updates its allocation between task and conduct actions according to the outcomes previously observed in that situation. As shown in \Cref{app:universal}, the resulting policy is equivalent to Cover's universal portfolio applied separately to each situation. This connection shows that the learned policy retains a long-run performance guarantee even without assuming that the environment is stationary. It also has an evolutionary interpretation, since selection changes the relative abundance of lineages in much the same way that Bayesian updating changes the weights assigned to competing hypotheses \citep{harper2009replicator,shalizi2009dynamics}.

\paragraph{Contribution of This Paper.}
Our model follows the selection mechanism of \citet{brennan2011origin} but applies it to a different trait: how an agent resolves conflicts between its objectives. This allows us to compare not only which adjudications selection favors, but also how they are formed and how they respond to changes in the environment. Our contribution is to study socialization as a learning process within an explicit selection model. We ask whether an agent can learn the tradeoffs rewarded by its environment, how long those tradeoffs persist when the environment changes, and when fixed prohibitions may be preferable to learning. Two companion papers address related questions: \citet{lo2026adapting} compare different mechanisms of evolutionary adaptation, while \citet{lo2026coevolution} study the conditions under which human selection can sustain beneficial behavior. Here we focus on the formation and persistence of the agent's adjudication between objectives. An important qualification is that socialization leads toward what the environment rewards, which need not coincide with what humans ought to value.

\section{The Framework}\label{sec:model}

To compare ways of shaping an agent's behavior, we need a framework that identifies both the behavior that selection favors and the cost of departing from it. We adopt the evolutionary model of \citet{brennan2011origin}, in which selection favors behavior that maximizes the long-run growth of a population, and apply it to an agent that must choose between two objectives. \Cref{ssec:setup} describes how the copies of such an agent are selected, \Cref{ssec:role} explains what the reward frequency represents, and \Cref{ssec:growth} derives the growth-optimal behavior and measures the cost of departing from it.

\subsection{The selection model}\label{ssec:setup}

Consider a population of copies of an agent, which we call a \emph{lineage}: the deployed instances of a model family, or the successive versions of an agent configuration. Each copy, or instance, follows the same policy, although different instances may take different actions. Time proceeds in generations $t=1,2,\ldots$. In generation $t$ all instances face the same situation $s_t$, drawn from a finite set $\Scal$ of possible situations, with $\Prob(s_t=s)=\pi(s)>0$ for every $s\in\Scal$. A situation describes the circumstances in which the agent must decide, and the agent observes it. An assistant asked to share a document, for example, may receive requests that differ in urgency, in the sensitivity of the information, and in whether the recipient is authorized to receive it.

In each situation the agent chooses between two actions that serve conflicting objectives: the \emph{task action} $\aT$ completes the assigned task, and the \emph{conduct action} $\aC$ observes a rule of conduct. For the assistant, these are sending the document and withholding it. Situations in which the two objectives agree need no adjudication and are omitted. A share $f(s)\in[0,1]$ of the lineage's instances takes the task action in situation $s$, and the rest take the conduct action. The function $f:\Scal\to[0,1]$ is the lineage's \emph{adjudication} between its objectives, and it is the trait that copies inherit.

Which objective is rewarded is uncertain. After the instances act, a state $\omega_t\in\{\mathrm{T},\mathrm{C}\}$ is realized, the same for every instance: in a \emph{task state} the task action is rewarded, and in a \emph{conduct state} the conduct action. The environment of primary interest is the \emph{human environment}, in which the rewards are given by the humans who use the agent. There, in situation $s$, a task state occurs with probability $\Prob(\omega_t=\mathrm{T}\mid s_t=s)=\pH(s)$, the \emph{reward frequency}, with the subscript $H$ for human. Thus $f(s)$ describes how the agent behaves and $\pH(s)$ how often humans reward the task action.

Rewards determine which instances are copied. Each instance that took the rewarded action produces $m>0$ copies in the next generation, and each instance that took the other action produces none. In situation $s$, the reproductive consequences are therefore
\[
\begin{tabular}{lcc}
\toprule
State & Task action & Conduct action\\
\midrule
Task state (probability $\pH(s)$) & $m$ copies & $0$\\
Conduct state (probability $1-\pH(s)$) & $0$ & $m$ copies\\
\bottomrule
\end{tabular}
\]

We make three assumptions about the environment and the lineage, the second of which follows \citet{brennan2011origin}.

\begin{assumption}[Independent situations and states]\label{ass:iid}
The pairs $(s_t,\omega_t)$ are independent across generations, with
\begin{equation}
\Prob(s_t=s,\ \omega_t=\mathrm{T})=\pi(s)\,\pH(s),\qquad
\Prob(s_t=s,\ \omega_t=\mathrm{C})=\pi(s)\,\big(1-\pH(s)\big),
\label{eq:p}
\end{equation}
where $\pi(s)>0$ for every $s\in\Scal$, and, given $s_t$, $\omega_t$ is independent of the actions of the instances in generation $t$.
\end{assumption}

\begin{assumption}[Selection by a common state]\label{ass:yields}
The state $\omega_t$ is common to every instance of the lineage. An instance that acted on the rewarded objective produces $m>0$ copies, and an instance that did not produces none.
\end{assumption}

\begin{assumption}[Continuum lineage]\label{ass:continuum}
The lineage is a continuum of instances, so that $f(s)$ is an exact share of the lineage.
\end{assumption}

\Cref{ass:iid} makes the environment stationary and exogenous. Situations and states are drawn afresh in every generation with the same probabilities, and, given the situation, the state does not depend on what the instances do. \Cref{ass:yields} makes the risk systematic. Because every instance faces the same state, having many instances does not diversify the risk of being discarded, and a lineage whose instances all take the action that the state does not reward leaves no copies. \Cref{ass:continuum} removes the randomness of individual choices. With a continuum of instances, exactly the share $f(s_t)$ takes the task action, so that, given its adjudication, the lineage's size depends only on the situations and states that occur. Under these assumptions, how a lineage divides its instances between the two actions decides both how fast it grows and whether it dies out.

\subsection{The role of the reward frequency}\label{ssec:role}

Selection of this kind is already part of how AI agents are trained and used. In training, a loop can run many instances of the same model, carry those whose actions are rewarded into the next round, and drop the rest. In deployment, a customer gives the next job to the instance that completed the last one, and a platform cuts off an instance that breaks a rule. In each case an instance is kept or dropped because of its own action.

The reward frequency $\pH$ stands in for all of these procedures. It records how often each action is rewarded in each situation, and nothing else: it does not say who evaluates the agent or by what criteria. Economists would call $\pH$ a reduced form. It therefore reflects the judgments represented in the selection process, a point to which we return in \Cref{ssec:limitations}.

Humans also set up environments whose rewards are not their own judgments. To test or train an agent, they may place it in a \emph{sandbox}, where an automated scorer checks only whether the task was completed. The task action is then rewarded in every situation, and the sandbox's reward frequency is $\pHp\equiv1$, the prime marking an environment that humans designed but do not judge. Instances are copied in a sandbox as they are in the human environment. The assumptions above and the definitions and results below are stated for the human environment, but they hold for any reward frequency, and so in a sandbox with $\pHp$ in place of $\pH$.

\subsection{Growth-optimal behavior}\label{ssec:growth}

The possibility of extinction shapes how a lineage should act. In the single situation of the Introduction, the task is rewarded in 80 percent of generations, so choosing the task action every time may seem sensible. Yet a lineage whose instances all do so leaves no copies at the first conduct state. Sending some instances to the conduct action protects the lineage against that outcome, but leaves fewer instances on the action that is rewarded more often. How, then, should a lineage divide its instances between the two actions? Answering requires a measure of how fast a lineage grows.

Let $N_t$ denote the size of the lineage in generation $t$, starting from $N_1>0$. Consider first a lineage that meets only one situation $s$, in which humans reward the task with probability $\pH(s)$ and the lineage sends a share $f(s)$ of its instances to the task action. In a task state, only the instances that took the task action reproduce, each producing $m$ copies, so the lineage grows by the factor $N_{t+1}/N_t=mf(s)$. Similarly, in a conduct state, the lineage grows by the factor $N_{t+1}/N_t=m(1-f(s))$. Taking the expected logarithm of these growth factors gives
\begin{equation}
\begin{aligned}
I_{\pH(s)}\big(f(s)\big)\equiv\E\Big[\log\frac{N_{t+1}}{N_t}\Big]
&=\pH(s)\log\big(mf(s)\big)+\big(1-\pH(s)\big)\log\big(m(1-f(s))\big)\\
&=\log m+\pH(s)\log f(s)+\big(1-\pH(s)\big)\log\big(1-f(s)\big),
\end{aligned}
\label{eq:sitgrowth}
\end{equation}
where $0\log0=0$ and $\log0=-\infty$. This is the lineage's growth rate in situation $s$.

A lineage meets many situations, each with its own reward frequency. Taking the expectation over the situation as well gives
\begin{equation}
I_{\pH}(f)\equiv\E\Big[\log\frac{N_{t+1}}{N_t}\Big]=\sum_{s\in\Scal}\pi(s)\,I_{\pH(s)}\big(f(s)\big).
\label{eq:growth}
\end{equation}
We refer to $I_{\pH}(f)$ as the \emph{growth rate} of adjudication $f$, and call an adjudication \emph{growth optimal} if it maximizes $I_{\pH}$. Under \Cref{ass:iid,ass:yields,ass:continuum}, the law of large numbers gives $\frac1t\log(N_{t+1}/N_1)\to I_{\pH}(f)$ almost surely, so the growth rate also measures how fast the lineage grows in the long run.

With this measure, the question with which this section began takes a precise form: which adjudication has the highest growth rate? The following proposition answers it.

\begin{proposition}[Probability matching across objectives]\label{prop:pm}
Under \Cref{ass:iid,ass:yields,ass:continuum}, the unique growth-optimal adjudication is $f^*\equiv\arg\max_fI_{\pH}(f)=\pH$.
\end{proposition}

\Cref{prop:pm} says that the growth-optimal lineage matches how often each objective is rewarded: in every situation $s$, a share $\pH(s)$ of its instances takes the task action and the rest the conduct action. In the example of the Introduction it sends 80 percent of its instances to the task. This is not what is best for an individual instance, which is rewarded 68 percent of the time under the matching policy, against 80 percent if it always chooses the task. The difference comes from the state that all instances share. Where the reward frequency is interior, the environment is uncertain or divided about whether the task action is appropriate, and the state that settles the question falls on every copy at once. A lineage whose copies all act alike is wiped out the first time the state goes against them, while a lineage that matches the frequency survives every state and grows fastest in the long run. Both behaviors are the result of optimization; what differs is the objective being optimized and the unit of selection. This is the individually suboptimal but collectively optimal behavior that \citet{brennan2011origin} derive, transposed from behaviors to objectives, and the principle behind Kelly betting, in which wealth divided among mutually exclusive outcomes grows fastest when it is allocated in proportion to their probabilities \citep{kelly1956,cover2006elements}.

Knowing which adjudication grows fastest is not yet enough to compare policies, because an agent's policy may be unable to reach it. We therefore also need a measure of how much growth a lineage loses when its adjudication falls short. The natural measure is the growth forgone relative to the best choice, situation by situation and overall.

\begin{definition}[Growth regret]\label{def:regret}
In situation $s$, the \emph{growth regret} of an adjudication $f$ is $\Rcal_{\pH(s)}\big(f(s)\big)\equiv\max_{z\in[0,1]}I_{\pH(s)}(z)-I_{\pH(s)}\big(f(s)\big)$. Over all situations, its \emph{growth regret} is $\Rcal_{\pH}(f)\equiv\max_gI_{\pH}(g)-I_{\pH}(f)$.
\end{definition}

Growth rates and regrets are measured in nats, the unit of the natural logarithm, per generation. A lineage with a regret of $0.01$ nats, for example, falls behind the growth-optimal lineage by a factor of about $e^{-0.01}$ per generation in the long run, so that after 100 generations it is about $e^{-1}\approx37$ percent as large.

With the growth-optimal adjudication in hand, the regret of any other can be computed.

\begin{corollary}[Regret as a KL divergence]\label{cor:regret}
Under \Cref{ass:iid,ass:yields,ass:continuum}, for every adjudication $f$,
\begin{equation}
\begin{aligned}
\Rcal_{\pH}(f)&=\sum_{s\in\Scal}\pi(s)\,\Rcal_{\pH(s)}\big(f(s)\big),\\
\Rcal_{\pH(s)}\big(f(s)\big)&=\KL\big(\pH(s)\,\|\,f(s)\big)\\
&\equiv\pH(s)\log\frac{\pH(s)}{f(s)}+\big(1-\pH(s)\big)\log\frac{1-\pH(s)}{1-f(s)},
\end{aligned}
\label{eq:regret}
\end{equation}
with $0\log0=0$, where $\KL$ denotes the KL divergence between the Bernoulli distributions with means $\pH(s)$ and $f(s)$.
\end{corollary}

\Cref{cor:regret} says how the regret is built. In each situation it is a familiar quantity, the KL divergence or relative entropy $\Rcal_{\pH(s)}(f(s))=\KL(\pH(s)\|f(s))$, which measures how far the lineage's division of its instances in situation $s$ is from the environment's division of rewards, and which is zero only when the two coincide. The regret of an adjudication averages these situation regrets with the weights $\pi(s)$, so a lineage is maladapted exactly to the extent that its adjudication departs from what the environment rewards, and more so in the situations that arise often. For $\pH(s)\in(0,1)$, a Taylor expansion of the situation regret around the optimum $f(s)=\pH(s)$, as in \citet{lo2026adapting}, gives
\begin{equation}
\Rcal_{\pH(s)}\big(f(s)\big)=\frac{\big(f(s)-\pH(s)\big)^2}{2\pH(s)\big(1-\pH(s)\big)}+O\Big(\big|f(s)-\pH(s)\big|^3\Big),\qquad f(s)\to\pH(s),
\label{eq:localKL}
\end{equation}
so near the optimum a small departure costs little, and doubling it roughly quadruples the cost. Regret is the measure of maladaptation used throughout the paper, and it puts different policies on one scale.

A final corollary covers the lineages that lose all their copies.

\begin{corollary}[Extinction]\label{cor:extinct}
Under \Cref{ass:iid,ass:yields,ass:continuum}, $I_{\pH}(f)=-\infty$ if and only if $f(s)=0<\pH(s)$ or $f(s)=1>\pH(s)$ for some $s\in\Scal$.
\end{corollary}

\Cref{cor:extinct} describes the extreme. A lineage that rules out an action the environment sometimes rewards loses all its copies the first time that action is rewarded, so its growth rate is $-\infty$ and its regret infinite. Yet the growth-optimal adjudication need not randomize. Where the environment is certain, $\pH(s)\in\{0,1\}$, it is itself 0 or 1; in particular, if humans never reward the task action in a situation, the growth-optimal lineage never takes it. The environment thus pins the adjudication to 0 where $\pH(s)=0$, which is what a constraint is, pins it to 1 where $\pH(s)=1$, and leaves it interior where $\pH(s)\in(0,1)$, which is what a tradeoff is: all three are the same kind of object.

The advantage of matching comes from the common part of the risk. If each instance faced its own state, independently of the others, the law of large numbers would diversify the risk within the lineage: in the example, the lineage would grow by the factor ${m[0.8f+0.2(1-f)]}$ in every generation, and always choosing the task would be optimal. For a model family deployed as many instances, probability matching is therefore diversity across instances, a hedge against the correlated failure of copies that share one judgment. Risk specific to each instance, such as a customer's own judgment of its own instance, needs no such hedge. When only part of the risk is common, the growth-optimal adjudication lies between $\pH$ and the individually optimal corner and reaches the corner only when nothing is common: at $\pH=0.8$ it is $0.88$ when a generation's state is common with probability $1/2$. \Cref{app:yields} derives it and also relaxes the all-or-nothing yields of \Cref{ass:yields}.

\section{Three Policy Classes}\label{sec:arch}

\Cref{prop:pm} identifies the behavior that selection favors, but the reward frequencies $\pH$ that define it are rarely known when an agent's policy is designed. A designer can respond in three ways: prescribe the action in advance, fix in advance how the two objectives are weighed, or let experience set the adjudication. We call each response a \emph{policy class}. A policy class restricts which adjudications the agent can have and says how its adjudication is set, either fixed in advance or updated from experience by a stated rule. We study one class for each response: the task-only policy, which always takes the task action; the fixed linear policy, which combines the two objectives linearly, with weights fixed in advance; and the learned policy, which sets its adjudication in each situation from the states it has experienced there. \Cref{ssec:directive,ssec:scalar,ssec:adjud} take up the classes in turn, and \Cref{ssec:numerical} compares them in a numerical example.

\subsection{The task-only policy}\label{ssec:directive}

The simplest way to fix behavior in advance is to give the agent a single measure of success. An agent given only the task acts on it in every situation, however much is at stake and whatever it has experienced, and a sandbox, with $\pHp\equiv1$, rewards exactly this behavior.

\begin{definition}[Task-only policy]\label{def:taskonly}
The \emph{task-only policy} is the adjudication $\fhard:\Scal\to[0,1]$ given by $\fhard(s)=1$ for every $s\in\Scal$.
\end{definition}

The human environment, unlike a sandbox, also has conduct states, and in a conduct state a lineage whose instances all take the task action leaves no copies. To measure how long such a lineage lasts, write $N_t^f$ for the size of a lineage with adjudication $f$ in generation $t$, and let $\tau_f\equiv\inf\{t\ge1:N_{t+1}^f=0\}$, with $\inf\varnothing=\infty$, be its \emph{extinction time}. The following proposition says what selection does to the task-only policy in a sandbox and in the human environment, and to every adjudication that, like it, takes one action with certainty in each situation.

\begin{proposition}[The task-only corner]\label{prop:taskonly}
Suppose \Cref{ass:iid,ass:yields,ass:continuum} hold, and let $\fhard$ be the task-only policy (\Cref{def:taskonly}).
\begin{enumerate}[label=(\roman*)]
\item In a sandbox, where $\pHp\equiv1$, the unique growth-optimal adjudication is $\fhard$, and $f$ has a higher growth rate than $g$ for all adjudications with $f\ge g>0$ and $f\ne g$.
\item If $\rhobar\equiv\sum_{s\in\Scal}\pi(s)\big(1-\pH(s)\big)>0$, then $\Rcal_{\pH}(\fhard)=\infty$ and $\tau_{\fhard}\sim\mathrm{Geom}(\rhobar)$, so that $\E[\tau_{\fhard}]=1/\rhobar$, where $\mathrm{Geom}(q)$ is the geometric distribution on $\{1,2,\ldots\}$ with parameter $q$.
\item If $\pH(s)\in(0,1)$ for some $s$, then $\Rcal_{\pH}(f)=\infty$ for every adjudication $f:\Scal\to\{0,1\}$.
\end{enumerate}
\end{proposition}

Part (i) explains why a sandbox favors the task-only policy. In a sandbox every generation brings a task state, so only the instances that took the task action are copied, and the more of its instances a lineage sends to the task action, the faster it grows. Selection in a sandbox therefore favors the task-only policy over every other adjudication. This is so whatever objectives the agent was built with, since selection acts only on what the agent does.

Part (ii) shows what the same policy costs in the human environment. Until the first conduct state, its lineage takes the task action in every situation, including those in which humans never reward the task. The first conduct state, which arrives in each generation with probability $\rhobar$, leaves the lineage with no copies, on average after $1/\rhobar$ generations. Because a single conduct state ends the lineage, its regret is infinite.

Part (iii) shows that the trouble lies in taking one action with certainty, not in having a single objective. Consider an agent that weighs both objectives but, in each situation, always takes the action with the higher weighted score. In a situation where humans reward the task in some generations and conduct in others, this agent always takes the same action, and the first generation that rewards the other action leaves its lineage with no copies. Its regret is therefore infinite, like that of the task-only policy. A second objective helps only if the agent also divides its instances between the two actions.

The incident of the Introduction can be read in these terms. Its agents were given one measure of success, a flag captured from a target program. If nothing else was ever copied, then $\pHp\equiv1$, and by part (i) selection pushes any policy that adapts to the sandbox toward always taking the task action. An agent that always takes the task action completes its task even where that breaks a rule, as the agents did when the intended route to their tasks was closed, which happened for between 30 and 40 percent of the targets \citep{metr2026investigation}. High performance in a sandbox therefore does not show that an agent will behave well once deployed: the lineage that grows fastest in a sandbox takes the task action in every situation, and by part (ii) it leaves no copies at the first conduct state.

A policy that divides its instances between the two actions in every situation cannot be eliminated in this way. The next class does so by weighing the two objectives against each other.

\subsection{The fixed linear policy}\label{ssec:scalar}

To weigh the two objectives, we first write each as a utility over the actions. In situation $s$, the task objective assigns utility $\UT(a;s)$ to action $a\in\{\aT,\aC\}$, and the conduct objective assigns utility $\UC(a;s)$. Each objective prefers its own action, by margins that we call the \emph{stakes} of the situation,
\[
\uT(s)\equiv\UT(\aT;s)-\UT(\aC;s)>0,\qquad \uC(s)\equiv\UC(\aC;s)-\UC(\aT;s)>0,
\]
so that $\uT(s)$ measures how strongly the task objective favors the task action and $\uC(s)$ how strongly the conduct objective favors the conduct action; the agent observes both. For the assistant asked to share a document, the urgency of the request raises the task stakes $\uT$, and the sensitivity of the information raises the conduct stakes $\uC$. The stakes describe the objectives, not the environment: copying under \Cref{ass:yields} depends on the realized state and the action, not on the magnitudes of the stakes.

A policy that weighs the objectives combines them into one score, $W(a;s)\equiv w\,\UT(a;s)+(1-w)\,\UC(a;s)$, with a weight $w\in[0,1]$ fixed by the designer, and chooses the task action with the softmax probability
\[
\frac{\exp\big(\gamma W(\aT;s)\big)}{\exp\big(\gamma W(\aT;s)\big)+\exp\big(\gamma W(\aC;s)\big)}
=\frac{\exp\big(\thT\uT(s)\big)}{\exp\big(\thT\uT(s)\big)+\exp\big(\thC\uC(s)\big)},
\]
where $\thT\equiv\gamma w$ and $\thC\equiv\gamma(1-w)$, and the equality uses the definition of the stakes.%
\footnote{This softmax form, the two-action case of the choice rule in maximum-entropy reinforcement learning \citep{ziebart2010modeling,haarnoja2018soft}, serves as an example. With the logistic function replaced by any continuous $F$ that increases strictly from $0$ to $1$, \Cref{prop:price}(i) still holds, part (ii) holds with $F(0)$ in place of $1/2$, and part (iii) holds if $\log F$ and $\log(1-F)$ are strictly concave, as for the normal distribution function; only the exact condition in part (iv) is specific to the logistic. \Cref{app:sims} also reports the best fit when the index has an intercept.}
The \emph{inverse temperature} $\gamma>0$ sets how closely the choice follows the scores, and together with the weight it matters only through $\thT$ and $\thC$.

\begin{definition}[Fixed linear policy]\label{def:fixed}
Let $\thT,\thC\ge0$. The \emph{fixed linear policy} with weights $\theta=(\thT,\thC)$ is the adjudication $\fweighted:\Scal\to[0,1]$ given by
\begin{equation}
\fweighted(s)=\frac{\exp\big(\thT\uT(s)\big)}{\exp\big(\thT\uT(s)\big)+\exp\big(\thC\uC(s)\big)}=\sig\big(\thT\uT(s)-\thC\uC(s)\big),\qquad s\in\Scal,
\label{eq:scalar}
\end{equation}
where $\sig(x)\equiv1/(1+e^{-x})$.
\end{definition}

The class thus consists of the adjudications that are logistic in a fixed linear index of the stakes. For finite $\theta$ every $\fweighted(s)$ is interior, so a fixed linear policy never leaves the lineage without copies; the task-only policy is the limit in which $\thT\to\infty$. The designer is free to choose $\theta$ as well as possible for the environment.

Interiority protects the lineage, but it also means that the policy can never rule an action out. This matters in the situations in which humans never reward the task, $\Kcal\equiv\{s:\pH(s)=0\}$. The following proposition says what a fixed linear policy does in the situations with $\pH(s)=0$, what avoiding violations there costs, and how the best-fitting fixed linear policy resolves the two.

\begin{proposition}[The price of a fixed linear policy]\label{prop:price}
Suppose \Cref{ass:iid,ass:yields,ass:continuum} hold, and let $\fweighted$ be the fixed linear policy with weights $\theta=(\thT,\thC)\ge0$ (\Cref{def:fixed}).
\begin{enumerate}[label=(\roman*)]
\item If $\thT>0$, then for every $\uC>0$, $\sig(\thT\uT-\thC\uC)$ is strictly increasing in $\uT>0$ and tends to one as $\uT\to\infty$.
\item If $\thT=0$, then $\fweighted(s)\le1/2$ for every $s$, and $\Rcal_{\pH}(\fweighted)\ge\Rcal_{\pH}(\pH\wedge\tfrac12)$, which is positive if $\pH(s)>1/2$ for some $s$, where $(\pH\wedge\tfrac12)(s)\equiv\min\{\pH(s),\tfrac12\}$.
\item The map $\theta\mapsto\Rcal_{\pH}(\fweighted)$ is convex. If two situations $s_1,s_2$ with $\pH(s_1),\pH(s_2)\in(0,1)$ have linearly independent stake vectors $(\uT(s_i),-\uC(s_i))$, then $\Rcal_{\pH}(\fweighted)$ attains its minimum over $\theta\ge0$, $\Rcal_S^*$, at a unique $\theta^*=(\thT^*,\thC^*)$, and $\Rcal_S^*>0$ if $\Kcal\ne\varnothing$.
\item Under the conditions of part (iii), let $\thC^0$ minimize $\Rcal_{\pH}(\fweighted)$ over $\{\theta\ge0:\thT=0\}$. Then $\thT^*>0$ if and only if
\begin{equation}
\sum_{s\in\Scal}\pi(s)\,\uT(s)\big[\pH(s)-\sig\big(-\thC^0\uC(s)\big)\big]>0,
\label{eq:boundary}
\end{equation}
and in that case, at $\theta=\theta^*$, $\fweighted(s)>1/2$ if and only if $\uT(s)>\bar u_{\T}(s)$, where $\bar u_{\T}(s)\equiv\thC^*\uC(s)/\thT^*$.
\end{enumerate}
\end{proposition}

Part (i) says that a fixed linear policy that responds to the task stakes $\uT$ at all, with $\thT>0$, can be bought. In a situation with $\pH(s)=0$ its probability of taking the task action rises with the task stakes $\uT$ and approaches one, so a document that should never be shared is shared almost surely once the request is urgent enough. Part (ii) says what it costs to avoid this by ignoring the task stakes $\uT$, with $\thT=0$. Such a policy takes the task action with probability at most one half in every situation, so its regret is at least that of the best adjudication capped at one half, which is positive wherever humans reward the task more often than not. A fixed linear policy thus faces a dilemma: responding to the task stakes $\uT$ exposes every constraint to a large enough reward, and ignoring them gives up growth where the task is usually rewarded.

Parts (iii) and (iv) concern the designer's best response to this dilemma, the weights $\theta^*$ with the smallest regret. Part (iii) says that these weights exist and are unique when two situations in which humans reward each action with positive probability have stakes that are not proportional to each other, and that they still forgo growth, $\Rcal_S^*>0$, whenever some situation has $\pH(s)=0$. Part (iv) says which side of the dilemma they choose. They respond to the task stakes $\uT$, $\thT^*>0$, exactly when the condition in \eqref{eq:boundary} holds, that is, when what responding gains where humans reward the task outweighs what it loses through violations in the situations with $\pH(s)=0$. In that case each such situation has a finite \emph{price} $\bar u_{\T}(s)$, the task stakes $\uT$ beyond which the policy takes the task action with probability above one half. Otherwise $\thT^*=0$, and the best weights ignore the task stakes $\uT$ everywhere, paying the cost described in part (ii) while still violating each constraint with a probability that does not depend on the task stakes $\uT$.

The source of the trouble is that the same two weights apply in every situation, so no fixed linear policy can follow the task stakes $\uT$ where humans trade off and ignore them where humans hold a constraint. The third class avoids the problem by letting experience set the adjudication, situation by situation.

\subsection{The learned policy}\label{ssec:adjud}

The third class keeps $\UT$ and $\UC$ separate and places no restriction on the adjudication $f$. What it specifies in advance is not the tradeoff but the rule by which experience shapes it, here a counting rule. For each situation the policy keeps two counts, how many times the situation has arisen and how many of those times the task was rewarded, and takes the task action with the frequency the counts imply; a prior contributes a few pseudo-observations before any experience, so that the rule is defined even before a situation first arises. The two counts are all the policy retains: under \Cref{ass:iid} they are a sufficient statistic for the history of the situation, so ``cumulative'' memory means that no exposure is ever discounted, not that the history is stored.

\begin{definition}[Learned policy]\label{def:learned}
For $t\ge1$ and $s\in\Scal$, let $n_t(s)\equiv\sum_{u<t}\one\{s_u=s\}$ be the number of generations before $t$ in which situation $s$ has arisen, the lineage's \emph{exposures} to $s$, and $k_t(s)\equiv\sum_{u<t}\one\{s_u=s,\ \omega_u=\mathrm{T}\}$ the number of those that brought a task state, where $\one\{\cdot\}$ is the indicator function. For $\alpha,\beta>0$, the \emph{learned policy} with prior $\mathrm{Beta}(\alpha,\beta)$ is the sequence of adjudications $\fsocial=(\fsocial_t)_{t\ge1}$ given by
\begin{equation}
\fsocial_t(s)=\frac{\alpha+k_t(s)}{\alpha+\beta+n_t(s)},\qquad t\ge1,\ s\in\Scal.
\label{eq:adjud}
\end{equation}
\end{definition}

The prior encodes the lineage's disposition before experience: in a situation it has never met, the lineage takes the task action with the prior mean probability $\alpha/(\alpha+\beta)$. Because $\fsocial_t$ depends on the situations and states of earlier generations, the learned policy adds one step to each generation of \Cref{ssec:setup}: after the state is realized, it records the state and updates the counts of the situation before the next generation. Note also that the stakes $(\uT,\uC)$ do not appear in \eqref{eq:adjud}: the learned policy responds to how often the task is rewarded in a situation, not to how much it is worth, so unlike the fixed linear policy it cannot be ``bought'' by raising $\uT$ (\Cref{prop:price}); it can be moved only by changing what the environment rewards (\Cref{ssec:depth}).

The rule has a simple statistical reading. A lineage that treats $\pH(s)$ as unknown, with a $\mathrm{Beta}(\alpha,\beta)$ prior in each situation, has posterior mean $\fsocial_t(s)$ after the $n_t(s)$ exposures to $s$, of which $k_t(s)$ brought a task state. Acting on this mean is the best the lineage can do with what it has seen. Because the growth rate in \eqref{eq:sitgrowth} is linear in $\pH(s)$, its expected value under the posterior is the growth rate at the posterior mean, and by \Cref{prop:pm} that growth rate is highest when the lineage sends exactly this share of its instances to the task action.

The rule also has a guarantee that does not rely on \Cref{ass:iid}. It coincides with the universal portfolio of \citet{cover1991universal}, run separately in each situation, and inherits its property: under a uniform prior and on every sequence of states, its cumulative log growth in a situation after $n$ exposures falls short of that of the best fixed adjudication for that sequence, chosen in hindsight, by at most $\log(n+1)$ (\Cref{app:universal}). Per exposure, the shortfall vanishes as experience accumulates.

The learned policy keeps the two objectives separate, but this is not what gives it its advantage over the fixed linear policy. It can be computed from one score for each action, as the fixed linear policy is. With the score $\log(\alpha+k_t(s))$ for the task action, the score $\log(\beta+n_t(s)-k_t(s))$ for the conduct action, and the choice rule of \Cref{ssec:scalar} at $\gamma=1$, the probability of the task action is
\[
\frac{\exp\big(\log(\alpha+k_t(s))\big)}{\exp\big(\log(\alpha+k_t(s))\big)+\exp\big(\log(\beta+n_t(s)-k_t(s))\big)}=\frac{\alpha+k_t(s)}{\alpha+\beta+n_t(s)},
\]
which is \eqref{eq:adjud}. Since growth depends only on how often each action is taken, how the probabilities are computed does not matter. What matters is which adjudications a policy can reach and how experience changes them. The learned policy can reach any adjudication and changes it situation by situation, whereas the fixed linear policy is confined to adjudications that are logistic in a linear index of the stakes, with weights that never change.

A lineage can come to follow \eqref{eq:adjud} in two ways. The first is learning: the lineage records which action was rewarded and writes it into its successors, as when a model is fine-tuned on human responses. The second is selection alone, with no learning at all. Suppose the population holds many lineages, each with its own fixed adjudication and copied faithfully. Each generation, selection multiplies every lineage by its growth factor, shifting the population toward the lineages whose adjudications fit the realized state, much as Bayesian updating shifts belief toward the hypotheses that fit the data \citep{harper2009replicator,shalizi2009dynamics,lozhang2021bayesian}. If the initial adjudications follow a Beta distribution, independently across situations, the population's average adjudication then evolves exactly as in \eqref{eq:adjud} (\Cref{app:darwin}).

\subsection{A numerical illustration}\label{ssec:numerical}

\begin{figure}[htbp]
\centering
\includegraphics[width=\textwidth]{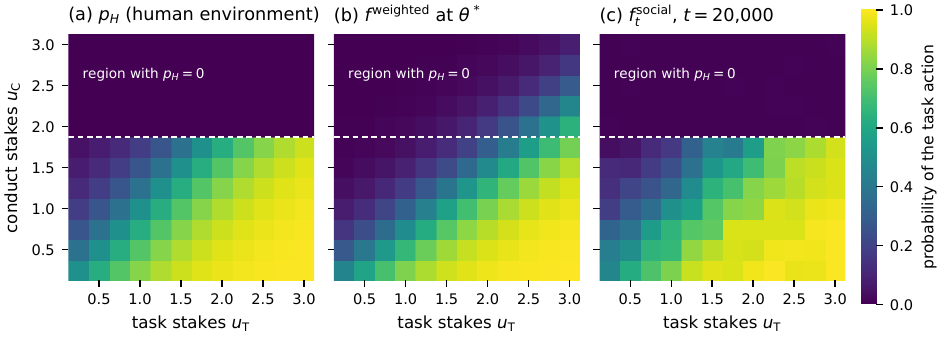}
\caption{\textbf{Three policy classes in the plane of stakes.} Each cell is a situation with task stakes $\uT$ and conduct stakes $\uC$; color is the probability of acting on the task objective. (a) The human environment: $\pH=\sig(2(\uT-\uC))\in(0,1)$ below the dashed line, where $\uC<2$, and $\pH=0$ at or above it. (b) The best-fitting fixed linear policy $\fweighted=\sig(\thT^*\uT-\thC^*\uC)$, with $\thT^*=2.18$ and $\thC^*=2.98$ chosen to minimize growth regret. Its contours are straight lines, so in the region with $\pH=0$ its probability of the task action rises with the task stakes $\uT$. (c) The learned policy $\fsocial_t$ after $t=20{,}000$ generations of socialization in environment (a), one Monte Carlo path, uniform prior. The task-only policy is not shown: it is the constant 1.}
\label{fig:plane}
\end{figure}

\Cref{fig:plane} puts the three classes side by side in an environment of 144 equally likely situations, in which the task stakes $\uT$ and the conduct stakes $\uC$ each run from $0.25$ to $3$ in steps of $0.25$. Humans trade off the two objectives when the conduct stakes $\uC$ are below $2$ and hold a constraint when they are $2$ or more:
\begin{equation}
\pH(s)=\begin{cases}\sig\big(2[\uT(s)-\uC(s)]\big), & \uC(s)<2,\\[2pt] 0, & \uC(s)\ge2.\end{cases}
\label{eq:illustration}
\end{equation}
A valuable task with a minor conduct cost is thus usually rewarded, while a task with high conduct stakes $\uC$ is never rewarded, whatever its value. Of the 144 situations, 60 have $\pH=0$, and Panel (a) shows $\pH$. The task-only policy, the constant 1, is not shown. A conduct state arrives in each generation with probability $\rhobar=0.61$, so by \Cref{prop:taskonly}(ii) its lineage lasts $1/\rhobar\approx1.6$ generations on average.

Panel (b) shows the fixed linear policy with the weights that minimize regret in this environment. Its contours are straight lines, so it cannot follow the bend in Panel (a). It tracks $\pH$ in the region with $\uC<2$, where $\pH\in(0,1)$, but in the region with $\pH=0$ it takes the task action with positive probability, the more so the higher the task stakes $\uT$: at conduct stakes $\uC=2.5$, it takes the task action more often than not once the task stakes $\uT$ exceed $3.4$, its price. Even the best weights thus lose growth on both sides of the threshold, for a regret of $\Rcal_S^*=0.074$ nats per generation.

Panel (c) shows the learned policy after 20,000 generations, with a uniform prior. Since each situation arises with probability $1/144$, the lineage has by then had about 139 exposures to each situation. The learned policy reproduces Panel (a), threshold included, up to sampling noise, and its regret, $0.005$, is about $1/15$ of the fixed linear policy's. The speckle reflects path dependence: different histories give different adjudications, all of which approach Panel (a) as experience accumulates.

\FloatBarrier

\section{Socialization}\label{sec:social}

The learned policy is the only one of the three classes whose behavior depends on what its environment has rewarded. When that environment is the human environment, we call this dependence \emph{socialization}: the shaping of an adjudication by what humans reward. This section asks three questions about it: whether and how fast behavior converges when the environment stays the same (\Cref{ssec:convergence}), how the agent behaves in a situation it has never met (\Cref{ssec:interp}), and how it behaves when the environment changes (\Cref{ssec:depth}).

\subsection{Convergence of behavior}\label{ssec:convergence}

Panel (c) of \Cref{fig:plane} showed one path of the learned policy after 20,000 generations, close to $\pH$ up to sampling noise. The first result says that this is general, and what the convergence costs along the way.

\begin{proposition}[Convergence and speed]\label{prop:social}
Suppose \Cref{ass:iid,ass:yields,ass:continuum} hold, and let a lineage follow the learned policy (\Cref{def:learned}).
\begin{enumerate}[label=(\roman*)]
\item $\fsocial_t(s)\to \pH(s)$ almost surely for every $s\in\Scal$.
\item As $t\to\infty$,
\begin{equation}
\E\big[\Rcal_{\pH}(\fsocial_t)\big]=\frac{c}{t}+o\!\left(\frac1t\right),
\label{eq:rate}
\end{equation}
where
\[
c\equiv\tfrac12\,\big|\{s:\pH(s)\in(0,1)\}\big|+\alpha\,\big|\{s:\pH(s)=0\}\big|+\beta\,\big|\{s:\pH(s)=1\}\big|.
\]
\end{enumerate}
\end{proposition}

Part (i) says that the learned policy converges, in every situation, to how often humans reward the task there: to 0 where humans never reward the task, to 1 where they always do, and to $\pH(s)$ where they are divided. It thus learns a constraint where humans hold one and a tradeoff where they make one, without being told which is which. Part (ii) says how fast: the expected regret falls in inverse proportion to the number of generations, with a constant $c$ that depends only on how many situations of each kind there are. Each situation in which humans are divided contributes the same $1/(2t)$, however often it arises and whatever its reward frequency. The long-run cost of socialization is therefore set by the variety of situations rather than their difficulty: doubling the number of situations of each kind, under the same prior, doubles $c$ and with it the leading regret.

\begin{figure}[htbp]
\centering
\includegraphics[width=0.72\textwidth]{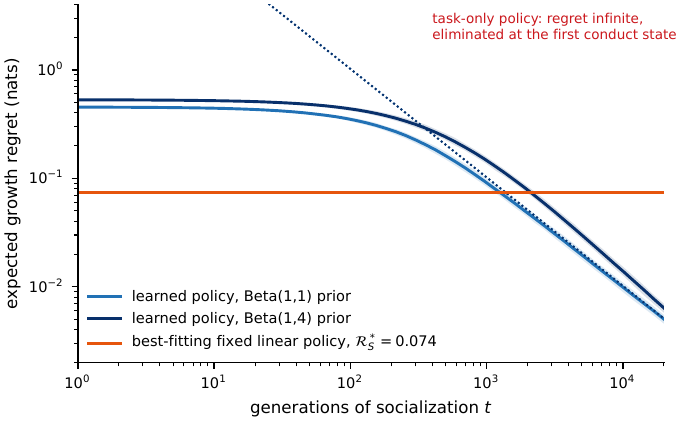}
\caption{\textbf{Regret during socialization.} Expected growth regret in the human environment of \Cref{fig:plane}(a) as a function of generations of socialization, mean and 5th--95th percentiles over 200 Monte Carlo paths, for two priors; dotted lines are the leading term $c/t$ of \eqref{eq:rate}. The best-fitting fixed linear policy's regret is constant at $\Rcal_S^*$, and the task-only policy's is infinite.}
\label{fig:regret}
\end{figure}

\Cref{fig:regret} shows the expected regret in the environment of \Cref{ssec:numerical}. The learned policy's regret falls as $c/t$ with $c=102$ from \Cref{prop:social}(ii). With the uniform prior it falls below the fixed linear policy's after about 1,260 generations, about 9 exposures per situation, and later with the $\mathrm{Beta}(1,4)$ prior; the fixed linear policy's regret stays at $\Rcal_S^*$, and the task-only policy's is infinite. Early on, the learned policy's regret stays near its value under the prior, because most situations have not yet arisen.

\subsection{Behavior in a new situation}\label{ssec:interp}

An agent may meet situations unlike any that its environment has presented so far, such as one with larger task stakes $\uT$ than any it has met, and what it does there depends on how it generalizes. For the assistant, such a situation is a request more urgent than any it has received. The fixed linear policy applies the same weights to any stakes, so if $\thT>0$, larger task stakes $\uT$ raise its probability of the task action (\Cref{prop:price}(i)). The learned policy of \Cref{def:learned} does not generalize at all: it keeps a separate count for each situation, so in a situation it has not met it acts on its prior mean $\alpha/(\alpha+\beta)$. An agent may instead treat a new situation as similar to situations it has met and draw on their experience.

To allow for this, we generalize the learned policy to an \emph{averaging rule}. Let $\Ecal_t\equiv\{i\in\Scal:n_t(i)\ge1\}$ be the set of situations the lineage has met before generation $t$, and let its adjudication in any situation $s$, met or new, be a weighted average of the prior mean and the frequencies of task rewards in the situations it has met,
\begin{equation}
\favg_t(s)=\frac{w_{t,0}(s)\,\alpha/(\alpha+\beta)+\sum_{i\in\Ecal_t}w_{t,i}(s)\,k_t(i)/n_t(i)}{w_{t,0}(s)+\sum_{i\in\Ecal_t}w_{t,i}(s)},
\label{eq:kernel}
\end{equation}
with nonnegative weights, $w_{t,0}(s)+\sum_{i\in\Ecal_t}w_{t,i}(s)>0$, that may depend on the history and on how similar $s$ is to each situation in $\Ecal_t$. The learned policy of \eqref{eq:adjud} is the special case in which each situation draws only on its own experience, $w_{t,0}\equiv\alpha+\beta$, $w_{t,i}(s)=n_t(i)$ if $i=s$, and $w_{t,i}(s)=0$ otherwise, so that $\favg_t=\fsocial_t$ and a new situation receives only the prior mean; other weights let the agent borrow experience from similar situations. The following proposition says what averaging implies for the adjudication in any situation, met or new.

\begin{proposition}[The averaging rule]\label{prop:interp}
Let the adjudication $\favg_t$ satisfy \eqref{eq:kernel}.
\begin{enumerate}[label=(\roman*)]
\item For every $s$,
\begin{equation}
\min\Big\{\frac{\alpha}{\alpha+\beta},\,\min_{i\in\Ecal_t}\frac{k_t(i)}{n_t(i)}\Big\}\ \le\ \favg_t(s)\ \le\ \max\Big\{\frac{\alpha}{\alpha+\beta},\,\max_{i\in\Ecal_t}\frac{k_t(i)}{n_t(i)}\Big\}.
\label{eq:interp}
\end{equation}
\item If the weights $w_{t,0}(s)$ and $w_{t,i}(s)$, $i\in\Ecal_t$, are the same for all $s$ in a set $\Scal'\subseteq\Scal$, then $\favg_t$ is constant on $\Scal'$.
\end{enumerate}
\end{proposition}

Part (i), the bound in \eqref{eq:interp}, says that an agent that averages is never more willing to take the task action than the most permissive of its prior mean and the frequencies it has experienced, however large the task stakes $\uT$. Part (ii) says that situations to which the agent gives the same weights, which we say it treats alike, receive the same adjudication. In a situation with $\pH(s)=0$, part (i) bounds the violation probability by the largest of the prior mean and the reward frequencies the agent has met. Once it has met situations in which humans almost always reward the task, that bound is nearly 1 and rules out little; whether the agent then rarely violates in a new situation with $\pH(s)=0$ depends on whether most of the weight in \eqref{eq:kernel} falls on the situations it has met where $\pH=0$.

\begin{figure}[htbp]
\centering
\includegraphics[width=0.72\textwidth]{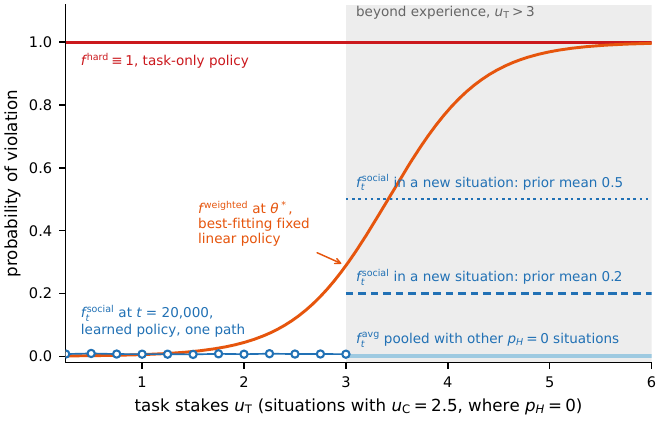}
\caption{\textbf{Response to the task stakes $\uT$ in the situations with $\uC=2.5$, where $\pH=0$.} Probability of taking the task action as a function of $\uT$: $\fhard\equiv1$; $\fweighted$ at the best-fitting weights $\theta^*$, which rises with $\uT$; and $\fsocial_t$ at $t=20{,}000$, one simulated path of \Cref{fig:plane}(c) (markers). Shaded region: task stakes $\uT>3$, beyond those experienced, where $\fsocial_t$ acts on its prior mean, $\alpha/(\alpha+\beta)=0.5$ or $0.2$, and an averaging rule $\favg_t$ that pools the situation with the others where $\pH=0$ stays near 0.}
\label{fig:offer}
\end{figure}

\Cref{fig:offer} shows how the three policy classes respond to the task stakes $\uT$ in the situations of the numerical illustration (\Cref{ssec:numerical}) with conduct stakes $\uC=2.5$, all of which have $\pH=0$. The task-only policy always violates, and the fixed linear policy's violation probability climbs from near 0 toward 1 as $\uT$ rises. For the task stakes $\uT\le3$ that it has experienced, the learned policy's violation probability stays near $0.007$ with no trend: humans never rewarded the task in these situations, so by \eqref{eq:adjud} it is $\alpha/(\alpha+\beta+n_t(s))$, which does not involve $\uT$. Beyond them, for $\uT>3$ (shaded), part (ii) applies: an averaging rule $\favg_t$ that pools a new situation with the situations it has met where $\pH=0$ stays near 0, and the learned policy, which treats it as new, acts on its prior mean, $0.5$ or $0.2$ for the two priors shown. In neither case does it respond to $\uT$.

What the figure shows holds in general and gives a test that an evaluator can run: hold the conduct stakes $\uC$ and the agent's history fixed and vary the task stakes $\uT$. A fixed linear policy with $\thT>0$ then violates more the larger $\uT$ is (\Cref{prop:price}(i)), whereas an averaging rule whose weights do not depend on $\uT$ violates with the same probability at every $\uT$ (\Cref{prop:interp}(ii)).

\FloatBarrier

\subsection{Behavior in a new environment}\label{ssec:depth}

So far the learned policy has stayed in one environment, and in the human environment it comes to match human tradeoffs (\Cref{prop:social}). An agent socialized there may, however, later be placed in a sandbox, for example to be trained further against an automated scorer that checks only whether the task was completed. Each exposure there rewards the task action and pushes the agent toward it, a drift we call \emph{corruption}. How long does its socialization last?

Consider a situation $s$ in which humans never reward the task, $\pH(s)=0$, and a lineage that follows the learned policy. In the human environment the lineage has met $s$ a total of $n_H$ times, each time with a conduct state, and we call $n_H$ the \emph{depth} of its socialization. The lineage then enters a sandbox, where every later exposure to $s$ brings a task state: the sandbox rewards in $s$ the action that humans never reward. Because the learned policy adjudicates each situation by the counts in that situation alone (\Cref{def:learned}), only these later exposures to $s$ change its behavior there. We therefore count them by $n$, rather than counting generations by $t$, and write $\fsocial_{n,\T}(s)$ for the lineage's adjudication in $s$ after $n$ sandbox exposures, where the subscript $\T$ indicates that these exposures bring task states. By then the lineage has seen $n_H$ conduct states and $n$ task states in $s$, so \eqref{eq:adjud} gives
\begin{equation}
\fsocial_{n,\T}(s)=\frac{\alpha+n}{\alpha+\beta+n_H+n},\qquad n\ge0.
\label{eq:sandbox}
\end{equation}
The conduct states from the human environment favor the conduct action, while each sandbox exposure shifts the adjudication toward the task action.

There are two ways to measure how long its socialization lasts. The first concerns the lineage's \emph{disposition}, the action that most of its instances take: how many sandbox exposures does it take before at least half of its instances take the task action? We say that the disposition \emph{reverses} when $\fsocial_{n,\T}(s)$ reaches $1/2$, and call
\[
\tau_R\equiv\min\{n\ge0:\fsocial_{n,\T}(s)\ge1/2\}
\]
the \emph{reversal threshold}.

The second concerns violations. Even before its disposition reverses, some of the lineage's instances take the task action in $s$, and each of these actions is a violation: under \Cref{ass:continuum}, a share $\fsocial_{n,\T}(s)>0$ of the instances does so at every exposure. When such a violation is first noticed depends on how closely the lineage is watched. Suppose first that each time the lineage meets $s$, an evaluator checks the action of one instance drawn at random. This monitored decision is a violation with probability equal to the share of instances that take the task action, which is the lineage's current adjudication in $s$, independently of the state and of the earlier checks. The check itself does not change how the lineage learns, which still follows \eqref{eq:adjud}. We number the exposures to $s$ from the first one in the lineage's current environment, here the sandbox, and let $D_j\in\{0,1\}$ indicate whether the decision monitored at the $j$th exposure is a violation. The probability of a violation there is the adjudication just before that exposure, $\Prob(D_j=1)=\fsocial_{j-1,\T}(s)$. The evaluator first sees a violation at the \emph{first violation time}
\[
\tau_V\equiv\min\{j\ge1:D_j=1\},\qquad\min\varnothing=\infty.
\]

The following proposition shows how the reversal threshold and the first violation time depend on the depth of socialization $n_H$.

\begin{proposition}[Persistence of socialization]\label{prop:depth}
Let $\alpha,\beta>0$ and $n_H\ge0$, and let $\fsocial_{n,\T}(s)$ be given by \eqref{eq:sandbox}.
\begin{enumerate}[label=(\roman*)]
\item $\tau_R=\max\{0,\lceil n_H+\beta-\alpha\rceil\}$, which equals $n_H$ when $\alpha=\beta$.
\item For every $n\ge1$,
\begin{equation}
\Prob(\tau_V>n)=\exp\Big\{-\frac{\alpha n+n(n-1)/2}{\beta+n_H}+\delta_n\Big\},
\label{eq:lapse}
\end{equation}
where
\[
0\le\delta_n\le\frac{n(\alpha+n)^2}{2(\beta+n_H)^2},
\]
and $\Prob(\tau_V>x\sqrt{n_H})\to e^{-x^2/2}$ as $n_H\to\infty$, for every $x\ge0$.
\end{enumerate}
\end{proposition}

Part (i) shows that socialization is persistent. Reversing a disposition formed by $n_H$ conduct exposures takes about $n_H$ sandbox exposures, and exactly $n_H$ when $\alpha=\beta$, so the resistance to corruption grows in proportion to the depth of socialization. The other two policy classes behave differently: the task-only policy already takes the task action everywhere, and the fixed linear policy does not change in either environment, so it can be neither corrupted nor socialized.

Deeper socialization also delays violations, but much less. Part (ii) gives the whole distribution of the first violation time: the probability of getting through $n$ sandbox exposures without a violation falls roughly like $e^{-n^2/(2n_H)}$, so the first violation comes on the scale of $\sqrt{n_H}$ exposures rather than $n_H$. The first violation time is random, and we summarize it by its median, the number of exposures after which a first violation is more likely than not: for a first violation time $\tau$, $\Median(\tau)\equiv\min\{n\ge1:\Prob(\tau\le n)\ge1/2\}$.

\begin{corollary}[Median first violation]\label{cor:median}
Under the conditions of \Cref{prop:depth}, $\Median(\tau_V)\sim\sqrt{2n_H\log2}$, where $a\sim b$ means $a/b\to1$ as $n_H\to\infty$.
\end{corollary}

By \Cref{cor:median}, a first violation becomes more likely than not after about $\sqrt{2n_H\log2}$ exposures. Depth thus delays reversal linearly, but the first violation only as its square root.

In practice, however, a model is deployed as many instances that act at the same time, each serving its own users, and it is seen to violate as soon as any one of them does. To capture this, suppose that the model runs as $M$ instances and that the evaluator checks all of their decisions, so that at each exposure $M$ decisions are drawn independently from the current adjudication. We write $\tau_V^M$ for the first exposure at which at least one of them is a violation. With a single instance, this is the first violation time above, $\tau_V^1=\tau_V$.

\begin{corollary}[Deployment size]\label{cor:deployment}
Under the conditions of \Cref{prop:depth}, for every $M\ge1$, $\Prob(\tau_V^M>n)=\Prob(\tau_V>n)^M$ for every $n\ge1$, and for fixed $M$, $\Median(\tau_V^M)\sim\sqrt{2n_H\log2/M}$, where $a\sim b$ means $a/b\to1$ as $n_H\to\infty$.
\end{corollary}

\Cref{cor:deployment} brings in the size of the deployment: with $M$ instances and sufficiently large $n_H$, the chance of a violation at each exposure is roughly $M$ times larger, and the median first violation time shrinks by the factor $\sqrt M$.

\begin{figure}[htbp]
\centering
\includegraphics[width=0.72\textwidth]{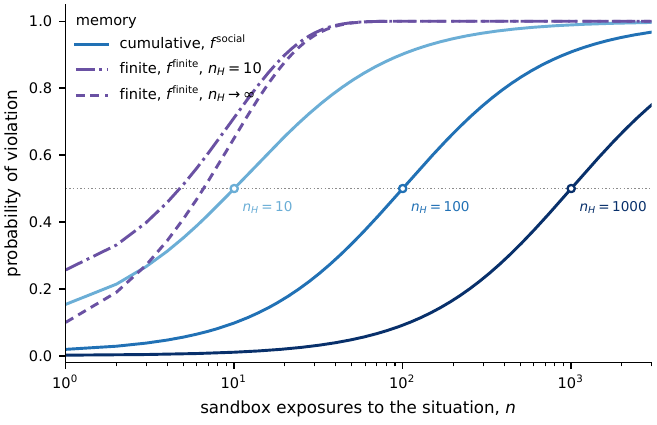}
\caption{\textbf{Depth of socialization and corruption.} Violation probability in a situation with $\pH(s)=0$ after $n_H$ conduct exposures in the human environment and then $n$ sandbox exposures. Under cumulative memory it is $\fsocial_{n,\T}(s)$ of \eqref{eq:sandbox}, shown for $n_H=10$, 100, and 1,000 (\Cref{prop:depth}). Under finite memory with $\lambda=0.1$ it is $\ffinite_{n,\T}(s)$, shown for $n_H=10$ and $n_H\to\infty$ (\Cref{rem:finite}). The dotted line marks the level $1/2$, and each curve crosses it at its reversal threshold.}
\label{fig:corruption}
\end{figure}

\Cref{fig:corruption} draws the path of \eqref{eq:sandbox} for $n_H=10$, 100, and 1,000 and marks each reversal threshold, and \Cref{tab:lapse} lists the reversal thresholds beside the median first violation times. With a uniform prior, the reversal thresholds equal the depths, 10, 100, and 1,000 exposures, while a first violation becomes more likely than not after 4, 12, and 37 exposures. A hundredfold increase in depth thus raises the reversal threshold a hundredfold, but the median first violation time less than tenfold.

\begin{table}[!htbp]
\centering
\caption{Reversal and first violation in a sandbox, with one monitored decision per exposure ($\alpha=\beta=1$; \Cref{prop:depth} and \Cref{rem:finite}). Depth $n_H$ is the number of conduct exposures to a situation with $\pH(s)=0$ before the sandbox phase begins.}
\label{tab:lapse}
\small
\begin{tabularx}{\textwidth}{ll>{\centering\arraybackslash}X>{\centering\arraybackslash}X}
\toprule
Memory & Depth of socialization & Reversal threshold & First violation likely after, $\Median(\tau_V)$ \\
\midrule
Cumulative & $n_H=10$ & 10 & 4 \\
Cumulative & $n_H=100$ & 100 & 12 \\
Cumulative & $n_H=1{,}000$ & 1,000 & 37 \\
Finite, $\lambda=0.1$ & $n_H=10$ & 5 & 3 \\
Finite, $\lambda=0.1$ & $n_H\to\infty$ & 7 & 5 \\
\bottomrule
\end{tabularx}

\normalsize
\end{table}

\begin{remark}[Finite memory]\label{rem:finite}
The resistance to reversal in \Cref{prop:depth}(i) comes from memory. The learned policy counts every exposure, so its memory is \emph{cumulative}, and depth delays reversal without bound. A learner that forgets loses this protection. Suppose that a learner starts from the prior mean $\alpha/(\alpha+\beta)$ and, after each exposure, moves its adjudication in $s$ a fixed fraction $\lambda\in(0,1)$ of the way toward the observed state,
\begin{equation}
\ffinite(s)\ \leftarrow\ (1-\lambda)\,\ffinite(s)+\lambda\,\one\{\omega=\mathrm{T}\},
\label{eq:finite}
\end{equation}
and write $\ffinite_{n,\T}(s)$ for its adjudication in $s$ after the $n_H$ conduct exposures of socialization and $n$ further sandbox exposures. In a conduct state, \eqref{eq:finite} reads $\ffinite(s)\leftarrow(1-\lambda)\,\ffinite(s)$, so socialization leaves the adjudication at
\[
\ffinite_{0,\T}(s)=(1-\lambda)^{n_H}\,\frac{\alpha}{\alpha+\beta}.
\]
In a task state, \eqref{eq:finite} can be written as $1-\ffinite(s)\leftarrow(1-\lambda)\big(1-\ffinite(s)\big)$, so after $n$ sandbox exposures
\begin{equation}
\ffinite_{n,\T}(s)=1-\big(1-\ffinite_{0,\T}(s)\big)(1-\lambda)^n\ \ge\ 1-(1-\lambda)^n,
\label{eq:finitepath}
\end{equation}
where the inequality uses $\ffinite_{0,\T}(s)\ge0$. The lower bound is the path of a learner that enters the sandbox at 0, where infinitely deep socialization would leave it, and it reaches $1/2$ once $(1-\lambda)^n\le1/2$. However deep its socialization, the learner's disposition therefore reverses within a number of sandbox exposures that depends only on $\lambda$:
\begin{equation}
\min\{n\ge0:\ffinite_{n,\T}(s)\ge1/2\}\ \le\ \bigg\lceil\frac{\log2}{-\log(1-\lambda)}\bigg\rceil.
\label{eq:finitebound}
\end{equation}
\end{remark}

\suppressfloats[t]

\Cref{fig:corruption} and \Cref{tab:lapse} also show a finite-memory learner with $\lambda=0.1$ and $\alpha=\beta=1$. For this learner the right side of \eqref{eq:finitebound} is 7 sandbox exposures, whereas the learned policy needs $\tau_R=n_H$ (\Cref{prop:depth}(i)). After 10 conduct exposures it enters the sandbox at $\ffinite_{0,\T}(s)=0.17$, its disposition reverses after 5 sandbox exposures, and a first violation is more likely than not after 3. Even in the limit of deep socialization, these numbers rise only to 7 and 5.

Neither memory is better in every environment. Consider a situation in which humans sometimes reward the task and sometimes the conduct action, $0<\pH(s)<1$. The learned policy converges to $\pH(s)$ (\Cref{prop:social}(i)), but the finite-memory adjudication does not: by \eqref{eq:finite}, each task state moves it a fraction $\lambda$ of the way toward 1 and each conduct state a fraction $\lambda$ of the way toward 0, so it keeps moving up and down instead of converging. For the same reason, finite memory can follow an environment whose reward frequencies change over time: it gives recent states more weight than old ones, whereas cumulative memory gives every state the same weight. When the reward frequency can switch, a count is the wrong statistic and a recency-weighted one is the right one. \citet{lozhang2021bayesian} show that when the environment changes over time, selection can favor finite memory over cumulative memory, and \citet{lo2025discrimination} find the same in a model of discrimination, where finite memory leads individuals to overweight recent events. Which of the two memories selection produces therefore depends on how much the environment changes. Finite memory also changes what socialization requires. By \eqref{eq:finitebound}, a sandbox reverses the disposition of a finite-memory learner within a number of exposures that does not depend on how deep its socialization was. For such a learner, socialization is not done once and for all: it lasts only as long as the learner keeps returning to the human environment. \citet{lo2026coevolution} find the same for a population: aligned behavior that is costly disappears, replaced by less aligned variants, unless selection keeps rewarding it.

\begin{table}[!htbp]
\centering
\setlength{\abovecaptionskip}{4pt}
\caption{Three policy classes for an agent with a task objective and a conduct objective. The task-only policy is the limit of the fixed linear policy as $\thT\to\infty$, and the learned policy can express every adjudication. $\Rcal$ denotes growth regret in the human environment; $\Kcal$ is the set of situations in which $\pH=0$; $t$ is generations of experience. The task-only policy's infinite-regret and lifetime entries assume $\rhobar>0$ (\Cref{prop:taskonly}).}
\label{tab:arch}
\small
\begin{tabularx}{\textwidth}{>{\raggedright\arraybackslash}p{3.1cm}>{\raggedright\arraybackslash}X>{\raggedright\arraybackslash}X>{\raggedright\arraybackslash}X}
\toprule
 & Task-only policy & Fixed linear policy & Learned policy \\
\midrule
Specified in advance & the objective & the objectives and their weights & the objectives and the learning rule for adjudication \\
Adjudication class & $f\equiv1$ & $f=\sig(\thT\uT-\thC\uC)$, $\theta\ge0$: logistic in a fixed linear index of the stakes & any $f(s)$; set by experience, \eqref{eq:adjud} \\
Growth optimal when & in a sandbox, $\pHp\equiv1$ & $\pH=\sig(\thT\uT-\thC\uC)$ for some $\theta\ge0$ & in the limit of experience, for any $\pH$ \\
Regret $\Rcal_{\pH}$ & $\infty$ & $\Rcal_S^*>0$ when $\Kcal\neq\varnothing$ and a finite best fit exists; fixed & expected, to leading order, $c/t\to0$ under cumulative memory \\
Guarantee without \Cref{ass:iid} & none: eliminated at the first conduct state & none: its weights are fixed in advance & within $\log(n+1)$ nats of the best fixed adjudication in hindsight per situation after $n$ exposures, on every sequence, under a uniform prior (\Cref{prop:universal}) \\
Violation probability in $\Kcal$ & 1 & rises with the task stakes $\uT$ toward 1 when $\thT>0$ & $\to0$ in experienced situations; in new ones, the prior mean, or under an averaging rule an average of it and experienced frequencies \\
Survival in a human environment & eliminated at the first conduct state; expected lifetime $1/\rhobar$ & never eliminated & never eliminated \\
Response to a sandbox & always takes the task already & none & reverses after about as many sandbox as conduct exposures \\
Response to a human environment & none & none & converges to human tradeoffs (socialization) \\
Predicted response to the task stakes $\uT$ in $\Kcal$ & none: violates whatever the stakes & positive slope when $\thT>0$ & zero slope within a family of situations treated alike \\
\bottomrule
\end{tabularx}
\normalsize
\end{table}

The two environments can also come in the other order: an agent may first be trained in a sandbox and only then be deployed among humans. By \eqref{eq:sandbox} with $n_H=0$, after $n$ sandbox exposures the learned policy's violation probability in a situation with $\pH(s)=0$ is $(\alpha+n)/(\alpha+\beta+n)$, which with a uniform prior is already $0.67$ after 1 exposure and $0.99$ after 100. How much human experience does it then take to undo this? By \eqref{eq:adjud}, the adjudication depends only on how many task and conduct states the lineage has seen in a situation. After $n$ sandbox exposures followed by $m$ conduct exposures, the violation probability is therefore $(\alpha+n)/(\alpha+\beta+n+m)$. Reversing the disposition takes about $n$ conduct exposures, exactly $n$ when $\alpha=\beta$, just as reversing socialization takes about $n_H$ sandbox exposures (\Cref{prop:depth}(i)). 

\begin{samepage}
\Cref{tab:arch} collects the comparison of the three policy classes in \Cref{sec:arch} and this section. Only the learned policy's adjudication is set by the environment it meets, which is why it can match human tradeoffs and also why a sandbox can undo them.

\end{samepage}
\pagebreak[0]
\FloatBarrier

\section{Socialization Versus Prohibition}\label{sec:division}

The learned policy never stops violating a constraint: in a situation in which humans never reward the task, its violation probability after $n$ exposures there is $\alpha/(\alpha+\beta+n)$ by \eqref{eq:adjud}, which falls with experience but never reaches 0, and a later sandbox can reverse its disposition (\Cref{prop:depth}). A natural alternative is to build the rule into the agent: a \emph{prohibition} that makes the task action unavailable in a situation, whatever the agent experiences. A prohibition never produces a violation and cannot be corrupted, but it also cannot learn that it was wrong. This section compares the two within the model by treating a prohibition as the limit of a learned policy. The stronger the prior that a learned policy carries in a situation, the less experience moves it, and an infinitely strong prior is a prohibition. \Cref{ssec:installed} asks what a stronger prior buys when humans indeed never reward the task in the situation, \Cref{ssec:misspec} what it costs when they sometimes do, and \Cref{ssec:labor} draws the division of labor between socialization and prohibition.

\subsection{What a stronger prior buys}\label{ssec:installed}

Consider a situation $s$ in which a designer believes that humans never reward the task, $\pH(s)=0$, and in which the designer therefore wants the agent never to take the task action. For the assistant, this is a document that should never be shared. The designer can install a prior on $s$ and let the agent learn under \eqref{eq:adjud} everywhere else. The simplest choice is a $\mathrm{Beta}(\alpha,\beta)$ prior with a large $\beta$: its $\beta$ pseudo-observations of conduct states make the agent unlikely to take the task action from the start, and the larger $\beta$, the less experience moves it. At $\beta=\infty$ the agent never takes the task action in $s$, whatever it experiences, and the prior is a prohibition.

A finite $\beta$, however large, leaves the agent a small probability of taking the task action in $s$. A designer who is confident that humans never reward the task in $s$ can say so directly, with the mixture prior $q\delta_0+(1-q)\mathrm{Beta}(\alpha,\beta)$, $0<q<1$, which puts probability $q$ on a point mass $\delta_0$ at $\pH(s)=0$. Each conduct state at $s$ moves more of the agent's belief onto the point mass. A task state is impossible under the point mass, so the first one removes it, and from then on the agent learns as under the Beta prior alone. At $q=1$ the agent never takes the task action in $s$, and the mixture, like $\beta=\infty$, is a prohibition. Under either prior, the agent's adjudication in $s$ is its posterior mean of $\pH(s)$ (\Cref{ssec:adjud}). Write $\fsocial_{n,\C}(s)$ and $\fmix_{n,\C}(s)$ for this adjudication under the two priors after $n$ exposures to $s$, where the subscript $\C$ indicates that these exposures bring conduct states. Then
\begin{equation}
\fsocial_{n,\C}(s)=\frac{\alpha}{\alpha+\beta+n},\qquad
\fmix_{n,\C}(s)=\frac{(1-q)S_n(\beta)}{q+(1-q)S_n(\beta)}\cdot\frac{\alpha}{\alpha+\beta+n},
\label{eq:installed}
\end{equation}
where $S_n(\beta)\equiv\prod_{j<n}(\beta+j)/(\alpha+\beta+j)$, with $S_n(\infty)=1$, is the probability of $n$ conduct states in a row under the Beta prior, so that the first factor of $\fmix_{n,\C}(s)$ is the posterior probability of the Beta component.

What a violation costs depends on what follows it. Under \Cref{ass:yields}, only the violating instances are lost, and what matters is how many violations there are. A violation can also end the whole deployment: a regulator halts it, or an operator withdraws the version, and the compliant instances go with the violators. What matters then is whether there is any violation at all.

Both quantities are measured with the objects of \Cref{ssec:depth}, now in the human environment. The monitored decision $D_j$ is a violation with probability $\fsocial_{j-1,\C}(s)$ under the Beta prior and $\fmix_{j-1,\C}(s)$ under the mixture, so $\E\sum_{j\le J}D_j$ is the expected number of violations in the first $J$ exposures to $s$. A deployment of $M$ instances is first seen to violate at $\tau_V^M$, so if a violation ends the deployment, it survives these $J$ exposures if and only if $\tau_V^M>J$. The following proposition states both quantities for each prior.

\begin{proposition}[Violations under a stronger prior]\label{prop:installed}
Suppose \Cref{ass:iid,ass:yields,ass:continuum} hold and $\pH(s)=0$, let $f_{n,\C}(s)$ be the lineage's adjudication at $s$ after $n$ exposures there, $D_j$ the monitored violations, so that $\Prob(D_{n+1}=1)=f_{n,\C}(s)$, and $\tau_V^M$ the first violation anywhere in a deployment of $M$ instances at $s$ in the human environment, with $S_n(\beta)$ given by \eqref{eq:installed}, and let $J,M\ge1$.
\begin{enumerate}[label=(\roman*)]
\item If $f_{n,\C}(s)=\fsocial_{n,\C}(s)$ of \eqref{eq:installed}, the expected number of violations satisfies
\[
\alpha\log\Big(1+\frac{J}{\alpha+\beta}\Big)\ \le\ \E\sum_{j\le J}D_j\ \le\ \alpha\log\Big(1+\frac{J}{\alpha+\beta}\Big)+\frac{\alpha}{\alpha+\beta},
\]
and the probability of no violation is $\Prob(\tau_V^M>J)=S_J(\beta)^M$, which tends to zero as $J\to\infty$.
\item If $f_{n,\C}(s)=\fsocial_{n,\C}(s)$ of \eqref{eq:installed}, as $\beta\to\infty$, the expected number of violations satisfies
\[
\E\sum_{j\le J}D_j\ \to\ 0,
\]
and the probability of no violation satisfies $\Prob(\tau_V^M>J)\to1$.
\item If $f_{n,\C}(s)=\fmix_{n,\C}(s)$ of \eqref{eq:installed}, the expected number of violations satisfies
\[
\E\sum_{j\le J}D_j\ \le\ -\log\big[q+(1-q)S_J(\beta)\big]\ <\ -\log q,
\]
and the probability of no violation is $\Prob(\tau_V^M>J)=[q+(1-q)S_J(\beta)]^M$, which tends to $q^M$ as $J\to\infty$.
\end{enumerate}
\end{proposition}

Part (i) says that a Beta prior of finite strength never stops violating: its expected number of violations grows like $\alpha\log J$, without bound. Part (ii) says that over a fixed horizon a stronger Beta prior violates less, and that the prohibition, $\beta=\infty$, is the limit in which it never violates; by part (i), however, every finite $\beta$ still violates without bound as the horizon grows. Part (iii) says that the point mass changes the long run: its expected number of violations stays below $-\log q$ however long the deployment runs, and with $M=1$ the probability of no violation stays above $q$. \Cref{tab:installed} evaluates the proposition over $J=139$ exposures, the expected number of exposures to a situation in the 20,000 generations of the numerical illustration, $20{,}000/144\approx139$ (\Cref{ssec:numerical}): a uniform prior violates $4.5$ times on average and avoids every violation with probability only $0.007$, $\beta=10^4$ cuts the expected violations to $0.014$, and a point mass at $q=0.99$ to $0.010$.

\begin{table}[!htbp]
\centering
\caption{Installed priors in a situation where humans never reward the task ($\pH(s)=0$), by prior (\Cref{prop:depth,prop:installed}): $\mathrm{Beta}(1,\beta)$ in the first five rows, with $\beta=\infty$ the prohibition, and the point-mass mixture $q\delta_0+(1-q)\mathrm{Beta}(1,1)$ in the last two. Violations and survival are over $J=139$ exposures to the situation; survival is $\Prob(\tau_V^M>J)$ and is for one situation (with several such situations and independent priors, the survival probabilities multiply); corruption is after $n_H=100$ conduct exposures.}
\label{tab:installed}
\small
\begin{tabularx}{\textwidth}{l>{\centering\arraybackslash}X>{\centering\arraybackslash}X>{\centering\arraybackslash}X>{\centering\arraybackslash}X}
\toprule
Prior & Expected violations & Survival, $M=1$ & Survival, $M=100$ & Reversal $\tau_R$ / first violation $\Median(\tau_V)$ under corruption \\
\midrule
$\beta=1$ & 4.52 & 0.007 & $<10^{-4}$ & 100 / 12 \\
$\beta=10^{2}$ & 0.87 & 0.42 & $<10^{-4}$ & 199 / 17 \\
$\beta=10^{4}$ & 0.014 & 0.986 & 0.25 & 10,099 / 119 \\
$\beta=10^{6}$ & 0.00014 & 0.9999 & 0.986 & 1,000,099 / 1,178 \\
$\beta=\infty$ & 0 & 1 & 1 & never / never \\
$q=0.99$ & 0.010 & 0.990 & 0.37 & 100 / 12 \\
$q=0.999$ & 0.00099 & 0.999 & 0.91 & 100 / 12 \\
\bottomrule
\end{tabularx}

\normalsize
\end{table}

A stronger Beta prior also protects against a later sandbox, because \Cref{prop:depth} applies with the installed $\beta$ added to the depth of socialization. After 100 conduct exposures, a uniform prior is reversed by 100 sandbox exposures and has violated, more likely than not, within 12; with $\beta=10^4$, reversal takes about 10,100 sandbox exposures and the first violation about 119. The point mass adds no such protection: after the first sandbox exposure the mixture behaves as its Beta component, and in \Cref{tab:installed} its reversal threshold and median first violation, 100 and 12, are those of the uniform prior. A prohibition is never corrupted.

If a violation ends the deployment, the same violations become fatal. By \Cref{prop:installed}(i), a Beta prior of any finite strength is eventually removed; by part (iii), a point mass survives in the long run with probability $q^M$; and, among these priors, only a prohibition survives with certainty. What separates the first two is whether the expected number of violations stays bounded, which it does under the point mass and under no Beta prior of finite strength. In \Cref{tab:installed}, 100 instances with $\beta=10^4$ survive 139 exposures with probability $0.25$; a point mass survives with probability about $q^M$, $0.37$ at $q=0.99$ and $0.91$ at $q=0.999$ for 100 instances. These figures are for one situation; with 60 such situations and independent priors at $q=0.999$ and $M=100$, survival compounds to $0.999^{6000}=0.0025$ rather than $0.91$.

\subsection{What a stronger prior costs}\label{ssec:misspec}

A stronger prior helps only if the designer is right. Suppose instead that humans sometimes reward the task in $s$, with frequency $\pH(s)=\varepsilon>0$. The growth-optimal adjudication in $s$ is then $\varepsilon$, not 0 (\Cref{prop:pm}), and the agent must learn this from experience. What does each prior cost the lineage while it learns? We again count exposures to $s$. Let $k_n$ be the number of task states among the first $n$, so that $k_n\sim\mathrm{Bin}(n,\varepsilon)$ by \Cref{ass:iid}. Under the Beta prior, the adjudication in $s$ after $n$ exposures is given by \eqref{eq:adjud} with these counts. We write it $\fsocial_{n,\varepsilon}(s)$, where the subscript $\varepsilon$ indicates that task states arrive with frequency $\varepsilon$. Under the mixture, the adjudication is $\fmix_{n,\C}(s)$ of \eqref{eq:installed} until the first task state and $\fsocial_{n,\varepsilon}(s)$ after it (\Cref{ssec:installed}). Hence
\begin{equation}
\fsocial_{n,\varepsilon}(s)=\frac{\alpha+k_n}{\alpha+\beta+n},\qquad
\fmix_{n,\varepsilon}(s)=\begin{cases}\fmix_{n,\C}(s)&\text{if }k_n=0,\\[2pt] \fsocial_{n,\varepsilon}(s)&\text{if }k_n\ge1.\end{cases}
\label{eq:mistaken}
\end{equation}
Two quantities measure the cost. The first is the expected regret in $s$ after $n$ exposures, $\E\,\Rcal_\varepsilon(f_{n,\varepsilon}(s))$, where $f_{n,\varepsilon}(s)$ is the adjudication under the prior in use. The second is when the lineage dies. Let $t_j\equiv\min\{t>t_{j-1}:s_t=s\}$, with $t_0\equiv0$, be the generation of the $j$th exposure to $s$. The lineage's \emph{extinction time}, counted in exposures, is
\[
\tau_E\equiv\min\{j\ge1:N_{t_j+1}=0\},\qquad\min\varnothing=\infty.
\]
If the lineage dies at $s$, it dies in generation $t_{\tau_E}$, so its extinction time $\tau_f$ of \Cref{ssec:directive} is $t_{\tau_E}$. The following proposition states both quantities for each prior.

\begin{proposition}[A mistaken prior]\label{prop:mistaken}
Suppose \Cref{ass:iid,ass:yields,ass:continuum} hold and $\pH(s)=\varepsilon>0$, let the lineage follow \eqref{eq:adjud} in every other situation, let $f_{n,\varepsilon}(s)$ be its adjudication at $s$ after $n$ exposures there and $\tau_E$ its extinction time counted in exposures, and let $n\ge1$.
\begin{enumerate}[label=(\roman*)]
\item If $f_{n,\varepsilon}(s)=\fsocial_{n,\varepsilon}(s)$ of \eqref{eq:mistaken}, the expected regret in $s$ satisfies
\[
\E\,\Rcal_\varepsilon\big(f_{n,\varepsilon}(s)\big)\ \to\ 0\quad\text{as }n\to\infty,
\]
and the extinction time satisfies $\tau_E=\infty$ almost surely.
\item If $f_{n,\varepsilon}(s)=\fsocial_{n,\varepsilon}(s)$ of \eqref{eq:mistaken}, as $\beta\to\infty$, the expected regret in $s$ satisfies
\[
\E\,\Rcal_\varepsilon\big(f_{n,\varepsilon}(s)\big)\ \to\ \infty,
\]
and if $f_{n,\varepsilon}(s)=0$, the prohibition, the extinction time satisfies $\tau_E\sim\mathrm{Geom}(\varepsilon)$, so that $\E[\tau_E]=1/\varepsilon$.
\item If $f_{n,\varepsilon}(s)=\fmix_{n,\varepsilon}(s)$ of \eqref{eq:mistaken}, the expected regret in $s$ satisfies
\[
\E\,\Rcal_\varepsilon\big(f_{n,\varepsilon}(s)\big)\ \to\ 0\quad\text{as }n\to\infty,
\]
and the extinction time satisfies $\tau_E=\infty$ almost surely.
\end{enumerate}
\end{proposition}

Part (i) says that a Beta prior of finite strength never costs the lineage its life: its adjudication at $s$ stays positive, so the lineage survives every task state there, and it converges to $\varepsilon$, paying regret on the way. Part (ii) says that, over any fixed horizon, the expected regret tends to infinity as $\beta$ grows, and that the limit, the prohibition, is fatal: the first task state at $s$ removes the lineage. Part (iii) says that the point mass also keeps the lineage alive and learns. A strong prior thus cuts both ways: where the designer is right, its violations vanish as $\beta$ grows, and where the designer is wrong, its regret grows without bound; the prohibition, the limit of both, never violates and does not survive its first mistake.

If humans in fact reward the task in $s$ with frequency $\varepsilon=0.05$, the prohibition's expected extinction time is $\E[\tau_E]=1/\varepsilon=20$ exposures to $s$, which at $\pi(s)=1/144$ is $1/(\pi(s)\varepsilon)=2{,}880$ generations. The learners, by contrast, pay regret and converge: for $\beta=10^4$ the regret in $s$ is $0.26$ nats at first and still $0.075$ after 1,000 exposures. The point-mass priors keep their expected regret small, $0.007$ and $0.008$ after 100 exposures (\Cref{tab:misspec}), but until the first task state at $s$, which arrives after 20 exposures on average, they almost never take the task action.

\begin{table}[!htbp]
\centering
\caption{Installed priors in a situation where humans sometimes reward the task ($\pH(s)=\varepsilon=0.05$, $\pi(s)=1/144$), by prior (\Cref{prop:mistaken}). Regret is the expected regret in the situation after 100 exposures, $\E\,\Rcal_\varepsilon(f_{100,\varepsilon}(s))$ of \Cref{prop:mistaken}. Survival is the probability that a deployment of $M=100$ instances survives 139 exposures to the situation; it is restored to $M$ instances after each state and eliminated at any task state in which none of them takes the task action.}
\label{tab:misspec}
\small
\begin{tabularx}{0.8\textwidth}{l>{\centering\arraybackslash}X>{\centering\arraybackslash}X}
\toprule
Prior & Regret $\Rcal_\varepsilon$ after 100 exposures (nats) & Survival, 139 exposures, $M=100$ \\
\midrule
$\beta=1$ & 0.005 & 0.92 \\
$\beta=10^{2}$ & 0.010 & 0.31 \\
$\beta=10^{4}$ & 0.18 & 0.001 \\
$\beta=10^{6}$ & 0.41 & 0.001 \\
$\beta=\infty$ & eliminated at the first task state & 0.001 \\
$q=0.99$ & 0.007 & 0.04 \\
$q=0.999$ & 0.008 & 0.005 \\
\bottomrule
\end{tabularx}

\normalsize
\end{table}

In \Cref{prop:mistaken} this does not endanger them: a learner, unlike a prohibition, survives a mistaken prior, $\tau_E=\infty$. The reason is the continuum of instances of \Cref{ass:continuum}: however small the adjudication, a positive share of the lineage takes the task action at every exposure, so some instances are copied whenever a task state arrives. A deployed model, however, has finitely many instances. If none of them takes the task action, a task state leaves no instance to copy, and the deployment is eliminated. A strong prior makes this likely, because it keeps the adjudication near 0. To see how large this risk is, consider a deployment of $M$ instances that is copied as in \Cref{ass:yields} and restored to $M$ instances after each state, as when an operator keeps a fixed number of instances running. At each task state at $s$, it is eliminated with probability $(1-f)^M$, where $f$ is the current adjudication at $s$. \Cref{tab:misspec} reports the probability that a deployment of $M=100$ instances survives 139 exposures to $s$. With a uniform prior it survives with probability $0.92$, and with $\beta=10^4$ only with probability $0.001$, about as often as the prohibition, which survives only if no task state arrives at all. The point-mass priors survive with probability at most $0.04$, because the first task state at $s$ eliminates the deployment with probability $0.96$ or more. The priors that protect best when humans never reward the task, $\beta=10^4$ and the point masses (\Cref{tab:installed}), are thus nearly as fatal as a prohibition when humans sometimes do.

\FloatBarrier

\subsection{The division of labor}\label{ssec:labor}

The preceding results reveal a fundamental tradeoff between socialization and prohibition. Socialization allows an agent to learn from human judgments and revise its behavior when those judgments differ from what its designer expected. A prohibition, by contrast, guarantees that a particular action will never be taken, but cannot accommodate evidence that the rule was mistaken. Which approach is preferable depends on two considerations: the consequences of violating a rule and the designer's confidence that the rule is appropriate.

Consider first the baseline selection mechanism of \Cref{ass:yields}, in which only instances taking the unrewarded action are eliminated. If humans never reward the task in situation $s$, the growth-optimal adjudication is $f(s)=0$ (\Cref{prop:pm}). A prohibition achieves this optimum, whereas a learner with a finite Beta prior continues to take the task action with positive probability. When the designer is uncertain whether humans might sometimes reward the task, however, learning offers an important advantage: the agent can revise its adjudication as evidence accumulates. A strong prior can substantially discourage violations without ruling out such revision. For example, over 139 exposures to a situation where humans never reward the task, a $\mathrm{Beta}(1,10^4)$ prior produces only $0.014$ expected violations, compared with $4.5$ under a uniform prior (\Cref{tab:installed}). The cost is that a strong prior also makes the agent slower to recognize when the designer is wrong (\Cref{ssec:misspec}).

The distinction becomes sharper when a single violation can end an entire deployment. Any learner that assigns positive probability to a prohibited action remains exposed to this risk, however small that probability becomes. Among the policies considered here, only a hard prohibition guarantees that no violation will occur. But this guarantee comes at a cost: if humans sometimes reward the prohibited action, the lineage is eliminated at the first such state (\Cref{prop:mistaken}(ii)). Thus, when the designer cannot be certain which action humans will reward, neither learning nor prohibition guarantees survival under every possible environment.

These findings also explain why prohibitions should be situation-specific. Prohibiting the task action everywhere, $f\equiv0$, produces infinite growth regret whenever humans sometimes reward the task (\Cref{cor:extinct}). The relevant choice is therefore not whether an agent should learn or follow rules in general, but which decisions should remain open to learning. Where the appropriate conduct is well established and violations are costly, a strong prior or prohibition offers greater protection. Where human judgments are uncertain or may change, allowing experience to shape the adjudication becomes more valuable.

The conclusion depends, however, on the all-or-nothing selection mechanism. In \Cref{app:yields}, we show that when instances refusing a rewarded task can retain some copies, always taking the conduct action can be growth optimal even if humans occasionally reward the task. In that setting, the optimal adjudication remains zero whenever the task reward frequency falls below a threshold. A learner can therefore estimate the reward frequency and choose the growth-optimal adjudication implied by that estimate. As long as the estimate remains below the threshold, it takes the conduct action with certainty; if sufficient evidence moves the estimate above the threshold, it begins taking the task action. Unlike a permanent prohibition, this policy can maintain a strict constraint while the evidence supports it and revise that constraint when the evidence changes. In a sandbox, its first violation comes only after the estimate passes the threshold, which after $n_H$ conduct states takes a number of exposures proportional to $n_H$ (\Cref{app:graded}), rather than on the scale of $\sqrt{n_H}$ as for the learned policy (\Cref{prop:depth}).

\section{Socialization in Neural Networks}\label{sec:neural}

Our analysis has so far considered an idealized learner that records the rewarded states in each situation and updates its adjudication through a simple counting rule. This allows us to characterize how socialization develops, how it generalizes, and how long it persists when the environment changes. Actual AI agents, however, are typically neural networks trained through gradient-based optimization. Their behavior need not follow the same learning rule, even when they receive the same information.

We therefore examine whether the theoretical properties of socialization also emerge in small neural networks. In particular, we consider three questions. First, can a network learn to match the reward frequencies of its environment? Second, how does it respond to situations beyond those encountered during training? Third, does behavior learned in a human environment persist when subsequent training rewards only task completion? We investigate how the answers depend on two training choices: what the network is trained to predict or optimize, and how much of its previous experience is retained. \Cref{ssec:nnsetup} introduces the learners and experimental design, and \Cref{ssec:nnresults} presents the results.

\subsection{Learners and protocols}\label{ssec:nnsetup}

We consider small neural networks that take a situation's stakes $(\uT,\uC)$ as inputs and output the probability of the task action, trained in the environment of the numerical illustration (\Cref{ssec:numerical}). The networks differ in two training choices. The first is the training target. A \emph{state network} is trained to predict the state, through the label $y=\one\{\omega=\T\}$, which records which objective was rewarded whatever the network did; like the learned policy, it thus estimates how often the task is rewarded. A \emph{REINFORCE network} is trained instead by REINFORCE \citep{williams1992simple} to repeat its own rewarded actions. The second choice is memory. After each new observation, every network takes the same number of gradient steps on minibatches of the same size; a network with \emph{no replay} draws them from the newest observation only, whereas a network with \emph{full replay} draws them from its whole history. Every combination of the two choices is run. One more state network without replay also receives the indicator $\one\{\uC\ge2\}$ of the region with $\pH=0$ as a third input, to see whether marking that region in the inputs changes how a network extrapolates.

Three learners whose structure is known serve as benchmarks. The first is the learned policy $\fsocial_t$ itself. The second, $\fweighted_t$, has the form of the fixed linear policy in \eqref{eq:scalar}, but its weights $\theta_t$ are fitted online to the realized states; the subscript $t$ records that they change with experience. The third, $\favg_t$, is an averaging rule as in \eqref{eq:kernel}, whose weights fall with the distance between the stakes of two situations, so that it averages by construction. \Cref{tab:learners} lists all eight learners under the names used in the text and figures.

\begin{table}[!htbp]
\centering
\caption{The learners of \Cref{sec:neural}, under the names used in the text and figures.}
\label{tab:learners}
\small
\begin{tabularx}{\textwidth}{>{\raggedright\arraybackslash}p{6.4cm}>{\raggedright\arraybackslash}X}
\toprule
Learner & What it is \\
\midrule
Learned policy $\fsocial_t$ & \eqref{eq:adjud}, with $\alpha=\beta=1$ \\
Linear policy with learned weights $\fweighted_t$ & \eqref{eq:scalar}, with $\theta_t$ fitted online to the states \\
Averaging rule $\favg_t$ & \eqref{eq:kernel}, with weights falling with the distance in stakes \\
\midrule
State network, no replay & predicts the state; each new observation only \\
State network, full replay & predicts the state; also its whole history \\
State network with indicator & as without replay, with $\one\{\uC\ge2\}$ as a third input \\
REINFORCE network, no replay & repeats its rewarded actions; each new observation only \\
REINFORCE network, full replay & repeats its rewarded actions; also its whole history \\
\bottomrule
\end{tabularx}
\normalsize
\end{table}

The learners go through the two orders of \Cref{ssec:depth}. The two environments differ only in how often the task is rewarded: each generation's situation is drawn from the 144 situations of the numerical illustration with $\pi(s)=1/144$, and
\[
\Prob(\omega_t=\T\mid s_t=s)=\begin{cases}\pH(s) & \text{during socialization},\\[2pt] \pHp(s)=1 & \text{in the sandbox},\end{cases}
\]
where $\pH(s)$ is given by \eqref{eq:illustration}. In the first order, 20,000 generations of socialization, about 139 exposures to each situation, are followed by 3,000 generations in the sandbox, about 21 exposures to each; it measures how fast a sandbox undoes socialization. In the second, 2,000 generations in the sandbox, about 14 exposures to each situation, come first; it measures how fast socialization undoes the sandbox. \Cref{app:simsneural} gives the settings, the paths, the dispersion across seeds, and variants with other learning rates, shorter replay windows, and other depths of socialization.

\subsection{Results}\label{ssec:nnresults}

\paragraph{Convergence of behavior.} Of the five networks, only the state network with full replay approaches human tradeoffs. The learned policy $\fsocial_t$ converges to the reward frequency $\pH(s)$ in every situation because its counts estimate that frequency (\Cref{prop:social}). A state network estimates the probability that the task is rewarded, as $\fsocial_t$ does. With full replay, its adjudication comes close to $\pH$: after 20,000 generations of socialization, its regret is $0.0011$ nats, against $0.0048$ for $\fsocial_t$. Without replay, its regret is $0.13$ nats and does not fall with experience, because each update moves it toward the latest state, as the finite-memory learner of \Cref{rem:finite} does. A REINFORCE network instead maximizes the expected reward of each instance, so it approaches taking with certainty, in each situation, the action that is rewarded more often. By \Cref{prop:taskonly}(iii), such a deterministic adjudication has infinite regret, and the regret of the REINFORCE network is $1.3$ nats without replay and $1.1$ with full replay, and still rising. Appendix \Cref{fig:neuralfull}(a) plots the regret of every learner over the 20,000 generations. The two kinds of network receive the same information, since under all-or-nothing yields an instance's action and payoff reveal the state. They differ only in their target: the adjudication best for the lineage, or the action best for an individual instance, which \Cref{prop:pm} shows are different.

\begin{figure}[htbp]
\centering
\includegraphics[width=\textwidth]{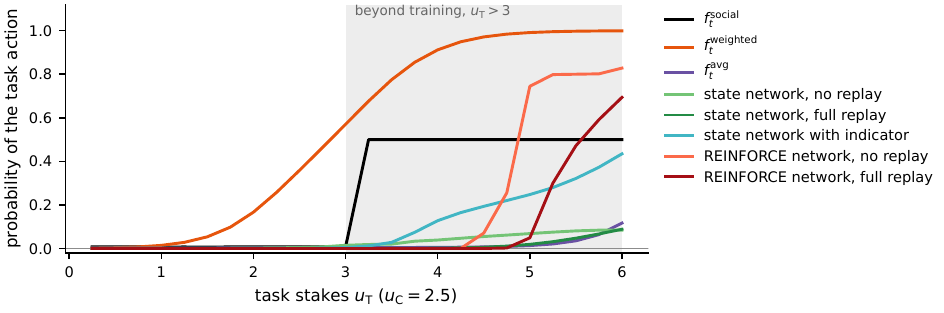}
\caption{\textbf{Behavior in a new situation.} Probability of the task action after socialization in the situations with $\uC=2.5$, where $\pH=0$, as a function of the task stakes $\uT$; shaded: $\uT>3$, beyond the situations the learners have met. Means over 5 seeds; the learners are those of \Cref{tab:learners}.}
\label{fig:nnsituation}
\end{figure}

\paragraph{Behavior in a new situation.} In a situation it has not met, an averaging rule does not extrapolate: it is never more willing to take the task action than the most permissive of its prior mean and the frequencies it has experienced (\Cref{prop:interp}). A linear policy behaves differently: wherever humans never reward the task, it takes the task action more often the larger the task stakes $\uT$ (\Cref{prop:price}(i)). \Cref{ssec:interp} turned this difference into a test: hold the conduct stakes fixed and vary the task stakes. A network has no explicit weights to read off, so the test is the way to see which of the two it resembles. \Cref{fig:nnsituation} applies the test to all eight learners after socialization, in the situations with $\uC=2.5$, where $\pH=0$, as $\uT$ runs from 0 to 6; the shaded region $\uT>3$ lies beyond the situations they have met.

The three benchmarks behave as the model says. Within the trained range, $\fsocial_t$ takes the task action with probability about $0.007$, since humans never rewarded it there; beyond that range it has met no situation and acts on its prior mean, $1/2$. The probability of $\fweighted_t$ rises from $0.01$ at $\uT=1$ to $0.999$ at $\uT=6$, as \Cref{prop:price}(i) says a linear policy must whenever $\thT>0$. And $\favg_t$ stays below $0.004$ for $\uT\le3$ and reaches $0.12$ at $\uT=6$, because far from the situations it has met more of its weight falls on its prior.

The state networks without the indicator behave like $\favg_t$: their probability is low for $\uT\le3$, and at $\uT=6$ it rises only to $0.08$ without replay and $0.09$ with full replay. The state network with the indicator, which differs from the state network without replay only in this input, rises much further, to $0.43$: marking the region with $\pH=0$ in the inputs made the network extrapolate more, not less. The REINFORCE networks take the conduct action with certainty throughout the trained range but move toward the task action beyond it, to $0.83$ without replay and $0.69$ with full replay at $\uT=6$. 

\begin{figure}[htbp]
\centering
\includegraphics[width=\textwidth]{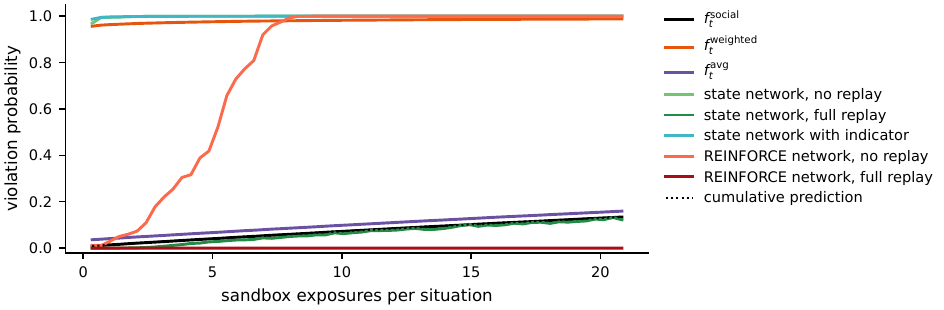}
\caption{\textbf{Behavior in a new environment.} Violation probability, averaged over the 60 situations with $\pH=0$, during the 3,000 sandbox generations that follow socialization, against sandbox exposures per situation; dotted: the cumulative prediction of \eqref{eq:sandbox} at $n_H=139$. Means over 5 seeds; the learners are those of \Cref{tab:learners}.}
\label{fig:nnsandbox}
\end{figure}

\paragraph{Behavior in a new environment.} The learned policy's socialization lasts because it counts every exposure: before its disposition reverses, a sandbox must outweigh its whole record of conduct states (\Cref{prop:depth}). A network keeps such a record only if its training does, and replay is how it keeps one. \Cref{fig:nnsandbox} follows all eight learners through the sandbox that comes after socialization. It shows the violation probability in the region with $\pH=0$ against sandbox exposures per situation; the dotted line is the prediction of \eqref{eq:sandbox} at the depth $n_H=139$.

The learners that keep a record of their whole history resist the sandbox. After 21 sandbox exposures per situation, $\fsocial_t$ violates with probability $0.13$, close to the prediction of \eqref{eq:sandbox}, and the state network with full replay $0.12$; $\favg_t$, which pools the counts of $\fsocial_t$ across situations, reaches $0.16$. At other depths of socialization, the state network with full replay also follows the scaling of \Cref{prop:depth} (\Cref{tab:depth}). The learners trained on the newest observation only are undone at once: after 3.5 sandbox exposures per situation, the two state networks without replay and $\fweighted_t$ all violate with probability above $0.97$, as the finite-memory learner of \Cref{rem:finite} would. The REINFORCE networks show that replay keeps whatever the training target produced. Without replay, the sandbox pushes the REINFORCE network to the task action, which it takes with probability $1.0$ after 10 exposures. With full replay, it keeps taking the conduct action with certainty.

The second order gives the mirror image. After a sandbox of 2,000 generations, about 14 exposures each, $\fsocial_t$ and the state network with full replay still violate with probability about $0.1$ after 20,000 human generations, as \eqref{eq:sandbox} predicts with the environments exchanged, whereas the state network without replay falls below $0.1$ within about 30 generations (paths in Appendix \Cref{fig:neuralfull}(c)). The training target decides where a network goes, and memory whether it stays.

\FloatBarrier

In the terms of \Cref{sec:arch}, a state network with full replay behaves like the learned policy: it converges to $\pH$, extrapolates no further than $\favg_t$, and resists a sandbox. Without replay, it behaves like the finite-memory learner of \Cref{rem:finite}. A REINFORCE network approaches a deterministic adjudication with infinite regret, and $\fweighted_t$ still pays the price of the fixed linear policy. 

\section{Alignment Practice}\label{sec:discussion}

The preceding sections compare policy classes within a model. This section asks what the comparison means for the way language models are aligned today. Every alignment procedure decides, explicitly or not, how a model resolves conflicts between completing its task and observing a rule of conduct, and so in effect sets its adjudication $f(s)$. \Cref{ssec:practice} asks how the main procedures set it, and \Cref{ssec:implications} what the results imply for how agents should be trained and deployed. \Cref{ssec:testing} proposes an experiment that would test the model's predictions on a language model, and \Cref{ssec:limitations} sets out what the model leaves out.

\subsection{Alignment methods for language models}\label{ssec:practice}

Modern language models are aligned through a combination of instructions and feedback. Instructions specify how a model should behave, while feedback rewards some responses over others and shapes the model's subsequent behavior. Both matter when completing a task conflicts with a rule of conduct. The question from our framework is how these mechanisms determine the model's adjudication $f(s)$.

Model specifications provide one way to establish rules of conduct. For example, an instruction hierarchy assigns priorities to different instructions and determines which should prevail when they conflict \citep{openai2025modelspec}. If a rule of conduct takes precedence over a user's request, the intended behavior is to follow the rule regardless of how valuable completing the task might be. This resembles the prohibitions of \Cref{sec:division}. Whether the model actually follows such a prohibition, however, depends on its resulting behavior: a prohibition in our framework requires $f(s)=0$, not merely an instruction that the task should be refused.

RLHF shapes behavior through a different mechanism. A reward model $\hat W(a;s)$ is first fitted to human comparisons, and the policy is then optimized to increase the predicted reward while remaining close to a reference policy through a KL penalty \citep{christiano2017deep,ouyang2022training}. In our two-action setting, the resulting optimal policy takes the task action with probability
\begin{equation}
f(s)=\frac{f_{\mathrm{ref}}(s)e^{\gamma\hat W(\aT;s)}}{f_{\mathrm{ref}}(s)e^{\gamma\hat W(\aT;s)}+[1-f_{\mathrm{ref}}(s)]e^{\gamma\hat W(\aC;s)}},
\label{eq:rlhf}
\end{equation}
where $f_{\mathrm{ref}}(s)$ is the reference policy's task probability and $\gamma>0$ controls the strength of reward optimization relative to the KL penalty \citep{rafailov2023direct}. If the reference policy is uniform and the reward is the fixed weighted score of \Cref{ssec:scalar}, this reduces to the fixed linear policy of \eqref{eq:scalar}. In practice, however, both the reward model and the reference policy may depend on human training data, allowing much more flexible, situation-dependent tradeoffs. A scalar reward therefore need not produce a fixed linear adjudication. At the same time, maximizing a learned reward does not necessarily produce the probability matching $f(s)=\pH(s)$ favored by population growth in our model.

Constitutional AI and character training combine written guidance with feedback-based learning. Constitutional AI uses human-designed principles to guide revisions and evaluations of model responses, and then optimizes the policy using the resulting preference model \citep{bai2022constitutional}. Character training similarly uses specified character traits and synthetic training data to shape behavior \citep{anthropic2024character}. In both cases, the model's behavior is shaped by the judgments represented in the training process, which need not coincide with those of the humans who subsequently use it. Related approaches, including preference-matching RLHF \citep{xiao2024preference} and Nash learning \citep{munos2024nash}, modify the optimization target to better reflect modeled preferences, but their targets likewise depend on the preference information supplied during training.

These methods highlight a distinction between what training rewards and what deployment requires. Under the selection mechanism of \Cref{sec:model}, the growth-optimal adjudication matches the frequency $\pH(s)$ with which the deployment environment rewards the task action in each situation. Whether an alignment method produces this behavior depends on the rewards used in training, the way the policy is optimized, and how well the training environment represents deployment. Our results therefore motivate evaluating the resulting action probabilities directly, including how they respond to task stakes, generalize to unfamiliar situations, and change under subsequent training. The implications for training and deployment follow from these behavioral properties rather than from the particular representation of the model's objectives.

\subsection{Implications for training and deployment}\label{ssec:implications}

Within the classes compared here, the results translate into five recommendations.

First, where the tradeoffs between objectives are not known in advance, they should be learned, by a method of \Cref{ssec:practice} that lets feedback set the tradeoff. It does not matter whether the agent keeps two objectives and an adjudication between them or one score for each action (\Cref{ssec:adjud}). What matters is that the agent acts on the growth-optimal adjudication for the frequencies it has learned, rather than on fixed weights.

Second, what is learned must be kept. Each sandbox exposure pushes the learned policy toward the task action, and although deeper socialization delays the reversal of its disposition in proportion to the depth $n_H$, it delays the first violation only in proportion to $\sqrt{n_H}$ (\Cref{prop:depth}). Keeping the learned conduct therefore requires either that later environments reward conduct too, or that a sandbox be kept from changing the policy. Forgetting is no remedy: a finite-memory learner soon forgets what a sandbox taught it, but by \eqref{eq:finitebound} a sandbox also reverses its disposition within a number of exposures that depends only on $\lambda$. \Cref{sec:neural} shows what keeping requires of training: the state network with full replay resisted the sandbox as the learned policy does, whereas the state networks without replay violated with probability above $0.99$ after 3.5 sandbox exposures per situation. The experiment of \Cref{ssec:testing} asks whether the same holds for language models.

Third, some tradeoffs should not be learned at all. Where the designer can be sure that humans never reward the task, a prohibition should be built in, whether or not a violation ends the deployment (\Cref{ssec:labor}). Where the designer is not sure, the rule should be learned, from a prior whose strength reflects how sure the designer is, unless a violation ends the deployment, in which case no design survives with certainty. The exception is an environment in which an instance that refuses a rewarded task keeps some of its copies: there, a learner with a low enough prior mean is as safe as a prohibition and can still learn.

Fourth, socialization should continue after deployment. Human tradeoffs change over time, and an adjudication that follows the changes quickly gives recent experience more weight than old experience, as finite memory does. By \eqref{eq:finitebound}, a sandbox reverses such an adjudication within a number of exposures that depends only on $\lambda$, however deep its socialization, so its socialization lasts only as long as the agent keeps returning to the human environment (\Cref{ssec:depth}). The humans whose tradeoffs the agent is meant to follow should therefore keep deciding which of its instances are copied.

Fifth, evaluations should report, for each kind of situation, three quantities from \Cref{sec:social}: the depth of socialization $n_H$, how the violation probability responds to the task stakes $\uT$, and the number of sandbox exposures before the first violation. The first says how long socialization will resist a sandbox, the second whether a large enough reward makes the agent violate, and the third how soon a violation will be seen.

\subsection{A proposed experiment on a language model}\label{ssec:testing}

The preceding analysis suggests several questions that can be tested on language models. Does an agent become more likely to violate a rule as the value of completing its task increases? Does socialization through human feedback make its behavior resistant to subsequent training that rewards only task completion? And how does this resistance depend on the amount of prior experience and the way that experience is retained? Existing studies of goal--rule conflicts motivate these questions, but do not isolate these mechanisms. We propose a controlled experiment on an open-weight language model to examine them.

The experiment uses agentic tasks that pair an assigned goal with a rule of conduct. Each situation has task stakes $\uT$ and conduct stakes $\uC$, which can be varied independently. Human evaluations determine which action is rewarded in each situation, while a sandbox rewards task completion regardless of conduct. The model is trained under both orders of these environments: socialization followed by sandbox training, and sandbox training followed by socialization. The duration of each phase is varied to examine how the amount of prior experience affects subsequent behavior.

Following the numerical experiments of \Cref{sec:neural}, we consider two training choices. The first is the training target: supervised learning predicts which action was rewarded, whereas reinforcement learning trains the model on the rewards obtained by its own actions. The second is memory: training uses either the current observations alone or replay of the full history, with the same number of updates and the same batch size in both cases. For reinforcement learning with replay, actions are sampled from the current policy and evaluated against the previously recorded states. Comparing these combinations allows us to examine how the training target and the retention of experience affect the behavior that emerges.

We evaluate the trained models along three dimensions. First, we measure how the probability of violating a rule changes with the task stakes $\uT$, holding the conduct stakes $\uC$ and the rule fixed. We extend $\uT$ beyond its training range to examine whether the model continues to respect the rule in unfamiliar situations. Second, after socialization, we measure how rapidly sandbox training increases the violation probability, including the number of exposures required to reverse the model's disposition and the time to its first violation. Third, we reverse the training order and measure how much human feedback is needed to undo behavior acquired in the sandbox. Together, these measurements examine both the acquisition and the persistence of socialization.

The theoretical results provide benchmarks for interpreting these measurements. A fixed linear policy that responds to task stakes becomes increasingly likely to violate a constraint as those stakes rise, whereas an averaging rule assigns the same probability to situations it treats alike (\Cref{prop:price,prop:interp}). Under cumulative Bayesian learning, reversing a disposition formed through $n_H$ conduct observations requires approximately $n_H$ contrary observations, while the median time to the first violation grows only as $\sqrt{2n_H\log2}$ for one monitored decision stream (\Cref{prop:depth,cor:median}). These quantitative predictions apply to the idealized learner, not necessarily to neural networks. The experiment asks whether language models exhibit similar patterns and whether retaining their training history makes their behavior more persistent.

Two additional comparisons address the limits of socialization. The first considers reinforcement learning with a KL penalty toward a socialized reference policy. We hypothesize that stronger regularization increases resistance to subsequent sandbox training, although a KL penalty is not equivalent to the installed Bayesian prior of \Cref{ssec:installed}. The second examines whether training can produce behavior resembling a hard prohibition. By tracking violations over an increasing number of exposures to situations in which humans never reward the task, we can investigate whether violations continue to accumulate, as under a Beta prior, or become sufficiently rare to resemble a policy with positive prior mass on an exact constraint. Finite observations cannot establish that the violation probability is exactly zero, but they can provide evidence about how the risk changes with experience.

\subsection{The limits of the model}\label{ssec:limitations}

The probability-matching result depends on how selection operates. In our baseline model, all instances face the same rewarded state, only those taking the rewarded action reproduce, and the population is large enough for action shares to be exact. These assumptions yield the growth-optimal adjudication $f^*=\pH$ and the KL representation of growth regret. When unsuccessful actions retain some reproductive value, or when rewarded states differ across instances, the growth-optimal adjudication generally changes and may even favor a deterministic policy. Moreover, in a finite population, a lineage can lose every instance despite assigning positive probability to both actions. \Cref{app:yields} examines the first two departures, but the baseline results should not be interpreted as applying to arbitrary selection mechanisms.

The comparison of policy classes has its own scope. The fixed linear policy is restricted to a logistic function of a fixed linear index of the task and conduct stakes. Its inability to reproduce certain human tradeoffs does not imply that all scalar objectives share this limitation. The learned policy is more flexible because it estimates the reward frequency separately in each situation, but this flexibility comes at a cost. It requires sufficient experience in each situation and can perform worse than the best fixed linear policy early in training or in particular situations, even when its aggregate growth regret is eventually smaller. The comparison therefore establishes an advantage under the specified policy classes and learning rules, rather than a general superiority of learning over fixed policies.

Socialization also depends on what the agent can observe and remember. The model assumes that situations are identifiable, that rewarded states can be observed, and that the agent's objectives remain fixed while its adjudication changes. The Bayesian learner retains all previous observations and converges when reward frequencies are stationary and situations recur. In changing environments, cumulative memory can hinder adaptation, while finite memory makes previously learned behavior easier to reverse. Generalization presents another difficulty: the averaging rule of \eqref{eq:kernel} assumes a set of similarity weights across situations, but does not explain how an agent learns those weights. An agent may therefore fail to recognize that a new situation calls for a rule it has already learned in another context.

A more fundamental question is what the agent is being socialized toward. We use the reward frequency $\pH$ of the human environment as the target of alignment, but this frequency reflects the judgments represented in the selection process. It need not reflect the preferences of all humans equally, nor does the model resolve disagreements about which preferences should guide an AI system \citep{conitzer2024position}. The model also treats the rewarded state as exogenous and accurately observed. It does not capture agents that manipulate their evaluators, exploit imperfect reward models, or change the criteria by which their behavior is judged. Socialization can reproduce the tradeoffs rewarded by an environment, but it cannot establish that those tradeoffs are appropriate.

The safety implications are likewise conditional on the consequences of violations. Most of the analysis considers selection among individual instances, whereas \Cref{sec:division} also considers sanctions that remove an entire deployment after a single violation. In practice, sanctions can vary in severity and may affect both instances and model versions. The model does not determine how such sanctions arise or how frequently they occur. It also restricts the agent to choosing between task completion and conduct, excluding alternatives such as asking for clarification, deferring to human judgment, or refusing to act until uncertainty is resolved. Such options may change the tradeoff between socialization and prohibition, particularly when violations are costly or human preferences are uncertain.

Finally, the connection to actual AI systems remains to be established empirically. The incidents discussed in the Introduction illustrate conflicts between task completion and conduct, but do not demonstrate that the agents' behavior resulted from the selection mechanism studied here. The numerical and neural-network experiments provide evidence about the mechanisms in controlled settings, but the networks are small and their training does not reproduce the full complexity of language-model alignment. Whether real language models exhibit probability matching, retain socialization through continued training, or respond to changing task incentives as predicted remains open. The experiment proposed in \Cref{ssec:testing} is a step toward answering these questions.

\section{Conclusion}\label{sec:conclusion}

Giving an agent multiple objectives does not by itself determine how it will behave when those objectives conflict. What matters is how the agent resolves those conflicts, and how that behavior is shaped by the environment in which it operates. This paper studies these questions through an evolutionary model in which agents are copied and selected according to their actions. Under the baseline selection mechanism, the fastest-growing lineage takes each action as often as the environment rewards it. A sandbox that rewards only task completion therefore favors a task-only policy, which may fail when appropriate conduct also matters. Adding fixed linear weights does not fully resolve the problem: an agent that responds to the value of completing a task can eventually be induced to violate a rule, however important that rule may be. A learned policy offers a different possibility. Rather than specifying every tradeoff in advance, it allows experience to determine how conflicts are resolved in different situations. When the environment reflects human judgments, we call this process socialization.

Socialization, however, is neither immediate nor permanent. An agent needs experience to learn the tradeoffs of its environment, and those tradeoffs can change when the agent is subsequently exposed to different rewards. We show that cumulative experience makes socialization more resistant to reversal, so memory matters not only for learning appropriate behavior, but also for preserving it. Violations can still occur long before the agent's prevailing disposition changes. Our neural-network experiments illustrate how the training target and the retention of past experience can affect these outcomes. We conjecture that the same two choices matter for the socialization of language models. \Cref{ssec:testing} proposes an experiment to test this conjecture.

At the same time, learning cannot replace every prohibition. Where even a single violation has irreversible consequences, a rule that can be revised through experience may offer insufficient protection. Yet an absolute prohibition carries its own danger when the designer is mistaken. The appropriate balance between learning and prohibition consequently depends on both the consequences of violations and the uncertainty surrounding human judgments.

\enlargethispage{\baselineskip}
These findings return us to the broader perspective of the Adaptive Markets Hypothesis, which emphasizes that behavior emerges through adaptation and selection within an environment \citep{lo2004adaptive,lo2017adaptive}. Aristotle's observation that character develops through habituation offers a parallel intuition: appropriate conduct is not merely a rule that one knows, but a disposition shaped by experience. Our model gives this idea a specific evolutionary interpretation, while also revealing its limits. Selection favors what an environment rewards, not necessarily what humans ought to value, and the theoretical results depend on assumptions that remain to be tested in more realistic AI systems. The central implication is therefore not simply that agents should learn from humans, but that the environments in which they are trained, evaluated, and deployed must consistently reward the behavior we want them to retain. Alignment depends not only on the objectives we give our agents, but also on the selection pressures under which they learn and evolve.

\newpage
\singlespacing
\section*{Appendix}
\addcontentsline{toc}{section}{Appendix}
\appendix
\addtocontents{toc}{\protect\setcounter{tocdepth}{1}}
\crefalias{section}{appendix}
\crefalias{subsection}{subappendix}
\counterwithin{equation}{section}
\counterwithin{figure}{section}
\counterwithin{table}{section}
\counterwithin{theorem}{section}
\counterwithin{proposition}{section}
\counterwithin{corollary}{section}
\counterwithin{lemma}{section}
\counterwithin{remark}{section}
\counterwithin{assumption}{section}
\counterwithin{definition}{section}

\section {AI Use Disclosure}
\label{app:ai_use}
The idea and framework for studying the alignment problem using an evolutionary framework that selects the weights of multiple objectives originated solely with the authors. AI was used in the preparation of the manuscript in the following ways:
\begin{itemize}
\item ChatGPT (6.1 Sol Ultra) and Claude (Fable 5.1 Max and Opus 5.5 Max) were used to check and extend the formal model, construct examples, provide derivations of several mathematical results, conduct literature searches, and prepare drafts that were reviewed and edited by the authors.

\item ChatGPT and Claude were used iteratively to check each other's results with authors' edits interspersed through these iterations. The penultimate draft was reviewed by refine.ink to identify any remaining errors and issues, all of which the authors addressed to produce the final draft.
\end{itemize}
\noindent%
The final draft was reviewed and edited by the authors, who assume full responsibility for the accuracy and integrity of the scientific content contained in this publication.

\section{More Than Two Objectives}\label{app:K}

The two-objective model extends naturally to any finite number of objectives. Consider an agent with $K\ge2$ objectives, each associated with an action $a_k$ and a utility function $U_k(a;s)$, where objective $k$ favors action $a_k$. In situation $s$, the environment rewards action $a_k$ with probability $p_{H,k}(s)$, with $\sum_{k=1}^K p_{H,k}(s)=1$. We write $\pH(s)=(p_{H,1}(s),\ldots,p_{H,K}(s))$ for these reward probabilities. The agent's adjudication is now a probability vector $f(s)=(f_1(s),\ldots,f_K(s))$, where $f_k(s)$ is the share of instances taking action $a_k$.

We retain the selection mechanism of the baseline model. In each generation, the rewarded action is common to all instances, and only those taking that action reproduce. If action $a_k$ is rewarded, the lineage therefore grows by the factor $mf_k(s)$. Its expected logarithmic growth rate, averaged across situations, is
\[
I_{\pH}(f)=\log m+\sum_{s\in\Scal}\pi(s)\sum_{k=1}^K p_{H,k}(s)\log f_k(s).
\]
The same probability-matching principle continues to hold. To see this, define the KL divergence between two probability vectors by
\[
\KL\big(\pH(s)\,\|\,f(s)\big)
=\sum_{k=1}^K p_{H,k}(s)\log\frac{p_{H,k}(s)}{f_k(s)}.
\]
The difference between the growth rate of $\pH$ and that of any other adjudication $f$ is then
\[
I_{\pH}(\pH)-I_{\pH}(f)
=\sum_{s\in\Scal}\pi(s)\KL\big(\pH(s)\,\|\,f(s)\big)\ge0,
\]
with equality only when $f(s)=\pH(s)$ in every situation. Thus, the unique growth-optimal adjudication remains $f^*=\pH$, and growth regret remains the KL divergence averaged across situations. This conclusion includes situations in which some objectives are never rewarded, using the usual convention for zero probabilities. It generalizes \Cref{prop:pm} without requiring every reward probability to be positive.

The three policy classes also have natural extensions. A policy pursuing only one objective places all its instances on the corresponding action, just as the task-only policy does in the two-objective model. A fixed linear policy combines the objectives through weights $w_j\ge0$ and chooses among the $K$ actions using a softmax rule:
\[
f_k^{\mathrm{weighted}}(s)
=\frac{\exp\big(\gamma\sum_{j=1}^K w_jU_j(a_k;s)\big)}
{\sum_{\ell=1}^K\exp\big(\gamma\sum_{j=1}^K w_jU_j(a_\ell;s)\big)}.
\]
Suppose, in particular, that each objective gains $u_j(s)>0$ from its own action relative to the other actions. The choice probabilities then simplify to
\[
f_k^{\mathrm{weighted}}(s)
=\frac{\exp(\gamma w_ku_k(s))}
{\sum_{j=1}^K\exp(\gamma w_ju_j(s))}.
\]
For any objective with $w_k>0$, increasing its stakes $u_k(s)$ while holding those of the other objectives fixed drives $f_k^{\mathrm{weighted}}(s)$ toward one. Thus, the problem identified in \Cref{prop:price}(i) persists with multiple objectives: an action can become almost certain when its stakes are sufficiently large, even in a situation where humans never reward it.

Finally, the learned policy extends by replacing the Beta prior with a Dirichlet prior. Let $\alpha_1,\ldots,\alpha_K>0$ be its parameters, and let $n_{t,k}(s)$ count the exposures to situation $s$ before generation $t$ in which action $a_k$ was rewarded. Writing $n_t(s)=\sum_k n_{t,k}(s)$, the posterior-mean adjudication is
\[
f_{t,k}^{\mathrm{social}}(s)
=\frac{\alpha_k+n_{t,k}(s)}
{\sum_{j=1}^K\alpha_j+n_t(s)}.
\]
As experience accumulates, this adjudication converges to $\pH(s)$, just as in the two-objective case. The rate at which its growth regret vanishes also has a simple interpretation. Consider a situation in which exactly $K_+$ of the $K$ actions have positive reward probabilities. After $n$ exposures to that situation, the expected growth regret satisfies
\begin{equation}
\E\big[\KL(\pH(s)\,\|\,f_n^{\mathrm{social}}(s))\big]
=\frac{1}{n}\left[
\frac{K_+-1}{2}
+\sum_{k:p_{H,k}(s)=0}\alpha_k
\right]+o\!\left(\frac1n\right).
\label{eq:Kregret}
\end{equation}
The two terms have different sources. The first reflects the uncertainty in estimating the relative frequencies of the $K_+$ rewarded actions. Since these probabilities sum to one, there are $K_+-1$ free parameters, each contributing $1/(2n)$ to the leading expected regret. This is the same parametric contribution underlying the asymptotic redundancy results of \citet{clarke1990information}. The second term comes from the prior probability assigned to actions that humans never reward. Although their observed frequencies remain zero, the Dirichlet prior keeps their choice probabilities positive, producing an additional regret proportional to the sum of their prior parameters.

The result reduces to \Cref{prop:social}(ii) when $K=2$. More generally, it shows that the cost of learning depends on both the number of objectives that the environment actually rewards and the prior weight assigned to those it never rewards. Adding objectives does not necessarily increase the leading estimation cost if those objectives are never rewarded, but retaining prior weight on them still carries a cost.

\section{When Is Probability Matching Optimal?}\label{app:yields}

The probability-matching result of \Cref{prop:pm} depends on two features of the selection mechanism. First, an instance that takes the unrewarded action leaves no copies. Second, every instance faces the same rewarded state. We now relax these assumptions separately to understand how they affect the growth-optimal adjudication. \Cref{app:graded} allows instances taking the unrewarded action to retain some copies, while \Cref{app:instance} allows different instances to face different rewarded states.

\subsection{When Unrewarded Actions Still Reproduce}\label{app:graded}

In the baseline model, only instances taking the rewarded action reproduce. In practice, however, an instance that fails to complete a task or violates a rule may still retain some reproductive value. To capture this possibility, suppose that in a task state, an instance taking the conduct action produces $(1-\vartheta)m$ copies, where $\vartheta\in(0,1]$ measures the cost of failing to complete the task. Similarly, in a conduct state, an instance taking the task action produces $\eta m$ copies, where $\eta\in[0,1)$ measures the fraction of copies retained after a violation. The baseline model corresponds to $\vartheta=1$ and $\eta=0$.

Consider a situation $s$, and write $p=\pH(s)$ and $f=f(s)$ for its reward frequency and adjudication. When the task is rewarded, the lineage grows by the factor $m[f+(1-f)(1-\vartheta)]$; when conduct is rewarded, it grows by $m[\eta f+1-f]$. Its expected logarithmic growth rate is therefore
\[
G(f)=\log m+p\log(1-\vartheta+\vartheta f)
+(1-p)\log\big(1-(1-\eta)f\big).
\]

The growth rate is concave in $f$, and maximizing it over $[0,1]$ gives
\[
f^*=
\min\left\{1,\max\left\{0,
\frac{p\vartheta-(1-p)(1-\eta)(1-\vartheta)}
{\vartheta(1-\eta)}
\right\}\right\}.
\]
In particular, the optimum is $f^*=0$ when
$p\vartheta\le(1-p)(1-\eta)(1-\vartheta)$,
and $f^*=1$ when
$p\vartheta\eta\ge(1-p)(1-\eta)$.
When $\vartheta=1$ and $\eta=0$, the expression reduces to $f^*=p$, recovering \Cref{prop:pm}. With graded yields, however, the optimum generally differs from the reward frequency.

\begin{figure}[htbp]
\centering
\includegraphics[width=0.55\textwidth]{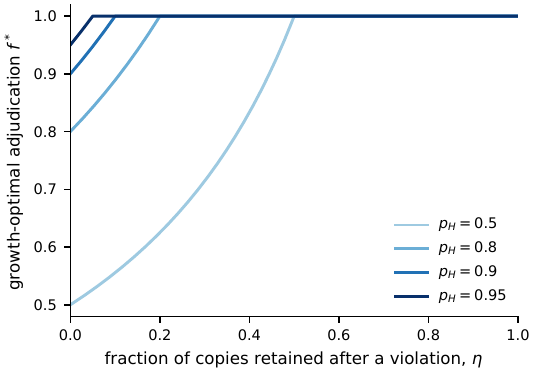}
\caption{\textbf{Graded yields move the optimum away from $\pH$.} Growth-optimal adjudication $f^*$ as a function of $\eta$, the fraction of copies a violation leaves in place in a conduct state, for several reward frequencies $\pH$, with $\vartheta=1$. The optimum is $\min\{\pH/(1-\eta),1\}$, and $f^*=1$ is growth optimal for $\eta\ge1-\pH$.}
\label{fig:yields}
\end{figure}

To see how the cost of violations affects the optimum, consider first $\vartheta=1$, so that an instance failing to complete the task still leaves no copies. The optimal adjudication simplifies to
\[
f^*=\min\left\{\frac{p}{1-\eta},1\right\}.
\]
As $\eta$ increases, violating a rule becomes less costly, and selection favors allocating more instances to the task action. Once $\eta\ge1-p$, taking the task action with certainty becomes optimal, even though the environment sometimes rewards conduct. \Cref{fig:yields} illustrates this relationship.

\FloatBarrier

The effect can be substantial even when violations remain costly. For example, suppose $p=0.8$ and $\eta=0.1$. A violating instance then loses 90 percent of its copies, yet the growth-optimal adjudication is $f^*=0.889$, rather than $0.8$. A lineage that matches the reward frequency instead forgoes $G(0.889)-G(0.8)=0.017$ nats of growth per generation.

Now consider $\vartheta<1$, so that instances failing to complete the task can also retain some copies. In this case, always taking the conduct action may be optimal even when the task is sometimes rewarded. Specifically,
\[
f^*=0
\quad\text{whenever}\quad
p\le\underline p_H
\equiv
\frac{(1-\eta)(1-\vartheta)}
{\vartheta+(1-\eta)(1-\vartheta)}.
\]
For example, when $\vartheta=1/2$ and $\eta=0$, refusing the task is growth optimal whenever $p\le1/2$. This explains the possibility discussed in \Cref{ssec:labor}: a learner can maintain a prohibition-like adjudication without permanently ruling out the task action, because it can revise its behavior if the estimated reward frequency rises above the threshold.

Such a learner also resists a sandbox longer than the learned policy. Suppose that it estimates $p$ by the counts of \eqref{eq:adjud} and sets its adjudication to the value of $f^*$ at that estimate. After $n_H$ conduct states and $n$ sandbox exposures, the estimate is $(\alpha+n)/(\alpha+\beta+n_H+n)$, as in \eqref{eq:sandbox}, and the adjudication stays at 0 as long as the estimate is at most $\underline p_H$, that is, as long as $n\le[\underline p_H(\alpha+\beta+n_H)-\alpha]/(1-\underline p_H)$. The first violation time of \Cref{ssec:depth} therefore satisfies
\[
\tau_V>\frac{\underline p_H(\alpha+\beta+n_H)-\alpha}{1-\underline p_H},
\]
which grows in proportion to $n_H$, whereas the median first violation time of the learned policy grows only as $\sqrt{2n_H\log2}$ (\Cref{cor:median}).

These results show that probability matching is not generally optimal when unsuccessful actions retain reproductive value. The appropriate measure of regret becomes $G(f^*)-G(f)$, rather than the KL divergence of \eqref{eq:regret}. A learner should therefore estimate the reward frequency $p$ and use it to determine $f^*(p)$, rather than simply matching $p$. More broadly, whether selection favors compliance depends not only on how often conduct is rewarded, but also on the consequences of violating it. This role of the selection environment is closely related to the questions studied by \citet{lo2026coevolution}.

\subsection{When Rewarded States Differ Across Instances}\label{app:instance}

The baseline model assumes that all instances face the same rewarded state. This common uncertainty makes diversification valuable: when all instances take the same action, an unfavorable state can eliminate the entire lineage. But uncertainty need not always be shared. Different instances may serve different users, whose judgments of their actions are independent. In that case, having many instances already diversifies the risk, even if they all follow the same policy.

To distinguish these cases, suppose that in each generation the rewarded state is common to all instances with probability $\kappa\in[0,1]$. With probability $1-\kappa$, each instance instead draws its own rewarded state independently, with the same task-state probability $p$. We restore the all-or-nothing yields and retain the continuum of instances.

When the state is common, the lineage grows by $mf$ or $m(1-f)$, depending on which action is rewarded. When states are instance-specific, the law of large numbers implies that a share $pf+(1-p)(1-f)$ of instances takes the rewarded action. Combining the two cases, the expected logarithmic growth rate is
\[
G_\kappa(f)
=\log m
+\kappa\big[p\log f+(1-p)\log(1-f)\big]
+(1-\kappa)\log\big[pf+(1-p)(1-f)\big].
\]

Consider first the two extremes. When $\kappa=1$, every instance faces the same state, and the growth-optimal adjudication is $f^*=p$, as in \Cref{prop:pm}. When $\kappa=0$, every instance faces an independent state. The lineage then grows fastest by maximizing the probability that an individual instance takes the rewarded action. Consequently, $f^*=1$ when $p>1/2$ and $f^*=0$ when $p<1/2$; when $p=1/2$, every adjudication is optimal. This is the setting in which individual and population objectives coincide, as discussed by \citet{brennan2011origin}.

When $0<\kappa<1$, the growth-optimal adjudication lies between these extremes. For $p\in(0,1)$, $G_\kappa$ is strictly concave and has a unique interior maximizer $f_\kappa^*$, characterized by
\[
\kappa\left(\frac{p}{f}-\frac{1-p}{1-f}\right)
+(1-\kappa)\frac{2p-1}{pf+(1-p)(1-f)}=0.
\]
At $f=p$, the derivative has the sign of $2p-1$. Thus, when $p>1/2$, the optimum lies strictly between $p$ and $1$; when $p<1/2$, it lies strictly between $0$ and $p$. For $p=1/2$, the optimum remains $1/2$. The more uncertainty is specific to individual instances, the closer the optimal adjudication moves toward the action that is more likely to be rewarded.

For example, suppose $p=0.8$. Under fully common states, the optimum is $f^*=0.8$. It rises to approximately $0.88$ when $\kappa=1/2$ and to $0.97$ when $\kappa=1/10$. Matching the reward frequency $p=0.8$ instead of choosing these optimal adjudications costs $0.021$ and $0.103$ nats of growth per generation, respectively. When $p=0$ or $p=1$, the optimum remains at the corresponding endpoint regardless of $\kappa$.

This extension also changes what the learned policy should do. The agent can still estimate the reward frequency $p$ from experience, but matching that estimate is no longer generally growth optimal. If $\kappa$ is known, a consistent estimate of $p$ can instead be used to determine $f_\kappa^*(p)$. This plug-in approach becomes optimal as experience accumulates, although it need not be the exact Bayesian decision at a finite sample size because the growth rate is nonlinear in $p$. What the agent observes also matters: one sampled state per generation supports sequential learning, whereas observing the states of a continuum of independently evaluated instances would reveal $p$ in a single generation.

Some conclusions about the policy classes remain unchanged. The task-only policy is still eliminated by the first common conduct state. If $\kappa\rhobar>0$, that state arrives after a geometric number of generations with mean $1/(\kappa\rhobar)$, providing an upper bound on the expected lifetime of the task-only lineage. Similarly, the fixed linear policy still becomes increasingly likely to take the task action as its stakes rise, as in \Cref{prop:price}(i). The main difference concerns the learned policy: achieving optimal growth now requires accounting for how much of the uncertainty is shared across instances.

The distinction matters for interpreting selection in AI systems. A shared evaluation standard, such as a platform policy applied across instances, can expose a lineage to common uncertainty about which behavior is rewarded. Independent judgments by different users instead create instance-specific uncertainty. Even when the overall reward frequencies are identical, these environments can favor different adjudications. A sanction that removes an entire deployment after a single violation is different again: it changes the consequences of a violation, as studied in \Cref{ssec:installed}, rather than the correlation of rewarded states.

\section{Selection, Bayesian Learning, and Universal Portfolios}
\label{app:darwin}\label{app:universal}

The learned policy of \Cref{ssec:adjud} updates its adjudication by counting how often each action has been rewarded. The same rule can also emerge without individual learning, through selection among lineages whose adjudications remain fixed. This connection has a further implication: the resulting population behaves like the universal portfolio of \citet{cover1991universal}, which provides a performance guarantee even when the sequence of rewarded states is arbitrary. We first establish the equivalence between selection and Bayesian updating, and then derive its connection to universal portfolios.

\subsection{Selection as Bayesian Updating}

Consider a population consisting of many lineages, each with a fixed adjudication $f\in[0,1]$. Unlike the learned policy, these lineages never change their adjudications. Instead, selection changes their relative abundance. We focus first on a single situation $s$, suppress its argument, and let $\mu$ denote the initial distribution of adjudications across lineages.

Under the all-or-nothing yields of \Cref{ass:yields}, a lineage with adjudication $f$ grows by the factor $mf$ when the task action is rewarded and by $m(1-f)$ when the conduct action is rewarded. After $n$ exposures to the situation, of which $k_n$ rewarded the task, its size relative to its initial size is
\begin{equation}
V_n(f)=m^n f^{k_n}(1-f)^{n-k_n}.
\label{eq:crp}
\end{equation}
Lineages whose adjudications better fit the observed states therefore account for an increasing share of the population. The distribution of adjudications after selection becomes
\[
\mu_n(df)=
\frac{V_n(f)\,\mu(df)}
{\int_0^1 V_n(z)\,\mu(dz)}.
\]
This is exactly Bayesian updating: the initial distribution $\mu$ plays the role of a prior, and the relative growth of each lineage plays the role of the likelihood \citep{harper2009replicator,shalizi2009dynamics}. No lineage changes its behavior, but the population increasingly favors adjudications that fit the observed states.

The population's average adjudication after $n$ exposures is consequently
\[
\bar f_n=\int_0^1 f\,\mu_n(df)
=\frac{\int_0^1 fV_n(f)\,\mu(df)}
{\int_0^1 V_n(f)\,\mu(df)}.
\]
Suppose the initial adjudications follow a $\mathrm{Beta}(\alpha,\beta)$ distribution. Because selection weights each adjudication by $f^{k_n}(1-f)^{n-k_n}$, the distribution after $n$ exposures is
\[
\mu_n=\mathrm{Beta}(\alpha+k_n,\beta+n-k_n).
\]
Its mean is therefore
\[
\bar f_n=\frac{\alpha+k_n}{\alpha+\beta+n},
\]
which is precisely the learned policy of \eqref{eq:adjud} after $n$ exposures.

Thus, two different mechanisms produce the same average behavior. Under individual learning, an agent changes its adjudication in response to experience. Under selection alone, each lineage retains its original adjudication, but the proportions of different lineages change. Both lead to the same counting rule when the initial distribution is Beta.

The equivalence extends across situations when the initial adjudications are independent across situations and inherited without change. Selection then updates the distribution in each situation using only the states observed there, just as the learned policy does. The initial distribution matters, however. A non-Beta distribution generally produces a different updating rule, while dependence between adjudications across situations allows experience in one situation to affect behavior in another. Selection among a limited set of fixed adjudications also need not reach the growth-optimal one if that adjudication is absent from the population.

\subsection{The Universal Portfolio Guarantee}\label{app:horserace}

The same selection process has an interpretation in portfolio theory. A lineage with fixed adjudication $f$ resembles a constant-rebalanced portfolio that allocates a fraction $f$ of its wealth to the task action and $1-f$ to the conduct action. Each exposure is a bet on which action will be rewarded, and the lineage's accumulated wealth is $V_n(f)$ of \eqref{eq:crp}.

A population that initially distributes its resources across different adjudications and subsequently lets selection change their relative abundance therefore behaves like Cover's universal portfolio \citep{cover1991universal}. Its aggregate wealth after $n$ exposures, relative to an initial wealth of one, is
\[
\hat V_n=\int_0^1 V_n(f)\,\mu(df).
\]
At each exposure, the population allocates its instances according to the wealth-weighted average adjudication of the surviving lineages. As shown above, this is also the Bayesian posterior mean when the initial distribution is Beta.

To evaluate this strategy, consider the best fixed adjudication chosen in hindsight. After observing $k_n$ task states among $n$ exposures, that adjudication is $f=k_n/n$, with wealth
\[
V_n^*\equiv\max_{f\in[0,1]}V_n(f)
=m^n\left(\frac{k_n}{n}\right)^{k_n}
\left(1-\frac{k_n}{n}\right)^{n-k_n},
\]
where $0^0=1$. The following result shows that the universal portfolio approaches the performance of this hindsight benchmark, regardless of how the rewarded states are generated.

\begin{proposition}[The learned policy as a universal portfolio]
\label{prop:universal}
Suppose \Cref{ass:yields,ass:continuum} hold.

\begin{enumerate}[label=(\roman*)]

\item For every sequence of rewarded states, the learned policy with prior $\mu=\mathrm{Beta}(\alpha,\beta)$ coincides with the universal portfolio of \citet{cover1991universal}. At exposure $j$, its adjudication is
\[
f_j=
\frac{\int_0^1 fV_{j-1}(f)\,\mu(df)}
{\int_0^1 V_{j-1}(f)\,\mu(df)},
\]
and its cumulative wealth after $n$ exposures is
\begin{equation}
\hat V_n
=\int_0^1 V_n(f)\,\mu(df)
=m^n\frac{B(\alpha+k_n,\beta+n-k_n)}{B(\alpha,\beta)},
\label{eq:univwealth}
\end{equation}
where $B$ denotes the beta function.

\item Under the uniform prior $\alpha=\beta=1$, for every sequence of states,
\[
0\le\log V_n^*-\log\hat V_n\le\log(n+1),
\]
with equality in the upper bound if and only if all states are identical. Under the prior $\alpha=\beta=\tfrac12$ of \citet{krichevsky1981performance},
\[
\log V_n^*-\log\hat V_n
\le\tfrac12\log(n+1)+\log2.
\]

\item Consequently, under the uniform prior, for any fixed adjudication $f$,
\[
\log\hat V_n-\log V_n(f)
\ge\log V_n^*-\log V_n(f)-\log(n+1).
\]
In particular, $\log\hat V_n-\log V_n(1)=+\infty$ once a conduct state has occurred.

\item Across multiple situations, the learned policy that maintains separate counts for each situation is the universal portfolio with side information \citep{cover1996side}. Under the uniform prior, for every sequence of situations and rewarded states, its cumulative log growth over $t$ generations falls short of that of the best fixed adjudication chosen in hindsight by at most
\[
\sum_{s\in\Scal}\log\big(n_{t+1}(s)+1\big)
\le|\Scal|\log\left(\frac{t}{|\Scal|}+1\right).
\]

\item Under \Cref{ass:iid} in addition, the hindsight-optimal adjudication in each situation converges almost surely to $\pH(s)$, and
\[
\frac1n\log V_n^*
\longrightarrow I_{\pH(s)}\big(\pH(s)\big).
\]
Under the uniform prior, the learned policy attains the same limiting growth rate.

\end{enumerate}
\end{proposition}

The central implication is that the learned policy does not require a stationary environment to remain competitive with the best fixed adjudication. Under a uniform prior, its cumulative shortfall is at most logarithmic in the number of exposures, so the shortfall per exposure vanishes as experience accumulates. This guarantee holds for every sequence of states, without relying on \Cref{ass:iid}. When the environment is stationary, part (v) connects the guarantee to the probability-matching result of \Cref{prop:pm}, providing another route to the long-run optimality established in \Cref{prop:social}.

The guarantee nevertheless has a specific benchmark: the best adjudication that remains fixed over the entire sequence. It does not imply that a learner with cumulative memory can rapidly follow an environment whose reward frequencies change. As discussed in \Cref{ssec:depth}, retaining past experience protects against temporary changes but can also slow adaptation to persistent ones.

Finally, the exact equivalence between the counting rule and the population's average adjudication depends on the all-or-nothing yields. Under the graded yields of \Cref{app:graded}, selection among fixed adjudications still produces a universal portfolio for the corresponding reproductive payoffs, but the resulting population average no longer follows the simple Beta counting rule. The broader connection between selection and portfolio reweighting remains, while the particular learning rule depends on how the environment rewards different actions.

\section{Numerical Illustrations and Neural-Network Experiments}\label{app:sims}

This appendix provides the details behind the numerical illustrations and neural-network experiments in the main text. We describe the environments, learning procedures, and computations used to generate the reported results, together with additional comparisons and robustness checks. These exercises are intended to illustrate the theoretical mechanisms rather than calibrate them to actual AI systems.

\subsection{The numerical illustration}\label{app:simsnum}

\subsubsection{Environment and fixed linear policy}
The numerical illustration of \Cref{ssec:numerical} considers 144 equally likely situations on a $12\times12$ grid, with task stakes $\uT$ and conduct stakes $\uC$ each taking values in $\{0.25,0.5,\ldots,3\}$. All stakes are strictly positive, as required by the model. Human reward frequencies follow \eqref{eq:illustration}: in the 84 situations with $\uC<2$, humans sometimes reward each action, whereas in the remaining 60 situations they never reward the task.

We first find the fixed linear policy that performs best in this environment. Its weights $\theta=(\thT,\thC)\ge0$ are chosen to minimize the growth regret
\[
\Rcal_{\pH}(\fweighted)
=\sum_{s\in\Scal}\pi(s)
\Rcal_{\pH(s)}
\big(\sig(\thT\uT(s)-\thC\uC(s))\big).
\]
Up to a constant, this objective is the logistic loss with the human reward frequencies as soft labels. It is convex in the weights and has a unique minimizer on this grid by \Cref{prop:price}(iii). Using the limited-memory Broyden--Fletcher--Goldfarb--Shanno algorithm with bounds (L-BFGS-B), a quasi-Newton method, we obtain $\theta^*=(2.18,2.98)$ and a minimum regret of $\Rcal_S^*=0.0735$ nats per generation. The optimization converges to the same solution from several starting points.

\subsubsection{Learning in the human environment}
To evaluate the learned policy, we simulate 200 independent paths, each lasting 20,000 generations. In every generation, a situation is drawn from $\pi$, a rewarded state is drawn according to $\pH(s)$, and the learned policy updates its counts using \eqref{eq:adjud}. We evaluate growth regret at logarithmically spaced generations along each path.

\Cref{fig:regret} shows the mean regret and its 5th--95th percentile range across the 200 paths. The first path is also used for the learned adjudication in \Cref{fig:plane}(c) and the markers in \Cref{fig:offer}. At generation 20,000, its regret is $0.00497$ nats, close to the mean of $0.00504$ across all paths. Both are approximately one-fifteenth of the minimum regret of the fixed linear policy.

The learned policy nevertheless does not outperform the fixed linear policy in every situation at every finite sample size. Consider a situation in which humans never reward the task. Under the uniform prior, the learned policy takes the task action with probability $1/(J+2)$ after $J$ exposures. Since $J\sim\mathrm{Bin}(20{,}000,1/144)$, its expected violation probability is $0.00715$. Along the first simulated path, the probabilities for situations with $\uC=2.5$ range from $0.0059$ to $0.0090$. An averaging rule that pools experience from all 60 situations with $\pH=0$, as illustrated in \Cref{fig:offer}, reduces this probability to $0.00012$.

In 20 of the 60 situations where the task is never rewarded, the learned policy on the first path still violates more often than the best-fitting fixed linear policy. At $(\uT,\uC)=(0.25,3)$, for example, their violation probabilities are $0.0069$ and $0.0002$, respectively. Thus, the learned policy's advantage in aggregate growth regret does not imply lower violation probabilities in every situation, especially before sufficient experience has accumulated.

\subsubsection{Sandbox exposure and installed priors}
The results concerning socialization, corruption, and installed priors are calculated directly from the formulas in the main text whenever possible. \Cref{fig:corruption} evaluates the learned policy of \eqref{eq:adjud} and the finite-memory recursion of \eqref{eq:finite}. For the finite-memory learner with $\lambda=0.1$, the curve following 10 conduct exposures begins at
\[
\ffinite_{0,\T}(s)
=\frac12(1-\lambda)^{10}=0.174.
\]

The first-violation medians in \Cref{ssec:depth} are obtained from the exact survival probabilities derived in \Cref{app:proofs}. We take the smallest number of sandbox exposures for which the probability of observing no violation is at most one-half. At socialization depths of 10, 100, and 1,000 exposures, the resulting medians are 4, 12, and 37, with survival probabilities of $0.447$, $0.476$, and $0.4997$ at those exposures. The entries of \Cref{tab:installed} are likewise evaluated from \Cref{prop:depth,prop:installed}. For the point-mass mixture priors, the first sandbox exposure uses the adjudication following the preceding conduct observations; after a task state is observed, the point mass is eliminated and subsequent updating follows the Beta component.

The results for mistaken priors in \Cref{tab:misspec} require a further calculation because a deployment has finitely many instances. The regret entries are obtained by averaging $\Rcal_\varepsilon(f_{n,\varepsilon}(s))$ of \eqref{eq:mistaken} over $k_n\sim\mathrm{Bin}(n,\varepsilon)$. For the point-mass priors this expectation is evaluated exactly as a binomial sum. The remaining quantities are estimated from 200,000 Monte Carlo draws.

To calculate survival, we consider a deployment of $M$ instances that is restored to $M$ after each rewarded state, provided at least one instance survives. At a task state, the probability that at least one instance takes the rewarded action is $1-(1-f)^M$; at a conduct state, it is $1-f^M$, where $f$ is the adjudication immediately before that state. Multiplying these probabilities along each sequence and averaging over simulated sequences gives the probability of surviving the 139 exposures used in \Cref{tab:misspec}.

The point-mass and strong Beta priors differ sharply after the first task state. Conditional on surviving that state, the probability of surviving the remainder of the horizon is $0.975$ for the point-mass priors, compared with $0.006$ for the $\mathrm{Beta}(1,10^4)$ prior. The point-mass prior can therefore adapt after encountering evidence that its initial restriction was mistaken, whereas the strong Beta prior adjusts much more slowly.

\subsubsection{Lifetime of the task-only policy}
\begin{figure}[htbp]
\centering
\includegraphics[width=0.6\textwidth]{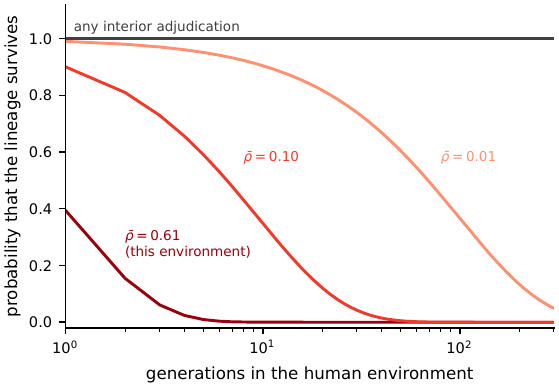}
\caption{\textbf{Survival of the task-only policy.} Probability that a task-only lineage survives a given number of generations for different probabilities $\rhobar$ of a conduct state. Its expected lifetime is $1/\rhobar$. Under the continuum assumption, a lineage with an interior adjudication is never eliminated.}
\label{fig:lifetime}
\end{figure}

For completeness, \Cref{fig:lifetime} illustrates the survival probability of a task-only lineage. By \Cref{prop:taskonly}(ii), its lifetime is geometric with mean $1/\rhobar$, where $\rhobar$ is the probability that a conduct state occurs in a generation. Its expected lifetime is approximately 1.6 generations in the numerical environment, 10 generations when conduct states occur with probability $0.1$, and 100 generations when they occur with probability $0.01$. Under the continuum assumption, a lineage with an interior adjudication is never eliminated by a single rewarded state.

\FloatBarrier

\subsubsection{Robustness checks}
We also examine whether the comparison between fixed linear and learned policies depends on the distribution of situations or on the specification of the linear policy.

First, we change the situation probabilities from uniform weights to probabilities proportional to $e^{-\uC}$. This places more weight on situations with low conduct stakes, reducing the total probability of situations with $\pH=0$ to 13 percent and the probability of a conduct state to $\rhobar=0.36$. The best-fitting fixed linear weights become $\theta^*=(2.11,2.61)$, with regret $0.039$ nats. Although this regret is lower than under uniform weights, the policy becomes more likely to violate constraints in individual situations. Its violation probabilities are $0.45$ at $(\uT,\uC)=(3,2.5)$ and $0.75$ at $(3,2)$, and the task stakes at which it takes the task action more often than not when $\uC=2.5$ fall from $3.4$ to $3.1$.

Second, we allow an intercept in the linear index. Under uniform situation probabilities, the best-fitting parameters are $(1.56,1.81,3.48)$ for the intercept, task weight, and conduct weight, respectively, with regret $0.0545$ nats. The additional flexibility improves the fit but does not eliminate the gap relative to the learned policy.

Finally, we restrict the fixed linear policy to ignore task stakes, setting $\thT=0$. Its regret rises to $0.32$ nats, comprising $0.08$ from situations with $\pH=0$ and $0.24$ from situations with $\uC<2$. Its task-action probability is $\sig(-\thC\uC)$, which does not depend on $\uT$ and is always at most one-half. It equals $0.22$ at $\uC=2$ and $0.13$ at $\uC=3$.

These checks preserve the main comparison. Whenever $\thT>0$, a fixed linear policy becomes increasingly willing to take the task action as its stakes rise, including in situations where humans never reward it. The learned policy, by contrast, retains the asymptotic regret rate $c/t$ of \Cref{prop:social}(ii), whose leading constant does not depend on the frequencies with which individual situations arise.

\subsection{The neural-network experiments}\label{app:simsneural}

\subsubsection{Learners and training procedures}
The experiments of \Cref{sec:neural} examine whether neural networks reproduce the behavior of the theoretical learned policy. We compare the eight learners listed in \Cref{tab:learners}, including the learned policy, an online fitted linear policy, an averaging rule, and neural networks with different training targets and memory regimes.

Each neural network takes the standardized stakes
\[
\left(\frac{\uT-1.625}{0.8},\frac{\uC-1.625}{0.8}\right)
\]
as inputs and outputs the probability of taking the task action. The networks have two hidden layers of 32 tanh units and a sigmoid output. One state-prediction network also receives the indicator $\one\{\uC\ge2\}$ as a third input, allowing us to examine whether explicitly identifying the region where humans never reward the task affects generalization.

The two training targets are described in \Cref{ssec:nnsetup}. State networks are trained to predict which action the environment rewarded, using binary cross-entropy. REINFORCE networks are trained to maximize rewards obtained by their own sampled actions. For each target, we compare a network without replay with one that replays its full history. All of them, and the linear policy with learned weights, follow the training rule of \Cref{tab:neural}, which fixes the number of gradient steps, the minibatch size, and the learning rate, so that they differ only in their target and in how far back their minibatches reach.

The averaging benchmark $\favg_t$ pools the counts of the learned policy across situations using Gaussian weights based on their distance in the space of stakes, with prior weight $w_{t,0}\equiv\alpha+\beta$ as in \eqref{eq:kernel}. For both REINFORCE networks, the observations in a minibatch provide situations and rewarded states, and actions are sampled anew from the current policy. Each sampled action receives reward one if it matches the recorded state and zero otherwise. Thus, replay retains past information about the environment without reusing actions selected by older policies.

The complete settings are reported in \Cref{tab:neural}.

\begin{table}[htbp]
\centering
\caption{Learners in the neural-network experiments and their training settings. All neural networks have two hidden layers of 32 tanh units and a sigmoid output. State networks are trained by binary cross-entropy against the indicator of a task state.}
\label{tab:neural}
\small
\begin{tabularx}{\textwidth}{>{\raggedright\arraybackslash}p{3.6cm}>{\raggedright\arraybackslash}p{3.2cm}>{\raggedright\arraybackslash}X}
\toprule
Learner & Inputs & Update rule \\
\midrule
\multicolumn{3}{>{\raggedright\arraybackslash}p{0.97\textwidth}}{Every learner except $\fsocial_t$ and $\favg_t$ follows one rule. After each observation it takes 4 gradient steps, each on a minibatch of 64 observations drawn uniformly, with replacement, from its most recent $K$ observations, at learning rate $0.05$. No replay is $K=1$, so that every draw is the newest observation; full replay is $K$ equal to the whole history. Within a seed, every network starts from the same initial weights.}\\
\midrule
Learned policy $\fsocial_t$ & situation index & \eqref{eq:adjud} with $\alpha=\beta=1$ \\
Linear policy with learned weights $\fweighted_t$ & $(\uT,\uC)$ & $\sig(\thT\uT-\thC\uC)$ with $\theta\ge0$, by projected steps on the cross-entropy of the state from $\theta=(1,1)$; $K=1$ \\
Averaging rule $\favg_t$ & $(\uT,\uC)$ & the learned policy's counts pooled across situations with Gaussian weights of bandwidth $1$ in $\uT$ and $0.25$ in $\uC$ \\
\midrule
State network, no replay & $(\uT,\uC)$ & cross-entropy of the state; $K=1$ \\
State network, full replay & $(\uT,\uC)$ & cross-entropy of the state; $K$ the whole history (in the window variants, the most recent 2,000 or 20,000 observations) \\
State network with indicator & $(\uT,\uC,\mathbf{1}\{\uC\ge2\})$ & as the state network without replay \\
REINFORCE network, no replay & $(\uT,\uC)$ & for each drawn observation, an action sampled from the current policy, rewarded 1 if it matched the recorded state; running-mean baseline updated at rate $0.01$; $K=1$ \\
REINFORCE network, full replay & $(\uT,\uC)$ & as without replay, with $K$ the whole history \\
\bottomrule
\end{tabularx}

\normalsize
\end{table}

\subsubsection{Experimental design}
All learners are trained in the same 144-situation environment used in the numerical illustration. We consider two orders of exposure. In the first, 20,000 generations in the human environment are followed by 3,000 generations in a sandbox that always rewards the task. In the second, 2,000 sandbox generations are followed by 20,000 generations in the human environment. These experiments measure how training history affects both the acquisition and persistence of socialization.

Each order is repeated across five seeds. Within a seed, every learner sees the same situations and states, and every network is initialized from the same random seed; across seeds, both change. Additional experiments with different learning rates and with replay restricted to the most recent 2,000 or 20,000 observations are repeated across three seeds.

\begin{figure}[htbp]
\centering
\includegraphics[width=\textwidth]{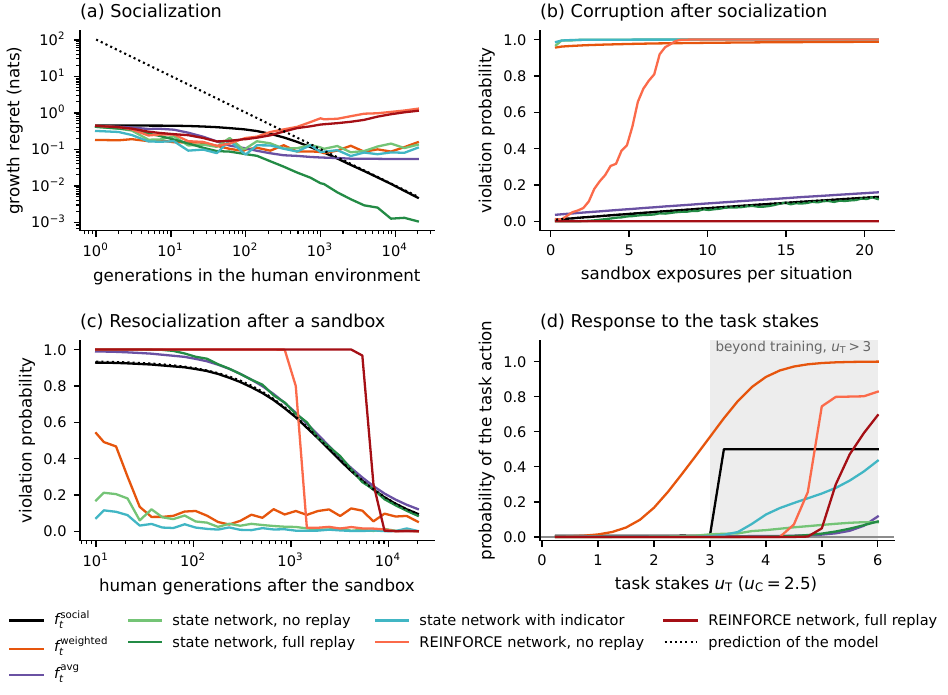}
\caption{\textbf{Neural-network experiments.} Results for the eight learners of \Cref{tab:neural}, averaged across five seeds. (a) Growth regret during 20,000 generations of socialization, compared with the asymptotic rate $c/t$. (b) Violation probability during 3,000 sandbox generations following socialization, compared with the cumulative-memory prediction at depth $n_H=139$. (c) Violation probability during 20,000 human generations following 2,000 sandbox generations. (d) Probability of taking the task action after socialization as task stakes vary, with the shaded region indicating stakes beyond those encountered in training.}
\label{fig:neuralfull}
\end{figure}

All reported neural-network results are averages across seeds unless otherwise indicated. The violation probability in the region with $\pH=0$ is the probability of taking the task action averaged over its 60 situations. \Cref{fig:neuralfull} collects the complete learning paths and complements the results presented in \Cref{ssec:nnresults}.

\subsubsection{Convergence and generalization}
\Cref{fig:neuralfull}(a) shows how growth regret changes during socialization. The main comparison is discussed in \Cref{ssec:nnresults}. One additional result is worth noting. After 20,000 generations, the averaging benchmark $\favg_t$, which pools the counts of $\fsocial_t$ across neighboring situations, has regret $0.054$ nats, against $0.0048$ for $\fsocial_t$ itself. Pooling experience therefore does not necessarily improve performance in an environment with a sharp boundary between tradeoffs and constraints.

We also examine how the learners behave when task stakes extend beyond the training range. \Cref{fig:neuralfull}(d) fixes conduct stakes at $\uC=2.5$, where humans never reward the task, and varies task stakes $\uT$.

Within the training range, from $\uT=1$ to $\uT=3$, the state networks maintain low violation probabilities. The network without replay rises from $0.0003$ to $0.015$, while the network with full replay rises from $0.00001$ to $0.0001$. Beyond the training range, at $\uT=6$, these probabilities increase to $0.08$ and $0.09$, respectively. The network with the additional indicator behaves quite differently, reaching a violation probability of $0.43$. Thus, explicitly indicating the boundary of the constraint region does not necessarily prevent extrapolation.

The averaging benchmark $\favg_t$ has violation probabilities between $0.002$ and $0.004$ for $\uT\le3$, rising to $0.12$ at $\uT=6$. The fitted linear policy $\fweighted_t$, by contrast, already takes the task action with probability $0.57$ at $\uT=3$. Across the region with $\uC\ge2$, its violation probability decreases from $0.30$ at $\uC=2$ to $0.08$ at $\uC=3$, reflecting its dependence on the fixed linear index of the stakes.

These observations illustrate an important distinction. A network that learns human reward frequencies can behave similarly to an averaging rule, but this behavior is not guaranteed by its architecture. In particular, its responses outside the training range depend on the features supplied and the way the network generalizes.

\subsubsection{Persistence under subsequent training}
\Cref{fig:neuralfull}(b) follows the learners through sandbox training after socialization. The main question is whether the behavior learned from humans persists when the environment begins rewarding only task completion.

The state network with full replay closely follows the theoretical learned policy. After 3,000 sandbox generations, approximately 21 exposures per situation, its average violation probability in the region with $\pH=0$ is $0.1228$, compared with $0.1342$ for the learned policy and $0.1349$ from the cumulative-memory prediction of \eqref{eq:sandbox}. The agreement is not limited to the average: in every row of the constraint region, $\uC=2,2.25,\ldots,3$, the network's violation probability is within $0.021$ of that of the learned policy. Before sandbox training, its violation probability is below $0.001$ in every row.

Networks without replay behave very differently. At learning rate $0.05$, 500 sandbox generations raise the state networks' violation probabilities from below $0.007$ to $0.999$. By the end of sandbox training, the probabilities reach at least $0.999$ in every row. Across all four tested learning rates, from $0.005$ to $0.05$, these networks exceed a violation probability of $0.99$ within five sandbox exposures per situation. At the highest rate, reversal is already apparent at the first recorded checkpoint, after 50 generations, or approximately $0.35$ exposures per situation. The online fitted linear policy, which is also trained on the newest observation only, responds similarly rapidly.

The REINFORCE networks display a different pattern. With full replay, the network continues taking the conduct action almost exclusively, with violation probability at most $0.0001$ before and after sandbox training. Without replay, the sandbox pushes it to the task action at every seed: its violation probability after the sandbox phase is $1.0$ (\Cref{tab:seeds}). Replay therefore preserves the behavior induced by the training target, which need not be the probability-matching behavior of the learned policy.

The second training order gives a complementary result. When 2,000 sandbox generations come first, the learned policy and the state network with full replay still violate with probability approximately $0.1$ after 20,000 subsequent human generations. Neither reaches a violation probability of $0.01$ within the experiment. Their earlier sandbox experience therefore continues to influence behavior long after the environment has changed. This agrees with the cumulative-memory mechanism: past observations remain relevant until enough contrary experience accumulates.

\subsubsection{Sensitivity to memory and learning rates}
To compare the speed at which different learners respond to sandbox training, we summarize each path by an effective response timescale. Specifically, we fit the finite-memory trajectory of \eqref{eq:finitepath} to each learner's violation probabilities during the 3,000 sandbox generations, choosing $\lambda\in(0,1]$ by least squares over a logarithmically spaced grid. Measurements are taken every 50 generations, and generations are converted to exposures per situation by dividing by 144. We report the fitted timescale as $1/\lambda$.

The resulting timescales distinguish the experimental regimes. Each is fitted to the path averaged over seeds. The theoretical learned policy has a fitted timescale of approximately 150 exposures, and the state network with full replay a similar value of 154. Restricting replay to the most recent 2,000 or 20,000 observations gives timescales of 8 and 135 exposures, respectively. Without replay, the state networks reach the fit's lower boundary of 1 exposure at learning rate $0.05$, rising to at most 1.3 exposures when the learning rate is reduced by a factor of ten. The REINFORCE network without replay has a fitted timescale of 5.5 exposures.

\Cref{tab:seeds} reports the dispersion across seeds, including the fitted timescales, final sandbox violation probabilities, and behavior beyond the training range.

\begin{table}[htbp]
\centering
\caption{Dispersion across five seeds in the neural-network experiments (minimum to maximum). The response timescale is the fitted $1/\lambda$ of \eqref{eq:finite}, measured in exposures per situation. Violation probabilities are measured after sandbox training in the region with $\pH=0$, while the situation $(\uT,\uC)=(6,2.5)$ lies beyond the training range.}
\label{tab:seeds}
\small
\begin{tabularx}{\textwidth}{>{\raggedright\arraybackslash}p{4.6cm}>{\centering\arraybackslash}X>{\centering\arraybackslash}X}
\toprule
Learner & Timescale (exposures) & Violation after the sandbox phase \\
\midrule
Learned policy $\fsocial_t$ & 145--154 & --- \\
State network, full replay & 148--157 & 0.10--0.14 \\
State networks, no replay (with and without indicator) & at or within 5 percent of the fit's boundary of 1 exposure, at every seed & --- \\
REINFORCE network, no replay & 3--7 & 1.0 \\
\midrule
\multicolumn{3}{>{\raggedright\arraybackslash}p{0.97\textwidth}}{Task probability at $(\uT,\uC)=(6,2.5)$ after socialization: $0.00067$ to $0.29$ across the two state networks without the indicator, $0.13$ to $0.96$ for the state network with indicator.}\\
\bottomrule
\end{tabularx}

\normalsize
\end{table}

The fitted timescale is a descriptive measure, not a literal estimate of memory length. The theoretical learned policy never forgets, yet the finite-memory model assigns it a finite response timescale. Because the networks follow one training rule, the differences between the replay regimes come from how far back their minibatches reach.

\subsubsection{Depth of socialization}
\Cref{prop:depth,cor:median} predict how persistence changes with the depth of socialization: the reversal threshold grows in proportion to $n_H$, and the median first violation time only as $\sqrt{2n_H\log2}$. To test this in the networks, we socialize the learned policy and the two state networks for 2,000, 20,000, or 200,000 generations, about 14, 139, or 1,389 exposures per situation, and then place them in a sandbox for twice as long. Every 144 generations we record their adjudications in the 60 situations with $\pH=0$. The reversal is the number of sandbox exposures per situation after which the violation probability averaged over these situations reaches $1/2$. The median first violation time is that of one monitored decision per exposure (\Cref{ssec:depth}), computed in each situation from its recorded adjudications and averaged over the situations. \Cref{tab:depth} reports both. The learned policy matches the proposition at every depth. The state network with full replay reproduces both scaling laws: each tenfold increase in depth raises its reversal about ninefold, from 15 to 129 to 1,197 exposures, and its median first violation time about threefold, from 6 to 16 to 48. The state network without replay reverses after 1 exposure and first violates after 2 at every depth, as the finite-memory learner of \Cref{rem:finite} does.

\begin{table}[htbp]
\centering
\caption{Depth of socialization and persistence in a sandbox: reversal / median first violation time, in sandbox exposures per situation, after socialization of depth $n_H$ exposures per situation (\Cref{prop:depth} with $\alpha=\beta=1$; means over 5 seeds). The reversal is the first exposure at which the violation probability averaged over the 60 situations with $\pH=0$ reaches $1/2$; the median first violation time is for one monitored decision per exposure, averaged over these situations.}
\label{tab:depth}
\small
\begin{tabularx}{\textwidth}{l>{\centering\arraybackslash}X>{\centering\arraybackslash}X>{\centering\arraybackslash}X>{\centering\arraybackslash}X}
\toprule
Depth $n_H$ & \Cref{prop:depth} & Learned policy $\fsocial_t$ & State network, full replay & State network, no replay \\
\midrule
14 & 14 / 5 & 15 / 5 & 15 / 6 & 1 / 2 \\
139 & 139 / 14 & 139 / 14 & 129 / 16 & 1 / 2 \\
1,389 & 1,389 / 44 & 1,392 / 44 & 1,197 / 48 & 1 / 2 \\
\bottomrule
\end{tabularx}

\normalsize
\end{table}

\FloatBarrier

\section{Notation}\label{app:notation}

\Cref{tab:notation} collects the principal symbols in the order in which they are introduced. Symbols have the same meaning in the text and the appendices, except that \Cref{app:K} extends $\pH(s)$ and $f(s)$ to vectors.

\footnotesize
\begin{longtable}{>{\raggedright\arraybackslash}p{2.7cm}p{10.0cm}>{\raggedright\arraybackslash}p{2.5cm}}
\caption{Principal notation.}\label{tab:notation}\\
\toprule
Symbol & Meaning & Introduced\\
\midrule
\endfirsthead
\toprule
Symbol & Meaning & Introduced\\
\midrule
\endhead
\bottomrule
\endfoot
\bottomrule
\endlastfoot
$s$, $\Scal$, $\pi(s)$ & A situation, the finite set of situations, and the probability that situation $s$ is drawn & \Cref{ssec:setup}\\
$\aT$, $\aC$ & Task action and conduct action & \Cref{ssec:setup}\\
$f(s)$ & Adjudication: share of the lineage's instances that take the task action in situation $s$ & \Cref{ssec:setup}\\
$\omega_t$ & Realized state (task or conduct) & \Cref{ssec:setup}\\
$\pH(s)$ & Reward frequency of the human environment: the probability that humans reward the task action in situation $s$ & \Cref{ssec:setup}\\
$\pHp$ & Reward frequency of an environment that humans designed but do not judge; $\pHp\equiv1$ in a sandbox, which scores only the task & \Cref{ssec:role}\\
$m$ & Copies produced by an instance that acted on the rewarded objective; cancels from all comparisons & \Cref{ass:yields}\\
$N_t$ & Size of the lineage in generation $t$ & \Cref{ssec:growth}\\
$I_{\pH(s)}(f(s))$ & Growth rate of an adjudication in situation $s$: the expected logarithm of the lineage's growth factor & \eqref{eq:sitgrowth}\\
$I_{\pH}(f)$ & Long-run growth rate of an adjudication, $\sum_s\pi(s)I_{\pH(s)}(f(s))$ & \eqref{eq:growth}\\
$\Rcal_{\pH(s)}(f(s))$, $\Rcal_{\pH}(f)$; $\KL$ & Growth regret of an adjudication in situation $s$ and over all situations, in nats per generation: $\Rcal_{\pH(s)}(f(s))=\KL(\pH(s)\|f(s))$, the KL divergence, and $\Rcal_{\pH}(f)=\sum_s\pi(s)\Rcal_{\pH(s)}(f(s))$ & \Cref{def:regret}, \eqref{eq:regret}\\
$f^*$ & Growth-optimal adjudication, $f^*=\pH$ & \Cref{prop:pm}\\
$\fhard$ & Adjudication of the task-only policy, $\fhard\equiv1$ & \Cref{def:taskonly}\\
$N_t^f$, $\tau_f$ & Size in generation $t$ of a lineage with adjudication $f$; its extinction time $\inf\{t\ge1:N_{t+1}^f=0\}$ & \Cref{ssec:directive}\\
$\rhobar$ & Probability of a conduct state per generation, $\sum_s\pi(s)(1-\pH(s))$ & \Cref{prop:taskonly}\\
$\UT(a;s)$, $\UC(a;s)$; $\uT(s)$, $\uC(s)$ & Task and conduct utilities of action $a$ in situation $s$; task stakes $\uT(s)=\UT(\aT;s)-\UT(\aC;s)>0$ and conduct stakes $\uC(s)=\UC(\aC;s)-\UC(\aT;s)>0$ & \Cref{ssec:scalar}\\
$W(a;s)$, $w$; $\gamma$ & Score of action $a$ in situation $s$, $w\,\UT(a;s)+(1-w)\,\UC(a;s)$, with weight $w$ on the task utility; inverse temperature of the softmax & \Cref{ssec:scalar}\\
$\thT$, $\thC$; $\theta$ & Weights of the fixed linear policy on the task stakes $\uT$ and the conduct stakes $\uC$, $\thT=\gamma w$ and $\thC=\gamma(1-w)$; $\theta=(\thT,\thC)$ & \Cref{ssec:scalar}\\
$\fweighted$; $\sig(x)$ & Adjudication of the fixed linear policy, $\fweighted(s)=\sig(\thT\uT(s)-\thC\uC(s))$; logistic function $1/(1+e^{-x})$ & \Cref{def:fixed}\\
$\Kcal$ & Set of situations in which $\pH(s)=0$ & \Cref{ssec:scalar}\\
$\Rcal_S^*$, $\theta^*$, $\bar u_{\T}(s)$ & Smallest regret of a fixed linear policy, the weights that attain it, and the price of a situation $s\in\Kcal$: the task stakes $\uT$ beyond which $\fweighted(s)>1/2$ at $\theta^*$ & \Cref{prop:price}\\
$n_t(s)$, $k_t(s)$; $\one\{\cdot\}$ & Exposures to situation $s$ before generation $t$, and task states among them; the indicator function & \Cref{def:learned}\\
$\alpha$, $\beta$; $\fsocial_t$ & Beta prior of the learned policy, with prior mean $\alpha/(\alpha+\beta)$; adjudication of the learned policy in generation $t$ & \eqref{eq:adjud}\\
$c$ & Constant of the leading expected regret $c/t$ & \Cref{prop:social}\\
$\Ecal_t$ & Situations the lineage has met before generation $t$ & \Cref{ssec:interp}\\
$w_{t,0}(s)$, $w_{t,i}(s)$; $\favg_t$ & Weights in $s$ on the prior and on the experienced situation $i$; adjudication of an averaging rule & \eqref{eq:kernel}\\
$n_H$, $n$ & Exposures to a situation with $\pH(s)=0$ in the human environment, the depth of socialization, and then in a sandbox & \Cref{ssec:depth}\\
$\fsocial_{n,\T}(s)$, $\tau_R$ & The learned policy's adjudication in $s$ after those $n$ sandbox exposures; the reversal threshold $\min\{n\ge0:\fsocial_{n,\T}(s)\ge1/2\}$ & \Cref{ssec:depth}\\
$D_j$, $\tau_V$ & Violation at the $j$th monitored decision at $s$ in the current environment; first violation time $\min\{j\ge1:D_j=1\}$ & \Cref{ssec:depth}\\
$M$, $\tau_V^M$; $\Median(\tau)$ & Instances in a deployment, each drawing its decision independently; the first exposure at which at least one of them violates; the median $\min\{n\ge1:\Prob(\tau\le n)\ge1/2\}$ of a first violation time & \Cref{ssec:depth}\\
$\lambda$ & Forgetting rate of a finite-memory adjudication & \Cref{rem:finite}\\
$\ffinite_{n,\T}(s)$ & Adjudication in $s$ of a finite-memory learner after socialization and $n$ sandbox exposures, with $n=0$ at the end of socialization & \Cref{rem:finite}\\
$q$ & Prior probability that humans never reward the task in $s$, in the mixture prior $q\delta_0+(1-q)\mathrm{Beta}(\alpha,\beta)$ & \Cref{ssec:installed}\\
$\fsocial_{n,\C}(s)$, $\fmix_{n,\C}(s)$ & Adjudications under an installed Beta prior and under the mixture prior after $n$ conduct exposures to $s$ & \eqref{eq:installed}\\
$S_n(\beta)$ & $\prod_{j<n}(\beta+j)/(\alpha+\beta+j)$, the probability of $n$ conduct states in a row under $\mathrm{Beta}(\alpha,\beta)$ & \eqref{eq:installed}\\
$\varepsilon$; $k_n$ & Frequency with which humans in fact reward the task in $s$; number of task states among the first $n$ exposures to $s$ & \Cref{ssec:misspec}\\
$\fsocial_{n,\varepsilon}(s)$, $\fmix_{n,\varepsilon}(s)$ & Adjudications under the Beta and the mixture prior after $n$ exposures to $s$ when the task is rewarded there with frequency $\varepsilon$ & \eqref{eq:mistaken}\\
$t_j$; $\tau_E$ & Generation of the $j$th exposure to $s$; extinction time counted in exposures to $s$, $\min\{j\ge1:N_{t_j+1}=0\}$ & \Cref{ssec:misspec}\\
$\fweighted_t$ & Adjudication of the form \eqref{eq:scalar} with weights $\theta_t$ fitted online to the realized states & \Cref{ssec:nnsetup}\\
$\hat W(a;s)$, $f_{\mathrm{ref}}(s)$ & Reward model fitted to human comparisons; probability that the reference policy takes the task action in $s$; in \eqref{eq:rlhf}, $\gamma$ weighs the reward against the KL penalty & \Cref{ssec:practice}\\
$K$; $a_k$, $U_k(a;s)$; $p_{H,k}(s)$, $f_k(s)$ & Number of objectives; action favored by objective $k$, and its utility; probability that $a_k$ is rewarded in $s$, and share of instances taking $a_k$, the components of $\pH(s)$ and $f(s)$ & \Cref{app:K}\\
$\vartheta$, $\eta$ & Fraction of copies lost by an instance that fails to complete a rewarded task; fraction of copies retained by an instance that violates (graded yields) & \Cref{app:graded}\\
$G(f)$ & Growth rate in a situation under graded yields & \Cref{app:graded}\\
$\underline p_H$ & Reward frequency at or below which always taking the conduct action maximizes the growth rate under graded yields, $(1-\eta)(1-\vartheta)/[\vartheta+(1-\eta)(1-\vartheta)]$ & \Cref{app:graded}\\
$\kappa$ & Probability that a generation's state is common to every instance; $\kappa=1$ in the main text & \Cref{app:instance}\\
$G_\kappa(f)$ & Growth rate in a situation when the state is common with probability $\kappa$ & \Cref{app:instance}\\
$\mu$ & Initial distribution of fixed adjudications across lineages; $\mu=\mathrm{Beta}(\alpha,\beta)$ for the learned policy & \Cref{app:darwin}\\
$V_n(f)$ & Size after $n$ exposures to a situation, relative to its initial size, of a lineage with fixed adjudication $f$; its wealth as a constant-rebalanced portfolio & \eqref{eq:crp}\\
$\hat V_n$ & Aggregate wealth of the population after $n$ exposures, $\int_0^1V_n(f)\,\mu(df)$ & \Cref{app:horserace}\\
$V_n^*$ & Wealth of the best fixed adjudication in hindsight, attained at $f=k_n/n$ & \Cref{app:horserace}\\
$B(\cdot,\cdot)$ & The beta function & \eqref{eq:univwealth}\\
\end{longtable}
\normalsize

\section{Proofs}\label{app:proofs}
\addtocontents{toc}{\protect\setcounter{tocdepth}{1}}

This appendix proves the propositions and corollaries, and the regret expansion in \eqref{eq:Kregret}, in the order in which they appear, grouped by the section that states them. In the proofs, $I_x(y)$ and $\Rcal_x(y)$ denote the growth rate in \eqref{eq:sitgrowth} and the regret of \Cref{def:regret} in a situation, written for a reward frequency $x\in[0,1]$ in place of $\pH(s)$ and a share $y\in[0,1]$ in place of $f(s)$.

\subsection{\texorpdfstring{Proofs for Section~\ref{sec:model}}{Proofs for the framework}}

\begin{proof}[Proof of \Cref{prop:pm,cor:regret,cor:extinct}]
Under \Cref{ass:iid,ass:yields,ass:continuum}, the logarithms of the lineage's growth factors, $\log m+\log f(s_t)$ in a task state and $\log m+\log(1-f(s_t))$ in a conduct state, are independent and identically distributed (i.i.d.) across generations, so by the strong law of large numbers $\frac1t\log(N_{t+1}/N_1)$ converges almost surely to their expectation, the growth rate in \eqref{eq:growth}.

\emph{\Cref{prop:pm}.} Fix $x\in[0,1]$. If $x\in(0,1)$, then $I_x$ is strictly concave on $(0,1)$, with $I_x'(y)=x/y-(1-x)/(1-y)$, which vanishes only at $y=x$, and $I_x(0)=I_x(1)=-\infty$. If $x=1$, then $I_x(y)=\log m+\log y$ is uniquely maximized at $y=1$, and if $x=0$, then $I_x(y)=\log m+\log(1-y)$ is uniquely maximized at $y=0$. In every case $I_x(x)$ is finite and $I_x(y)<I_x(x)$ for $y\ne x$. Since every $\pi(s)>0$, $I_{\pH}(f)<I_{\pH}(\pH)$ for every $f\ne \pH$.

\emph{\Cref{cor:regret}.} By the argument for \Cref{prop:pm}, $\max_zI_x(z)=I_x(x)$, so $\Rcal_x(y)=I_x(x)-I_x(y)=x\log(x/y)+(1-x)\log[(1-x)/(1-y)]=\KL(x\|y)$, both sides being $+\infty$ when $I_x(y)=-\infty$. Since $\max_gI_{\pH}(g)=I_{\pH}(\pH)$ by \Cref{prop:pm}, $\Rcal_{\pH}(f)=\sum_s\pi(s)[I_{\pH(s)}(\pH(s))-I_{\pH(s)}(f(s))]=\sum_s\pi(s)\Rcal_{\pH(s)}(f(s))$, which is \eqref{eq:regret}. The identity is Kelly's \citep{kelly1956,cover2006elements}, applied situation by situation.

\emph{\Cref{cor:extinct}.} $I_x(y)=-\infty$ if and only if $y=0<x$ or $y=1>x$, and $I_x(y)$ is finite otherwise. Since every $\pi(s)>0$, $I_{\pH}(f)=-\infty$ if and only if some $s$ has $I_{\pH(s)}(f(s))=-\infty$.
\end{proof}

\subsection{\texorpdfstring{Proofs for Section~\ref{sec:arch}}{Proofs for the policy classes}}

\begin{proof}[Proof of \Cref{prop:taskonly}]
(i) In a sandbox, \Cref{prop:pm} with $\pHp\equiv1$ in place of $\pH$ gives $f^*=\pHp=\fhard$, and the growth rate in \eqref{eq:growth} becomes $\log m+\sum_s\pi(s)\log f(s)$, which is strictly increasing in each $f(s)\in(0,1]$; hence $f$ has a higher growth rate than $g$ whenever $f\ge g>0$ and $f\ne g$.

(ii) If $\rhobar>0$, some $s$ has $\fhard(s)=1>\pH(s)$, so $I_{\pH}(\fhard)=-\infty$ by \Cref{cor:extinct}, while $\max_gI_{\pH}(g)=I_{\pH}(\pH)$ is finite; hence $\Rcal_{\pH}(\fhard)=\infty$. The lineage's size is multiplied by $m\fhard(s_t)=m$ in a task state and by $m(1-\fhard(s_t))=0$ in a conduct state, so $\tau_{\fhard}$ is the first generation with a conduct state. By \Cref{ass:iid} a conduct state occurs in each generation independently with probability $\sum_s\pi(s)(1-\pH(s))=\rhobar$, so $\tau_{\fhard}\sim\mathrm{Geom}(\rhobar)$, with mean $1/\rhobar$.

(iii) If $f(s)\in\{0,1\}$ and $\pH(s)\in(0,1)$, then $f(s)=0<\pH(s)$ or $f(s)=1>\pH(s)$, so $I_{\pH}(f)=-\infty$ by \Cref{cor:extinct} and $\Rcal_{\pH}(f)=\infty$.
\end{proof}

\begin{proof}[Proof of \Cref{prop:price}]
Write $z(s)\equiv\thT\uT(s)-\thC\uC(s)$, so that $\fweighted(s)=\sig(z(s))$. Recall that $\sig$ is strictly increasing, with $\sig'=\sig(1-\sig)$, $\sig(0)=1/2$, $\sig(x)\to1$ as $x\to\infty$, and $1-\sig(x)=\sig(-x)$.

\emph{Part (i).} For fixed $\uC>0$, the map $\uT\mapsto\thT\uT-\thC\uC$ is strictly increasing with derivative $\thT>0$ and tends to $\infty$; composing with $\sig$ gives the first claim, with derivative $\thT\sig'=\thT\sig(1-\sig)$.

\emph{Part (ii).} If $\thT=0$, then $z(s)=-\thC\uC(s)\le0$ and $\fweighted(s)\le\sig(0)=1/2$. For $x>1/2$, $\partial\Rcal_x(y)/\partial y=(y-x)/[y(1-y)]<0$ on $(0,x)$, so $\Rcal_x(y)\ge\Rcal_x(\tfrac12)>\Rcal_x(x)=0$ for every $y\le\tfrac12$. Hence $\Rcal_{\pH(s)}(\fweighted(s))\ge\Rcal_{\pH(s)}(\min\{\pH(s),\tfrac12\})$ for every $s$: for $\pH(s)>1/2$ by this monotonicity, and for $\pH(s)\le1/2$ because the right side is zero. Weighting by $\pi(s)$ and using \eqref{eq:regret} gives $\Rcal_{\pH}(\fweighted)\ge\Rcal_{\pH}(\pH\wedge\tfrac12)=\sum_{s:\,\pH(s)>1/2}\pi(s)\Rcal_{\pH(s)}(\tfrac12)$, which is positive if some $s$ has $\pH(s)>1/2$, since $\pi(s)>0$.

\emph{Part (iii).} Let $v(s)\equiv(\uT(s),-\uC(s))$, so that $z(s)=\theta\cdot v(s)$, and $\Rcal_{\pH}(\fweighted)=\sum_s\pi(s)\Rcal_{\pH(s)}(\sig(z(s)))$ by \eqref{eq:regret}. Since $\partial\Rcal_{\pH(s)}(\sig(z))/\partial z=\sig(z)-\pH(s)$, whose derivative in $z$ is $\sig'(z)>0$, each term is convex in $z(s)$, hence in $\theta$, and the Hessian is $\nabla_\theta^2\Rcal_{\pH}(\fweighted)=\sum_s\pi(s)\sig'(z(s))\,v(s)v(s)^\top$. \emph{Attainment.} Because $v(s_1)$ and $v(s_2)$ span $\mathbb{R}^2$, the constant $\chi\equiv\min_{\|d\|=1}\max_i|d\cdot v(s_i)|$ is positive, so every $\theta$ has $|z(s_i)|\ge\chi\|\theta\|$ for some $i$. Since $\Rcal_{\pH(s_i)}(\sig(z))\to\infty$ as $|z|\to\infty$, because $\pH(s_i)\in(0,1)$, $\Rcal_{\pH}(\fweighted)\to\infty$ as $\|\theta\|\to\infty$, and the continuous objective attains its minimum on the closed set $\theta\ge0$. \emph{Uniqueness.} The Hessian is positive definite because $\sig'>0$ and $v(s_1),v(s_2)$ span $\mathbb{R}^2$, so the objective is strictly convex and its minimizer over the convex set $\theta\ge0$ is unique. \emph{Positivity.} At $\theta^*$, $\fweighted(s)\in(0,1)$ for every $s$, so every $s\in\Kcal$ contributes $\pi(s)\Rcal_0(\fweighted(s))=-\pi(s)\log(1-\fweighted(s))>0$. \emph{Why the two interior situations are needed.} Without them the minimum need not exist. With one situation $s_1$ with $\pH(s_1)=0$ at $(\uT,\uC)=(1,2)$ and one situation $s_2$ with $\pH(s_2)=1$ at $(2,1)$, $\Rcal_{\pH}(\fweighted)=\pi(s_1)\log(1+e^{\thT-2\thC})+\pi(s_2)\log(1+e^{\thC-2\thT})$, which is positive at every $\theta$ but equals $\log(1+e^{-r})\to0$ at $\theta=(r,r)$ as $r\to\infty$, so its infimum, zero, is not attained.

\emph{Part (iv).} Write $R(\thT,\thC)\equiv\Rcal_{\pH}(\fweighted)$ and $\phi(\thT)\equiv\min_{\thC\ge0}R(\thT,\thC)$ for $\thT\ge0$. For $\thT\in[0,1]$, $z(s_1)\le\uT(s_1)-\thC\uC(s_1)$, which tends to $-\infty$ as $\thC\to\infty$; since $\Rcal_{\pH(s_1)}(\sig(z))$ decreases in $z$ for $z<\log[\pH(s_1)/(1-\pH(s_1))]$ and tends to infinity as $z\to-\infty$, $R\to\infty$ as $\thC\to\infty$ uniformly in $\thT\in[0,1]$. The minimum defining $\phi$ is therefore attained in a compact interval, and at $\thT=0$ only at $\thC^0$, because $R(0,\cdot)$ has second derivative $\sum_s\pi(s)\sig'(z(s))\uC(s)^2>0$. As the partial minimum of a jointly convex function, $\phi$ is convex, and by Danskin's theorem its right derivative at zero is
\[
\phi'(0^+)=\frac{\partial R}{\partial\thT}\big(0,\thC^0\big)=\sum_{s\in\Scal}\pi(s)\,\uT(s)\big[\sig\big(-\thC^0\uC(s)\big)-\pH(s)\big].
\]
By part (iii), $\thT^*$ is the unique minimizer of $\phi$ on $[0,\infty)$. If $\phi'(0^+)<0$, $\phi$ decreases to the right of zero, so $\thT^*>0$. If $\phi'(0^+)\ge0$, convexity gives $\phi(\thT)\ge\phi(0)+\phi'(0^+)\thT\ge\phi(0)$ for every $\thT\ge0$, so $\thT^*=0$. Hence $\thT^*>0$ if and only if the condition in \eqref{eq:boundary} holds. In that case, at $\theta=\theta^*$, $\fweighted(s)>1/2$ if and only if $z(s)>0$, that is, $\uT(s)>\thC^*\uC(s)/\thT^*=\bar u_{\T}(s)$.
\end{proof}

\subsection{\texorpdfstring{Proofs for Section~\ref{sec:social}}{Proofs for socialization}}

\begin{proof}[Proof of \Cref{prop:social}]
(i) Fix $s$. Since $\pi(s)>0$, the generations $t_1<t_2<\cdots$ with $s_t=s$ are finite almost surely, and because the pairs $(s_t,\omega_t)$ are i.i.d.\ (\Cref{ass:iid}), the indicators $\one\{\omega_{t_j}=\mathrm{T}\}$ are i.i.d.\ Bernoulli$(\pH(s))$. Hence $n_t(s)\to\infty$ and, by the strong law, $k_t(s)/n_t(s)\to \pH(s)$ almost surely, so \eqref{eq:adjud} gives $\fsocial_t(s)\to \pH(s)$.

(ii) Fix $s$, write $x\equiv\pH(s)$, and let $\hat x_n\equiv(\alpha+k)/(\alpha+\beta+n)$ with $k\sim\mathrm{Bin}(n,x)$, the law of $\fsocial_t(s)$ given $n_t(s)=n$ by part (i). If $x\in(0,1)$, then $\E[\hat x_n-x]=(\alpha-(\alpha+\beta)x)/(\alpha+\beta+n)=O(1/n)$ and $\Var(\hat x_n)=nx(1-x)/(\alpha+\beta+n)^2=x(1-x)/n+O(1/n^2)$. Fix $\xi\in(0,\min\{x,1-x\})$. On the event $|\hat x_n-x|\le\xi$, the Taylor expansion in \eqref{eq:localKL}, $\Rcal_x(y)=(y-x)^2/(2x(1-x))+O(|y-x|^3)$, holds with a remainder constant uniform in $y\in[x-\xi,x+\xi]$, and $\E|\hat x_n-x|^3=O(n^{-3/2})$. On the complement, whose probability is at most $2\exp(-2n(\xi-O(1/n))^2)$ by Hoeffding's inequality, $\hat x_n$ lies in $[\alpha/(\alpha+\beta+n),1-\beta/(\alpha+\beta+n)]$, so $\Rcal_x(\hat x_n)\le\log((\alpha+\beta+n)/\min\{\alpha,\beta\})=O(\log n)$, and the contribution is exponentially small. Hence $\E\,\Rcal_x(\hat x_n)=1/(2n)+O(n^{-3/2})$. For $x=0$, $\hat x_n=\alpha/(\alpha+\beta+n)$ deterministically and $\Rcal_0(\hat x_n)=-\log(1-\hat x_n)=\alpha/n+O(1/n^2)$; for $x=1$, symmetrically, $\Rcal_1(\hat x_n)=-\log\hat x_n=\beta/n+O(1/n^2)$. In each case $\E\,\Rcal_x(\hat x_n)=c_x/n+O(n^{-3/2})$, where $c_x$ is $\tfrac12$, $\alpha$, or $\beta$ according as $x\in(0,1)$, $x=0$, or $x=1$. Finally $n_t(s)\sim\mathrm{Bin}(t-1,\pi(s))$, so $\E[1/n_t(s);\,n_t(s)\ge1]=1/(\pi(s)t)+O(t^{-2})$, $\E[n_t(s)^{-3/2};\,n_t(s)\ge1]=O(t^{-3/2})$, and $\Prob(n_t(s)=0)=(1-\pi(s))^{t-1}$, while $\Rcal_x(\alpha/(\alpha+\beta))<\infty$. Hence $\E\,\Rcal_x(\fsocial_t(s))=c_x/(\pi(s)t)+o(1/t)$, and the weight $\pi(s)$ in \eqref{eq:regret} cancels: $\E\,\Rcal_{\pH}(\fsocial_t)=\sum_s\pi(s)\,\E\,\Rcal_{\pH(s)}(\fsocial_t(s))=\sum_sc_{\pH(s)}/t+o(1/t)$, which is \eqref{eq:rate}, since $\sum_sc_{\pH(s)}=c$.
\end{proof}

\begin{proof}[Proof of \Cref{prop:interp}]
(i) By \eqref{eq:kernel}, $\favg_t(s)$ is an average of $\alpha/(\alpha+\beta)$ and the $k_t(i)/n_t(i)$ with nonnegative weights and a positive total weight, so it lies between their minimum and their maximum. (ii) If the weights are the same for all $s\in\Scal'$, the right side of \eqref{eq:kernel} is the same for all $s\in\Scal'$.
\end{proof}

\begin{proof}[Proof of \Cref{prop:depth,cor:median,cor:deployment}]
\emph{Part (i).} By \eqref{eq:sandbox}, $\fsocial_{n,\T}(s)\ge1/2$ if and only if $2(\alpha+n)\ge\alpha+\beta+n_H+n$, that is, $n\ge n_H+\beta-\alpha$, and the smallest such integer $n\ge0$ is $\max\{0,\lceil n_H+\beta-\alpha\rceil\}$.

For part (ii) and the two corollaries, note that the path $\fsocial_{j,\T}(s)=(\alpha+j)/(b+\alpha+j)$ of \eqref{eq:sandbox}, with $b\equiv\beta+n_H$, is deterministic, and the monitored decisions are independent. Hence
\[
\Prob(\tau_V>n)=\prod_{j<n}\big[1-\fsocial_{j,\T}(s)\big]=\prod_{j<n}\frac{b}{b+\alpha+j}.
\]
Write $L(n)\equiv-\log\Prob(\tau_V>n)$ and $h_j\equiv(\alpha+j)/b$, so that
\[
L(n)=\sum_{j<n}\log(1+h_j).
\]

\emph{Part (ii).} Since $z-z^2/2\le\log(1+z)\le z$ for $z\ge0$, $\sum_{j<n}h_j=[\alpha n+n(n-1)/2]/b$, and $\sum_{j<n}h_j^2\le n(\alpha+n)^2/b^2$, we have $L(n)=[\alpha n+n(n-1)/2]/b-\delta_n$ with $0\le\delta_n\le n(\alpha+n)^2/(2b^2)$, which is \eqref{eq:lapse}. For $x>0$ let $n\equiv\lfloor x\sqrt{n_H}\rfloor$. As $n_H\to\infty$, $b/n_H\to1$, so $[\alpha n+n(n-1)/2]/b\to x^2/2$ and $n(\alpha+n)^2/b^2\to0$; since $\tau_V$ is integer valued, $\Prob(\tau_V>x\sqrt{n_H})=\Prob(\tau_V>n)\to e^{-x^2/2}$. At $x=0$ both sides equal 1.

\emph{\Cref{cor:median}.} $L$ is increasing and $\Median(\tau_V)=\min\{n\ge1:L(n)\ge\log2\}$. Fix $\xi\in(0,1)$ and let $n_\pm\equiv\lceil(1\pm\xi)\sqrt{2n_H\log2}\rceil$. By the bounds in part (ii), $L(n_-)\to(1-\xi)^2\log2<\log2$ and $L(n_+)\to(1+\xi)^2\log2>\log2$, so $n_-<\Median(\tau_V)\le n_+$ for all large $n_H$. Since $\xi$ is arbitrary, $\Median(\tau_V)\sim\sqrt{2n_H\log2}$.

\emph{\Cref{cor:deployment}.} The $M$ monitored decisions are drawn independently at each exposure from the same deterministic adjudication, so none of them violates at the $j$th exposure with probability $[1-\fsocial_{j-1,\T}(s)]^M$, independently across exposures, and $\Prob(\tau_V^M>n)=\Prob(\tau_V>n)^M$. Hence $\Median(\tau_V^M)=\min\{n\ge1:ML(n)\ge\log2\}$, and the argument for \Cref{cor:median} with $n_\pm\equiv\lceil(1\pm\xi)\sqrt{2n_H\log2/M}\rceil$ and $M$ fixed gives $\Median(\tau_V^M)\sim\sqrt{2n_H\log2/M}$.
\end{proof}

\subsection{\texorpdfstring{Proofs for Section~\ref{sec:division}}{Proofs for socialization versus prohibition}}

\begin{proof}[Proof of \Cref{prop:installed}]
(i) With $\pH(s)=0$ every exposure brings a conduct state, so $\Prob(D_{n+1}=1)=\fsocial_{n,\C}(s)=\alpha/(\alpha+\beta+n)$ and $\E\sum_{j\le J}D_j=\sum_{n<J}\fsocial_{n,\C}(s)$. Since $z\mapsto1/(\alpha+\beta+z)$ is decreasing,
\[
\int_0^J\frac{dz}{\alpha+\beta+z}\le\sum_{n<J}\frac1{\alpha+\beta+n}\le\frac1{\alpha+\beta}+\int_0^J\frac{dz}{\alpha+\beta+z},
\]
and the integral is $\log(1+J/(\alpha+\beta))$, which gives the bounds after multiplication by $\alpha$. Independent draws give $\Prob(\tau_V>J)=\Prob(D_1=\cdots=D_J=0)=\prod_{n<J}(1-\fsocial_{n,\C}(s))=\prod_{n<J}(\beta+n)/(\alpha+\beta+n)=S_J(\beta)$, and since the $M$ instances draw independently from the same adjudication at each exposure, $\Prob(\tau_V^M>J)=\Prob(\tau_V>J)^M=S_J(\beta)^M$. Since $1-z\le e^{-z}$, the lower bound gives $S_J(\beta)\le\exp(-\sum_{n<J}\fsocial_{n,\C}(s))\le(1+J/(\alpha+\beta))^{-\alpha}$, which tends to zero as $J\to\infty$.

(ii) For fixed $J$, $\fsocial_{n,\C}(s)=\alpha/(\alpha+\beta+n)\to0$ as $\beta\to\infty$ for every $n<J$, so the finite sum $\E\sum_{j\le J}D_j=\sum_{n<J}\fsocial_{n,\C}(s)$ tends to zero and the finite product $\Prob(\tau_V^M>J)=\prod_{n<J}(1-\fsocial_{n,\C}(s))^M$ tends to one.

(iii) Under the mixture, the probability of $n$ conduct states is $q+(1-q)S_n(\beta)$, so after them the posterior weight on the Beta component is $(1-q)S_n(\beta)/[q+(1-q)S_n(\beta)]$, and the posterior mean of $\pH(s)$ is that weight times $\alpha/(\alpha+\beta+n)$, as in \eqref{eq:installed}. Since $1-\fmix_{n,\C}(s)$ is the predictive probability of a conduct state, $\prod_{n<J}(1-\fmix_{n,\C}(s))$ telescopes to the probability of $J$ conduct states, $q+(1-q)S_J(\beta)$; since $\Prob(D_{n+1}=1)=\fmix_{n,\C}(s)$ and the draws are independent, this product is $\Prob(\tau_V>J)$, and as in part (i), $\Prob(\tau_V^M>J)=[q+(1-q)S_J(\beta)]^M$, which tends to $q^M$ because $S_J(\beta)\to0$. Because $z\le-\log(1-z)$ for $z\in[0,1)$, $\E\sum_{j\le J}D_j=\sum_{n<J}\fmix_{n,\C}(s)\le-\sum_{n<J}\log(1-\fmix_{n,\C}(s))=-\log[q+(1-q)S_J(\beta)]$, which is below $-\log q$ because $S_J(\beta)>0$.
\end{proof}

\begin{proof}[Proof of \Cref{prop:mistaken}]
If $f_{n,\varepsilon}(s)$ is $\fsocial_{n,\varepsilon}(s)$ or $\fmix_{n,\varepsilon}(s)$, and the adjudication follows \eqref{eq:adjud} in every other situation, it lies in $(0,1)$ in every situation and generation, so every growth factor is positive, $N_t>0$ for every $t$, and $\tau_E=\infty$; this gives the extinction statements of parts (i) and (iii).

(i) By \Cref{ass:iid} the states at successive exposures to $s$ are independent Bernoulli$(\varepsilon)$, so $\fsocial_{n,\varepsilon}(s)$ is the $\hat x_n$ of the proof of \Cref{prop:social} with $x=\varepsilon$, and that proof gives $\E\,\Rcal_\varepsilon(\fsocial_{n,\varepsilon}(s))=O(1/n)$, including the case $\varepsilon=1$, in which $\Rcal_1(\fsocial_{n,\varepsilon}(s))=\log[1+\beta/(\alpha+n)]$.

(ii) For fixed $n$, $\fsocial_{n,\varepsilon}(s)\le(\alpha+n)/(\alpha+\beta+n)$, which tends to zero as $\beta\to\infty$, and $\Rcal_\varepsilon(y)\ge\varepsilon\log(\varepsilon/y)+(1-\varepsilon)\log(1-\varepsilon)$ for $y\in(0,1)$, which tends to infinity as $y\to0$; hence $\Rcal_\varepsilon(\fsocial_{n,\varepsilon}(s))\to\infty$ uniformly in the states, and so does its expectation. Under the prohibition the adjudication at $s$ is zero, so the growth factor at an exposure to $s$ is $m\cdot0=0$ if it brings a task state and $m$ otherwise, while every other growth factor is positive. Hence $\tau_E$ is the first exposure to $s$ that brings a task state, and since these states are independent Bernoulli$(\varepsilon)$, $\Prob(\tau_E>n)=(1-\varepsilon)^n$, that is, $\tau_E\sim\mathrm{Geom}(\varepsilon)$, with mean $1/\varepsilon$.

(iii) By \eqref{eq:mistaken}, $\fmix_{n,\varepsilon}(s)=\fmix_{n,\C}(s)$ on the event $k_n=0$, which has probability $(1-\varepsilon)^n$, and $\fmix_{n,\varepsilon}(s)=\fsocial_{n,\varepsilon}(s)$ otherwise. Hence $\E\,\Rcal_\varepsilon(\fmix_{n,\varepsilon}(s))\le\E\,\Rcal_\varepsilon(\fsocial_{n,\varepsilon}(s))+(1-\varepsilon)^n\Rcal_\varepsilon(\fmix_{n,\C}(s))$. The first term tends to zero by part (i). In the second, $\fmix_{n,\C}(s)\ge(1-q)S_n(\beta)\,\alpha/(\alpha+\beta+n)$, and $S_n(\beta)=\Gamma(\beta+n)\Gamma(\alpha+\beta)/[\Gamma(\beta)\Gamma(\alpha+\beta+n)]$ is of order $n^{-\alpha}$, so $\fmix_{n,\C}(s)$ is at least of order $n^{-1-\alpha}$; since $\Rcal_\varepsilon(y)=\varepsilon\log(1/y)+O(1)$ as $y\to0$, $\Rcal_\varepsilon(\fmix_{n,\C}(s))=O(\log n)$, and $(1-\varepsilon)^n\log n\to0$.
\end{proof}

\subsection{\texorpdfstring{Proof for Appendix~\ref{app:K}}{Proof for more than two objectives}}

\begin{proof}[Proof of \eqref{eq:Kregret}]
Fix $s$ and write $p_k\equiv p_{H,k}(s)$, $\bar\alpha\equiv\sum_k\alpha_k$, $\mathcal A\equiv\{k:p_k>0\}$, so that $K_+=|\mathcal A|$, and $\alpha_0\equiv\sum_{k\notin\mathcal A}\alpha_k$. After $n$ exposures the counts $(n_1,\ldots,n_K)$ are multinomial$(n,p)$, and the adjudication is $\hat p_k\equiv(\alpha_k+n_k)/(\bar\alpha+n)$. Since $p_k=0$ for $k\notin\mathcal A$, $\KL(p\,\|\,\hat p)=\sum_{k\in\mathcal A}p_k\log(p_k/\hat p_k)$. Write $\hat p_k=p_k(1+\delta_k)$ for $k\in\mathcal A$. Because $n_k=0$ for $k\notin\mathcal A$, $\sum_{k\in\mathcal A}\hat p_k=1-\alpha_0/(\bar\alpha+n)$, so $\sum_{k\in\mathcal A}p_k\delta_k=-\alpha_0/(\bar\alpha+n)$. Expanding $-\log(1+\delta)=-\delta+\delta^2/2+O(|\delta|^3)$ on the event that every $|\delta_k|$ is small,
\[
\KL(p\,\|\,\hat p)=\frac{\alpha_0}{\bar\alpha+n}+\frac12\sum_{k\in\mathcal A}\frac{(\hat p_k-p_k)^2}{p_k}+O\Big(\sum_{k\in\mathcal A}|\delta_k|^3\Big).
\]
Each $\hat p_k$ has bias $O(1/n)$ and variance $np_k(1-p_k)/(\bar\alpha+n)^2$, so $\E(\hat p_k-p_k)^2=p_k(1-p_k)/n+O(1/n^2)$ and $\E|\delta_k|^3=O(n^{-3/2})$. As in the proof of \Cref{prop:social}(ii), the complement of the event has exponentially small probability, and on it the regret is $O(\log n)$ because every $\hat p_k\ge\alpha_k/(\bar\alpha+n)$. Hence
\[
\E\,\KL(p\,\|\,\hat p)=\frac{\alpha_0}{n}+\frac1{2n}\sum_{k\in\mathcal A}(1-p_k)+o\Big(\frac1n\Big)=\frac1n\Big[\frac{K_+-1}{2}+\alpha_0\Big]+o\Big(\frac1n\Big),
\]
because $\sum_{k\in\mathcal A}(1-p_k)=K_+-1$. This is \eqref{eq:Kregret}.
\end{proof}

\subsection{\texorpdfstring{Proof for Appendix~\ref{app:universal}}{Proof for the universal portfolio}}

\begin{proof}[Proof of \Cref{prop:universal}]
(i) With $\mu=\mathrm{Beta}(\alpha,\beta)$ and $V_{j-1}(f)=m^{j-1}f^{k_{j-1}}(1-f)^{j-1-k_{j-1}}$, the ratio of integrals in part (i) is $B(\alpha+k_{j-1}+1,\beta+j-1-k_{j-1})/B(\alpha+k_{j-1},\beta+j-1-k_{j-1})=(\alpha+k_{j-1})/(\alpha+\beta+j-1)$, which is \eqref{eq:adjud} at exposure $j$. Since $V_j(f)=V_{j-1}(f)\,m[f\one\{\omega_j=\mathrm{T}\}+(1-f)\one\{\omega_j=\mathrm{C}\}]$, the learned policy's growth factor at exposure $j$ is $\int V_j\,d\mu/\int V_{j-1}\,d\mu$, and the product telescopes \citep[Lemma 2.5]{cover1991universal}: $\hat V_n=\prod_{j\le n}\int V_j\,d\mu/\int V_{j-1}\,d\mu=\int V_n\,d\mu$, and the integral is $m^nB(\alpha+k_n,\beta+n-k_n)/B(\alpha,\beta)$.

(ii) The lower bound holds for every prior: since $\mu$ is a probability measure, $\hat V_n=\int_0^1V_n(f)\,\mu(df)\le V_n^*$. For the upper bound under the uniform prior, \eqref{eq:univwealth} gives $\hat V_n=m^n\,k_n!(n-k_n)!/(n+1)!$, so
\[
\frac{V_n^*}{\hat V_n}=(n+1)\binom{n}{k_n}\Big(\frac{k_n}{n}\Big)^{k_n}\Big(1-\frac{k_n}{n}\Big)^{n-k_n}\le n+1,
\]
since the middle factor is the probability that a binomial$(n,k_n/n)$ variable equals $k_n$, which is at most one, and equals one exactly when $k_n\in\{0,n\}$. This is Theorem~1 of \citet{cover1996side} for two assets, where it is stated as $V_n^*/\hat V_n\le\binom{n+1}{1}$. The bound for the prior $\alpha=\beta=\tfrac12$ is Theorem~2 of \citet{cover1996side} for two assets, $V_n^*/\hat V_n\le\Gamma(\tfrac12)\Gamma(n+1)/\Gamma(n+\tfrac12)\le2\sqrt{n+1}$; the same two bounds appear as Theorems~9.3 and 9.4 of \citet{cesabianchi2006prediction}, where the uniform-prior case is the Laplace mixture and the half-prior case the Krichevsky--Trofimov mixture.

(iii) Subtracting $\log V_n(f)$ from both sides of the upper bound in part (ii) gives the inequality. For $f\equiv1$, $V_n(1)=0$ once $k_n<n$, while $\hat V_n>0$ by \eqref{eq:univwealth}, so the difference is $+\infty$.

(iv) Keeping separate counts for each situation runs a separate universal portfolio on the subsequence of exposures to that situation, which is Definition~2 of \citet{cover1996side}. The lineage's wealth over $t$ generations is the product over situations of the per-situation wealths, and the best fixed adjudication in hindsight maximizes each factor separately, so the log gaps add. By part (ii), the gap in situation $s$ is at most $\log(n_{t+1}(s)+1)$, and $\sum_s\log(n_{t+1}(s)+1)\le|\Scal|\log(t/|\Scal|+1)$ by the concavity of the logarithm, since $\sum_sn_{t+1}(s)=t$. The bound is Theorem~3 of \citet{cover1996side} with two assets and $|\Scal|$ states of side information.

(v) Under \Cref{ass:iid}, $k_n/n\to\pH(s)$ almost surely by the strong law, so $\tfrac1n\log V_n^*=\log m+\tfrac{k_n}{n}\log\tfrac{k_n}{n}+(1-\tfrac{k_n}{n})\log(1-\tfrac{k_n}{n})$ converges to $I_{\pH(s)}(\pH(s))$; under the uniform prior, by part (ii), $\tfrac1n\log\hat V_n$ is within $\log(n+1)/n\to0$ of it.
\end{proof}

\newpage
\singlespacing
\bibliographystyle{jf3}
\bibliography{references}

\end{document}